%% file: ClassicThesis.tex
\RequirePackage{fix-cm} 
\documentclass[ twoside,openright,titlepage,numbers=noenddot,headinclude,
                footinclude=true,cleardoublepage=empty,abstractoff, 
                BCOR=5mm,paper=a4,fontsize=11pt,
                ngerman,american,%
                ]{scrreprt}

\input{classicthesis-config}

\usepackage{geometry}
\usepackage[skip=10pt plus1pt, indent=20pt]{parskip} 

\usepackage{float}
\usepackage{aliascnt}
\newaliascnt{eqfloat}{equation}
\newfloat{eqfloat}{h}{eqflts}
\floatname{eqfloat}{Equation}

\newcommand*{\ORGeqfloat}{}
\let\ORGeqfloat\eqfloat
\def\eqfloat{%
  \let\ORIGINALcaption\caption
  \def\caption{%
    \addtocounter{equation}{-1}%
    \ORIGINALcaption
  }%
  \ORGeqfloat
}

\usepackage{forest}
\definecolor{folderbg}{RGB}{124,166,198}
\definecolor{folderborder}{RGB}{110,144,169}

\def\Size{4pt}
\tikzset{
  folder/.pic={
    \filldraw[draw=folderborder,top color=folderbg!50,bottom color=folderbg]
      (-1.05*\Size,0.2\Size+5pt) rectangle ++(.75*\Size,-0.2\Size-5pt);  
    \filldraw[draw=folderborder,top color=folderbg!50,bottom color=folderbg]
      (-1.15*\Size,-\Size) rectangle (1.15*\Size,\Size);
  }
}

\usepackage{pdfpages}
\begin{document}
\frenchspacing
\raggedbottom
\selectlanguage{american} 
\pagenumbering{roman}
\pagestyle{plain}
\includepdf[pages=-]{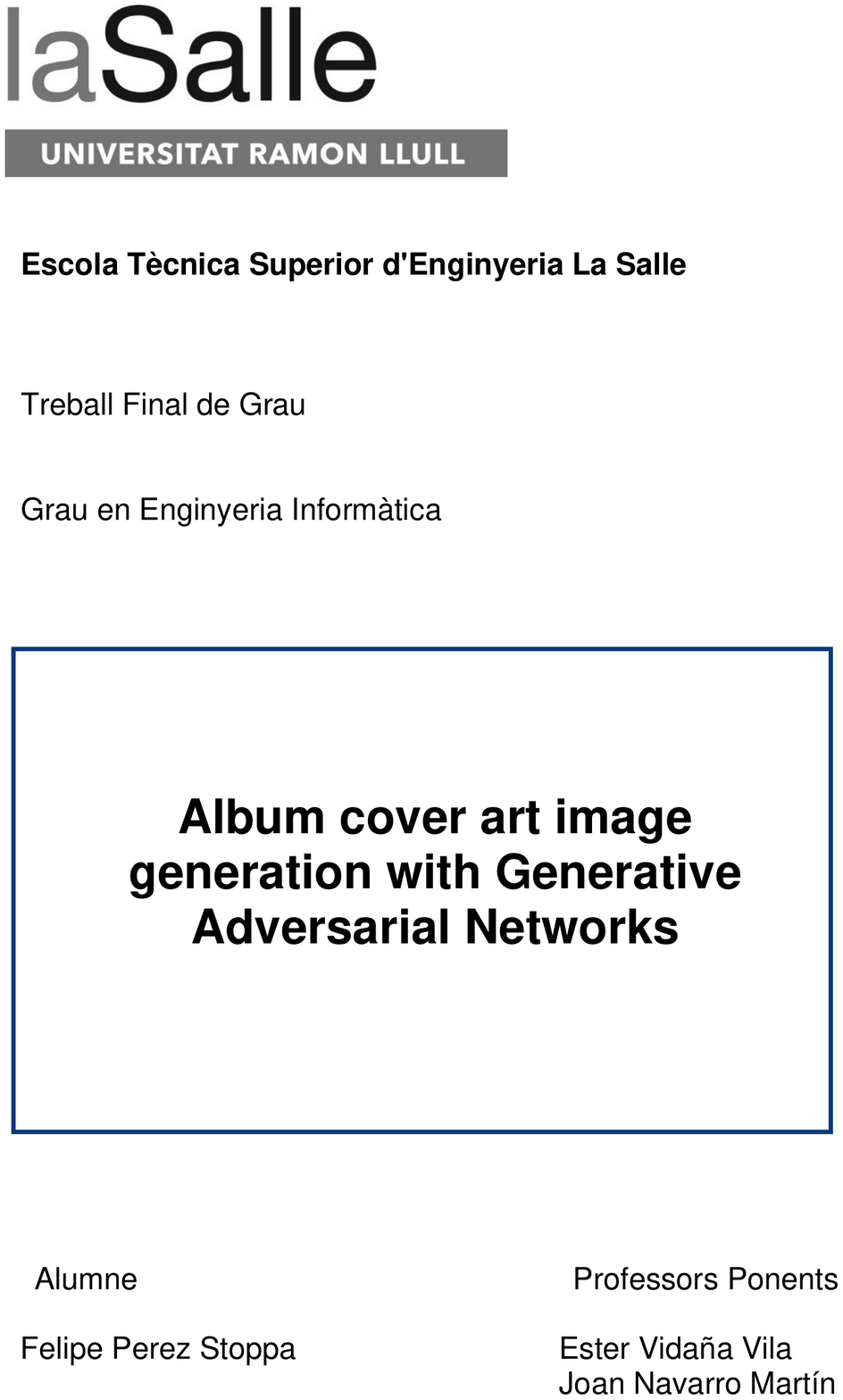}
\include{FrontBackmatter/Titlepage}
\include{FrontBackmatter/Titleback}
\cleardoublepage\include{FrontBackmatter/Abstract}
\pagestyle{scrheadings}
\cleardoublepage\include{FrontBackmatter/Contents}
\cleardoublepage\pagenumbering{arabic}


\cleardoublepage
\include{Chapters/00Introduction}

\cleardoublepage

\include{Chapters/01Theoretical}
\include{Chapters/02Setup}
\include{Chapters/IntroGAN}
\include{Chapters/CelebrityFacesDCGAN}
\include{Chapters/StyleGAN2}
\include{Chapters/05Conclusion}




\appendix
\part{Appendix}
\include{Chapters/Appendix}
\include{FrontBackmatter/Bibliography}
\end{document}

%% file: classicthesis-config.tex
\PassOptionsToPackage{utf8}{inputenc}
	\usepackage{inputenc}

\PassOptionsToPackage{eulerchapternumbers,listings,drafting,%
					 pdfspacing,
					 subfig,beramono,eulermath,parts}{classicthesis}                                        

\newcommand{\myTitle}{Album cover art image generation with Generative Adversarial Networks\xspace}

\newcommand{\myName}{Felipe Perez Stoppa\xspace}

\newcommand{\myFaculty}{Put data here\xspace}

\newcommand{\myUni}{Universitat Ramon Llull, La salle Campus Barcelona\xspace}

\newcommand{\myTime}{June 2022\xspace}
\newcommand{\myVersion}{ \xspace}

\newcounter{dummy} 
\providecommand{\mLyX}{L\kern-.1667em\lower.25em\hbox{Y}\kern-.125emX\@}

	\usepackage{babel}                  
    \usepackage{csvsimple}

\usepackage{csquotes}
\PassOptionsToPackage{%
	backend=bibtex8,bibencoding=ascii,%
	language=auto,%
	style=numeric-comp,%
    sorting=none, 
    maxbibnames=10, 
    natbib=true 
}{biblatex}
    \usepackage{biblatex}

    \usepackage{amsmath}

\PassOptionsToPackage{T1}{fontenc} 
    \usepackage{fontenc}     
\usepackage{textcomp} 
\usepackage{scrhack} 
\usepackage{xspace} 
\usepackage{mparhack} 
\usepackage{fixltx2e} 
\PassOptionsToPackage{printonlyused,smaller}{acronym} 
    \usepackage{acronym} 
    
\usepackage{tabularx} 
\newcommand{\tableheadline}[1]{\multicolumn{1}{c}{\spacedlowsmallcaps{#1}}}
\newcommand{\myfloatalign}{\centering} 
\usepackage{caption}
\usepackage{subfig}  

\usepackage{listings} 

\PassOptionsToPackage{pdftex,hyperfootnotes=false,pdfpagelabels}{hyperref}
    \usepackage{hyperref}  
\PassOptionsToPackage{pdftex}{graphicx}
    \usepackage{graphicx}

\hypersetup{%
    colorlinks=true, linktocpage=true, pdfstartpage=3, pdfstartview=FitV,%
    breaklinks=true, pdfpagemode=UseNone, pageanchor=true, pdfpagemode=UseOutlines,%
    plainpages=false, bookmarksnumbered, bookmarksopen=true, bookmarksopenlevel=1,%
    hypertexnames=true, pdfhighlight=/O,
    urlcolor=webbrown, linkcolor=RoyalBlue, citecolor=webgreen, 
    pdftitle={\myTitle},%
    pdfauthor={\textcopyright\ \myName, \myUni, \myFaculty},%
    pdfsubject={},%
    pdfkeywords={},%
    pdfcreator={pdfLaTeX},%
    pdfproducer={LaTeX with hyperref and classicthesis}%
}   

\makeatletter
\@ifpackageloaded{babel}%
    {%
       \addto\extrasamerican{%
                }%
       \addto\extrasngerman{%
                }%
    }{\relax}
\makeatother

\usepackage{classicthesis} 



%% file: FrontBackmatter/Titlepage.tex
\begin{titlepage}
    \begin{center}
        \large  

        \hfill

        \vfill

        \begingroup
            \color{Maroon}\spacedallcaps{\myTitle} \\ \bigskip
        \endgroup

        \spacedlowsmallcaps{\myName}

        \vfill



        \myTime

        \vfill                      

    \end{center}  
\end{titlepage}   

%% file: FrontBackmatter/Titleback.tex
\thispagestyle{empty}

\hfill

\vfill

\noindent\myName: \textit{\myTitle,} 
\textcopyright\ \myTime

%
%
%
%
%

%% file: FrontBackmatter/Abstract.tex
\pdfbookmark[1]{Abstract}{Abstract}
\begingroup
\let\clearpage\relax
\let\cleardoublepage\relax
\let\cleardoublepage\relax

\chapter*{Abstract}
\acp{GAN} were introduced by Goodfellow in 2014, and since then have become popular for constructing generative artificial intelligence models. However, the drawbacks of such networks are numerous, like their longer training times, their sensitivity to hyperparameter tuning, several types of loss and optimization functions and other difficulties like mode collapse.

Current applications of \acp{GAN} include generating photo-realistic human faces, animals and objects. However, I wanted to explore the artistic ability of \acp{GAN} in more detail, by using existing models and learning from them. This dissertation covers the basics of neural networks and works its way up to the particular aspects of \acp{GAN}, together with experimentation and modification of existing available models, from least complex to most. The intention is to see if state of the art \acp{GAN} (specifically StyleGAN2) can generate album art covers and if it is possible to tailor them by genre.

This was attempted by first familiarizing myself with 3 existing \acp{GAN} architectures, including the state of the art StyleGAN2. The StyleGAN2 code was used to train a model with a dataset containing 80K album cover images, then used to style images by picking curated images and mixing their styles.

\refstepcounter{dummy}
    \pdfbookmark[1]{Keywords}{Keywords}
    \markboth{\spacedlowsmallcaps{Keywords}}{\spacedlowsmallcaps{Keywords}}
    \chapter*{Keywords}
Machine learning | Neural Networks | Deep Learning | Generative model | Generative 
Adversarial Network | Style mixing

\pagebreak

\chapter*{Resumen}
Las \ac{GAN} fueron introducidos por primera vez por Goodfellow en 2014 y desde entonces se han hecho populares para construir modelos de inteligencia artificial generativa. Sin embargo, sus inconvenientes son numerosos, principalmente sus tiempos de entrenamiento largos, su sensibilidad a los cambios en los hiperparámetros, varias funciones de pérdida y optimización y otras dificultades como el 'mode collapse'.

Las aplicaciones actuales de las \acp{GAN} incluyen la generación de rostros humanos fotorrealistas y imágenes de animales y objetos. Sin embargo, quería explorar la habilidad artística de las \acp{GAN} con más detalle. Este trabajo cubre los conceptos básicos de las redes neuronales y avanza hasta los aspectos particulares de las \acp{GAN}, junto con la experimentación y modificación de modelos existentes, desde los menos hasta los más complejos. La intención es ver si las \acp{GAN} (concretamente el StyleGAN2) de última generación puede generar portadas de álbum y si es posible adaptarlas por género, por ejemplo, rock.

Esto lo intenté familiarizándome primero con 3 arquitecturas \ac{GAN} existentes, incluido el StyleGAN2. El código StyleGAN2 se utilizó para
entrenar un modelo con un conjunto de datos que contiene 80000 imágenes de portadas de álbumes y luego utilizarlo para generar imágenes eligiendo imágenes concretas y mezclando sus estilos.

\refstepcounter{dummy}
    \pdfbookmark[1]{Palabras Clave}{Palabras Clave}
    \markboth{\spacedlowsmallcaps{Palabras Clave}}{\spacedlowsmallcaps{Palabras Clave}}
    \chapter*{Palabras Clave}
Aprendizaje automático | Redes neuronales | Aprendizaje profundo | Modelos 
generativos | Generative Adversarial Network | Style mixing

\pagebreak

\chapter*{Resum}

Els \acp{GAN} van ser introduïts per primera vegada per Goodfellow el 2014, i des de llavors s'han fet populars per construir models d'intel·ligència artificial generativa. Tot i això, els seus inconvenients són nombrosos, principalment els seus temps d'entrenament més llargs, la seva sensibilitat a la modificació d'hiperparàmetres, diversos tipus de funcions de pèrdua i optimització i altres dificultats com el 'mode collapse'.

Les aplicacions actuals dels \ac{GAN}  inclouen la generació de cares humanes fotorealistes i imatges d'animals i objectes. Tanmateix, volia explorar l'habilitat artística de les \ac{GAN} amb més detall. Aquest TFG cobreix els fonaments de les xarxes neuronals i els aspectes particulars dels \ac{GAN}, juntament amb l'experimentació i la modificació dels models disponibles existents, des de menys complexos fins als mes complexos. La intenció és veure si les \acp{GAN} (concretament l'StyleGAN2) d'última generació poden generar portades d'àlbum i si és possible adaptar-les per gènere, per exemple, rock.

Això ho vaig fer familiaritzant-me primer amb 3 arquitectures \ac{GAN} existents, inclòs el StyleGAN2. El codi del StyleGAN2 es va utilitzar per entrenar un model amb un conjunt de dades que conté 80000 imatges de portades d'àlbums, i després utilitzar-lo per generar imatges triant imatges concretes i barrejant els seus estils.

\refstepcounter{dummy}
    \pdfbookmark[1]{Paraules Clau}{Paraules Clau}
    \markboth{\spacedlowsmallcaps{Paraules Clau}}{\spacedlowsmallcaps{Paraules Clau}}
    \chapter*{Paraules Clau}
Aprenentatge automàtic | Xarxes neuronals | Aprenentatge profund | Models
generatius | Generative Adversarial Network | Style mixing

\endgroup

%% file: FrontBackmatter/Contents.tex
\refstepcounter{dummy}
\pdfbookmark[1]{\contentsname}{tableofcontents}
\setcounter{tocdepth}{2} 
\setcounter{secnumdepth}{3} 
\manualmark
\markboth{\spacedlowsmallcaps{\contentsname}}{\spacedlowsmallcaps{\contentsname}}
\tableofcontents 
\automark[section]{chapter}
\renewcommand{\chaptermark}[1]{\markboth{\spacedlowsmallcaps{#1}}{\spacedlowsmallcaps{#1}}}
\renewcommand{\sectionmark}[1]{\markright{\thesection\enspace\spacedlowsmallcaps{#1}}}
\clearpage

\begingroup 
    \let\clearpage\relax
    \let\cleardoublepage\relax
    \let\cleardoublepage\relax
    \refstepcounter{dummy}
    \pdfbookmark[1]{\listfigurename}{lof}
    \listoffigures

    \vspace{8ex}

    \refstepcounter{dummy}
    \pdfbookmark[1]{\listtablename}{lot}
    \listoftables
        
    \vspace{8ex}
    
    \refstepcounter{dummy}
    \pdfbookmark[1]{\lstlistlistingname}{lol}
    \lstlistoflistings 

    \vspace{8ex}
       
    \refstepcounter{dummy}
    \pdfbookmark[1]{Acronyms}{acronyms}
    \markboth{\spacedlowsmallcaps{Acronyms}}{\spacedlowsmallcaps{Acronyms}}
    \chapter*{Acronyms}
    \begin{acronym}[UMLX]
        \acro{MSE}{Mean squared error}
        \acro{BCE}{Binary cross entropy}
        \acro{CCE}{Categorical cross entropy}
        \acro{GD}{Gradient Descent}
        \acro{SGD}{Stochastic Gradient Descent}
        \acro{MBSGD}{Mini Batch Stochastic Gradient Descent}
        \acro{BN}{Batch Normalization}
        \acro{CNN}{Convolutional Neural Network}
        \acro{D}{Discriminator}
        \acro{G}{Generator}
        \acro{GAN}{Generative Adversarial Network}
        \acro{DCGAN}{Deep Convolutional Generative Adversarial Network}
        \acro{ML}{Machine Learning}
        \acro{AI}{Artificial intelligence}
        \acro{NN}{Neural Network}
        \acro{STD}{Standard deviation}
        \acro{IS}{Inception score}
        \acro{FID}{Fréchet inception distance}
        \acro{MNIST}{Modified National Institute of Standards and Technology}
        \acro{OS}{Operating system}
        \acro{WSL}{Windows Subsystem for Linux}
        \acro{VM}{Virtual Machine}
        \acro{LTS}{Long Term Support}
        \acro{GPU}{Graphics Processing Unit}
        \acro{JS}{JavaScript}
        \acro{GUI}{Graphical User Interface}
        \acro{TF}{TensorFlow}
        \acro{CPU}{Central Processing Unit}
        \acro{FAIR}{Facebook's AI Research}
        \acro{API}{Application Programming Interface}
        \acro{URL}{Uniform Resource Locator}
        \acro{RGB}{Red, Green, Blue}
        \acro{ReL}{Rectified linear activation}
        \acro{ReLU}{Rectified linear activation unit}
        \acro{Tanh}{Hyperbolic tangent}
        \acro{ADA}{Adaptive Discriminator augmentation}
        \acro{FFHQ}{Flickr-Faces-HQ Dataset}
        \acro{VAE}{Variational Autoencoder}
        \acro{KIMG}{Thousands of Images}
        \acrodefplural{GPU}{Graphics Processing Units}
        \acrodefplural{CPU}{Central Processing Units}
        \acrodefplural{URL}{Uniform Resource Locators}
        \acrodefplural{VM}{Virtual Machines}
        \acrodefplural{GAN}{Generative Adversarial Networks}
        \acrodefplural{DCGAN}{Deep Convolutional Generative Adversarial Networks}
    \end{acronym}                     
\endgroup

%% file: Chapters/00Introduction.tex
\chapter{Introduction}\label{ch:introduction}

\section{Motivation and Framework}
The initial motivation that drew me to research \acp{GAN} and attempt to generate images with them was when we covered them at the end of our data mining course here at La Salle Barcelona. The last assignment consisted in picking a research paper and make a presentation with our findings.

My project partner and I chose one related to \acp{GAN}. The paper \cite{skandarani_gans_2021} looked into the capability of \acp{GAN} for generating medical images to augment existing datasets. This was because in the case of medical images, due to privacy concerns and the sensitivity of the subject, there are not as many images available in these datasets. As I read through it more and more times, I became more interested in knowing more about generative models. Eventually I landed on the 'This person does not exist'\footnote{\url{https://this-person-does-not-exist.com/en}} website (as most people that become interested with \acp{GAN} do) and I was amazed at the quality and precision with which an \ac{AI} could generate human faces.

\begin{figure}[bth]
        \myfloatalign
        {\label{fig:avatar1}
        \includegraphics[width=0.45\linewidth]{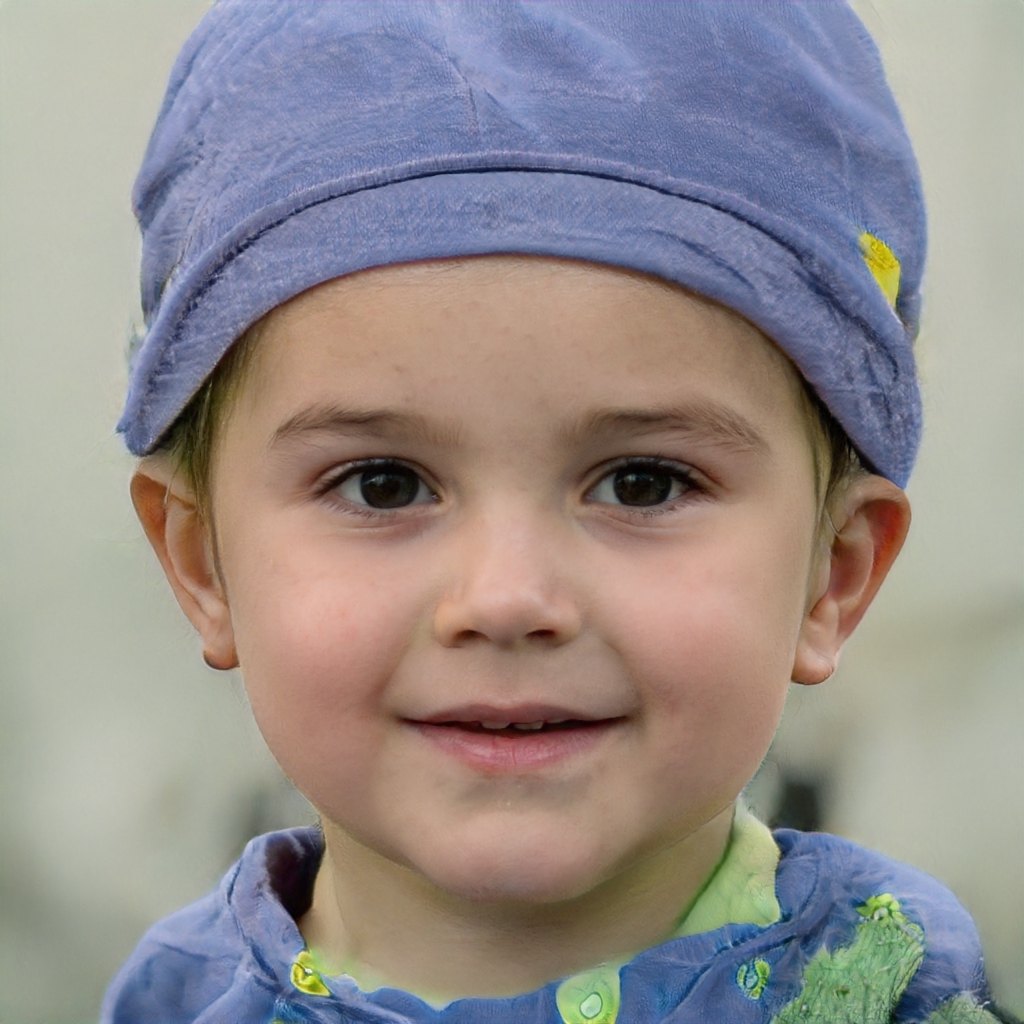}} \quad
        {\label{fig:avatar2}
        \includegraphics[width=0.45\linewidth]{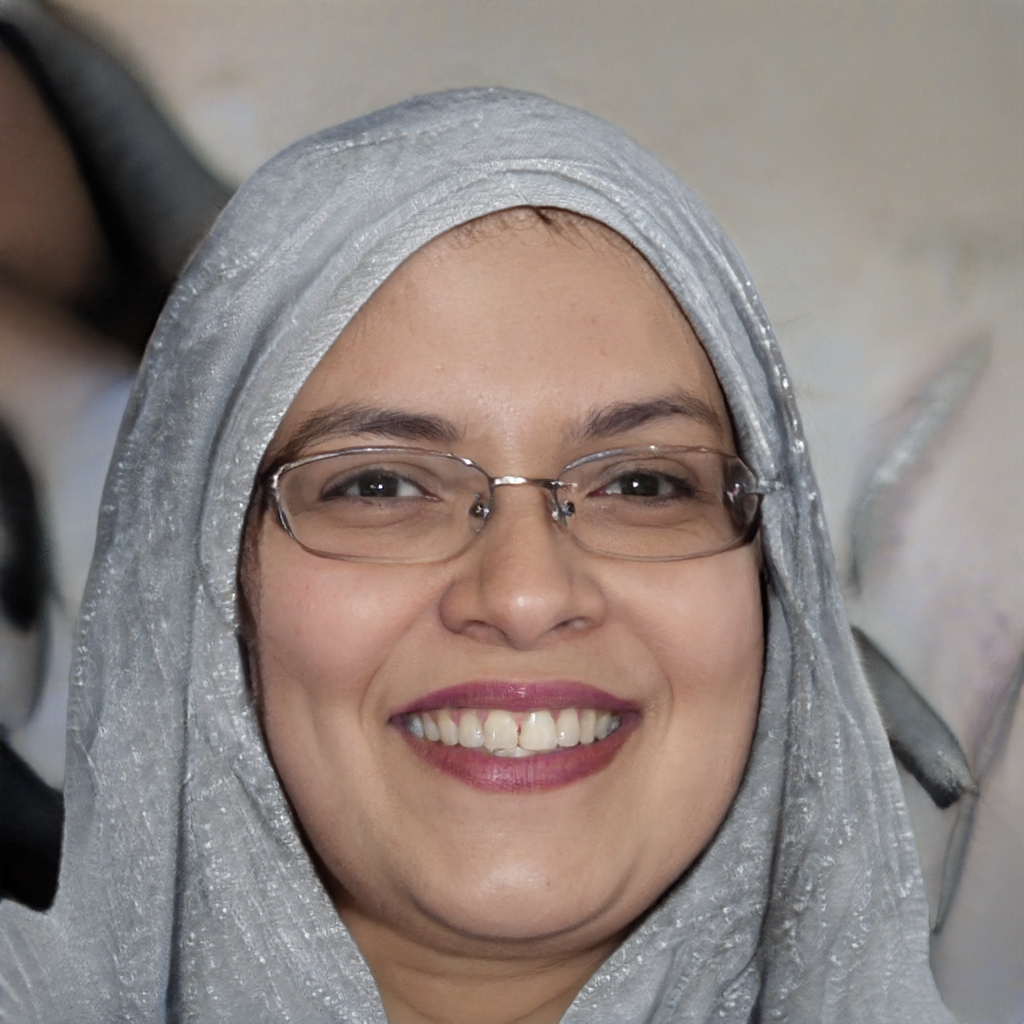}}\\
        \caption{2 images generated with a model using the original StyleGAN architecture. Images generated using the \url{https://this-person-does-not-exist.com/en} website.} 
\end{figure}

It also tied into the previously mentioned paper in more ways than one. Not only were \acp{GAN} being used for generating these faces, but the StyleGAN2 architecture made it possible to generate high quality credible faces with limited data.

This prompted me to want to further my knowledge in \ac{AI} and \acp{GAN}, specifically the StyleGAN2 architecture. I chose I wanted to generate album cover art as a goal to aid in researching the StyleGAN2 architecture, as this played into my interests in music and curiosity of seeing the results. It was also motivated by the high variability and number of features that album covers contain (text, color, shapes, objects, people and faces, etc.) and seeing which features and styles a StyleGAN2 model could pick up on.

There were also other architectures that came up when doing preliminary research for this task, specifically \acp{VAE}. I chose to focus my research on \acp{GAN} since some of them actually incorporate elements of \acp{VAE} architectures, specifically StyleGAN2.

\pagebreak

\section{Goal of this project}\label{goals}

The goal of this dissertation is to understand how the StyleGAN2 architecture works by generating album cover art. This involves not only reading the research paper on StyleGAN2 \cite{karras_training_2020} but also using the researcher's code available on GitHub \footnote{\label{stylegan2_ref}\url{https://github.com/NVlabs/stylegan2-ada-pytorch}} to train and generate images of my own, and be able to style them.

StyleGAN2 is \textquote{an alternative generator architecture for \acp{GAN} [...] . The new architecture leads to an automatically learned, unsupervised separation of high-level attributes (e.g., pose and identity when trained on human faces)} \cite{noauthor_nvlabsstylegan_2022}.

StyleGAN2 uses a technique called styling. Styling is a technique which involves mixing the learnt high level attributes present in vectors of images that have been projected to the latent space.

However, before understanding how StyleGAN2 works, it is important to introduce a theoretical framework of the necessary concepts for machine learning. Concepts such as \acp{NN}, their building blocks, their relation to \acp{GAN} and why StyleGAN2 is different than other models before it and what mechanisms allow it to style images.

The resulting album covers will be evaluated both subjectively and objectively.

The objective evaluation will consist in looking at the \ac{FID} result of the trained model, and comparing it to other scores obtained by the same \ac{GAN} trained on different datasets.

Since art is subjective, a successful generated album cover will be defined as an album cover in which the title (if any), color and background are distinct elements that have been picked up by the model and are manifested in the generated covers. This in and of itself is also partly subjective, since it could be argued that a black square is a valid album cover. However, since the task of a \ac{GAN} is to generate images that model the input distribution, and the input distribution will be a variety of album covers from different genres and styles, it is reasonable to expect the model to replicate features present in most covers.

\section{Objectives}\label{objectives}

In order to understand how the StyleGAN2 works and generating album cover art, the following objectives have been defined:

\begin{enumerate}

\item{\textbf{Set up a repeatable development and training environment by using a custom Docker container}}: The repeatable environment will consist of using a custom Docker container based on the Dockerfile provided in the StyleGAN2 GitHub page\footnote{\url{https://github.com/NVlabs/stylegan2-ada-pytorch/blob/main/Dockerfile}}. Some libraries will need to be added since this image will not only be used for training the model but also writing and running custom python code, which includes tasks such as creating datasets needed and plotting results.

\item{\textbf{Introducing theoretical framework of the model by providing explanations of the necessary concepts}}: The goal is to analyze the knowledge necessary to understand the architecture of \acp{GAN}, how they are trained and how they learn, and the benefits, limitations and challenges of different approaches of building \acp{GAN}

\item{\textbf{Familiarize myself with existing \ac{GAN} architecture by analyzing and experimenting with 3 different \ac{GAN} models}}: To aid in understanding the building blocks of \acp{GAN}, 3 different models will be Analyzed and used for experimentation.

    \begin{enumerate}
       \item{\textbf{Introductory GAN}}: Model programmed using the Keras Python library trained on the \ac{MNIST} handwritten digits dataset to generate black and white handwritten digits at 24x24 resolution.
       
       \item{\textbf{DCGAN}}: Model originally used to generate faces using the CelebA\footnote{\url{https://mmlab.ie.cuhk.edu.hk/projects/CelebA.html}} dataset. The goal is to train it with the Album covers dataset\footnote{\label{album_cover_ds}\url{https://www.kaggle.com/datasets/greg115/album-covers-images}} for this application. This model generates images in color as opposed to black and white like the previous one and the output resolution is higher, at 64x64.
       
       \item{\textbf{StyleGAN2}}: Model with progressive architecture and other improvements such as \ac{ADA}. The goal is to also train it on the Album covers dataset \footnotemark[\value{footnote}] and generate styled images by mapping select images to the model's latent space and mixing their styles.
    \end{enumerate}

Their architectures must be analyzed, as well as the dataset used, what format of input data the model expects as well as what output data it will generate, the training process and all hyperparameters used. Any nuances of the architecture will need to be discussed.

The experimentation phase should involve generating images with the different models and the discussing the results. Any improvements done to the architecture, the data or the initialization of the model will also be discussed where relevant.

\item{\textbf{Generate styled images by using the trained StyleGAN2 model}}: Using the StyleGAN2 repository's\footnote{\url{https://github.com/NVlabs/stylegan2-ada-pytorch}} code to train a model and use helper scripts provided to map images to the latent space. Once mapped, the resulting latent space vectors can be used to experiment with different methods of style mixing \ref{goals}.

\end{enumerate}

\pagebreak

\section{Related Work}

This section covers all related work, specifically the most prominent works that I drew from for inspiration and/or used as research material.

\subsection{Generative Adversarial Networks}

\citet{goodfellow_generative_2014} first proposed the \ac{GAN} framework with their research paper titled 'Generative Adversarial Networks'. They propose an architecture where there are 2 networks, the Generator and the Discriminator, competing to 'outsmart' the other. The Generator can be thought of as an art counterfeiter, while the Discriminator can be thought of as a police person trying to tell counterfeits from real pieces of art. The Generator is trained to maximize the chance the Discriminator will make a mistake i.e. trained to fool the Discriminator, while the Discriminator continually learns and becomes better at telling fakes from  reals\cite{goodfellow_generative_2014}.

\begin{figure}[bth]
\centering
\includegraphics[width=0.7\linewidth]{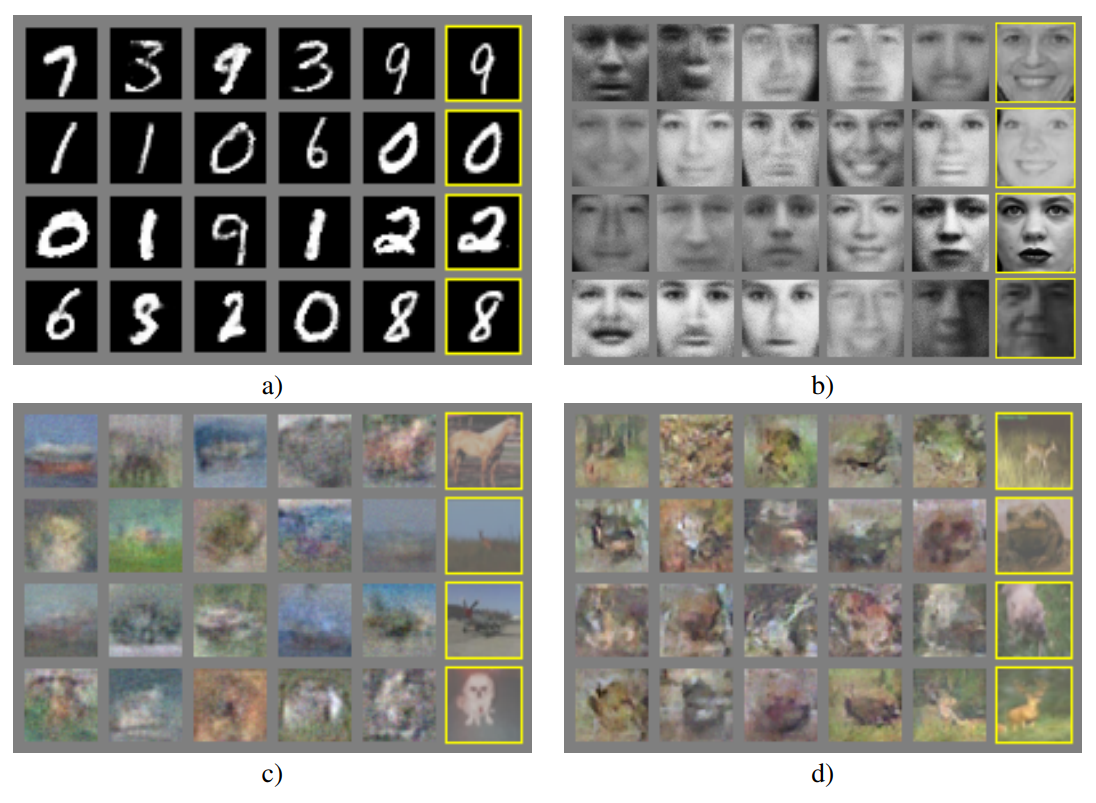}
\caption{Results from training the \ac{GAN} with different datasets. The rightmost column shows the nearest training example to the generated images. Source: \cite{goodfellow_generative_2014}.}
\label{fig:first_gan_results}
\end{figure}

Datasets: a) MNIST\footnote{\url{http://yann.lecun.com/exdb/mnist/}} b) TFD\footnote{\url{https://www.kaggle.com/general/50987}} c) CIFAR-10\footnote{\url{https://www.cs.toronto.edu/~kriz/cifar.html}} (fully connected model) d) CIFAR-10\footnotemark[\value{footnote}] (convolutional discriminator and “deconvolutional” generator).

This paper introduced the base knowledge needed to understand \ac{GAN}. A notable fact to extract from the paper is that these models lack theoretical guarantees, but their practical performance shows they are viable as a framework for generative models \cite{goodfellow_generative_2014}.

\subsection{GANs for Medical Image Synthesis}

This research paper was the first I heard about \acp{GAN}. The study consists of training different \ac{GAN} architectures on select medical image datasets with the goal to generate synthetic data that could be used to extend the available data. They tested a variety of \acp{GAN}, and while some of them achieved results that could fool experts in a visual Turing test, they data they generated was found to not always be reliable.

\begin{figure}[bth]
\centering
\includegraphics[width=0.7\linewidth]{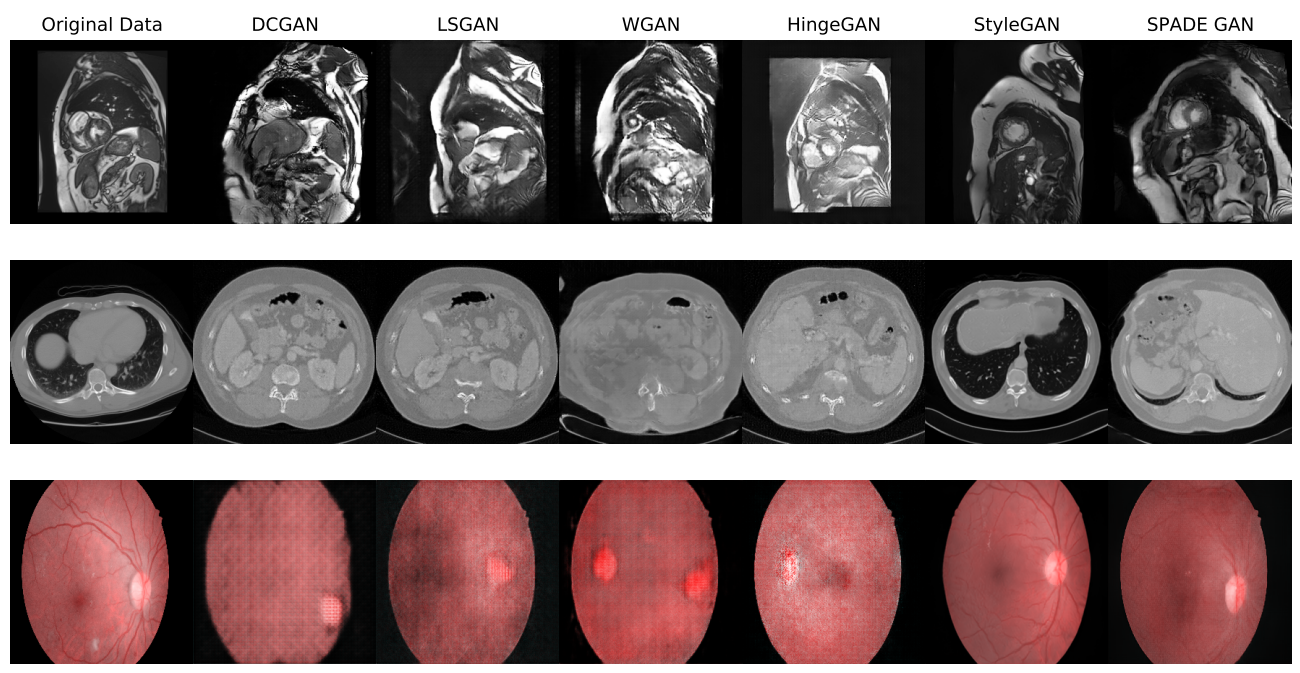}
\caption{Examples of generated images for each GAN on the ACDC, SLiver07 and
IDRID datasets. Source: \cite{skandarani_gans_2021}.}
\label{fig:medical_image_gan}
\end{figure}

Datasets: ACDC\footnote{\url{https://www.creatis.insa-lyon.fr/Challenge/acdc/}}, SLiver07\footnote{\url{https://sliver07.grand-challenge.org/}}, IDRID\footnote{\url{https://idrid.grand-challenge.org/}}.

This research paper sparked my interest since it took on a major task, expanding medical image datasets. Not only are the images themselves seemingly very complex, but the nature of the subject is also delicate, as unreliable data making its way into datasets that are used by other researchers could spell disaster.

\subsection{StyleGAN2 - Training Generative Adversarial Networks with Limited Data}
\begin{figure}[bth]
\centering
\includegraphics[width=1\linewidth]{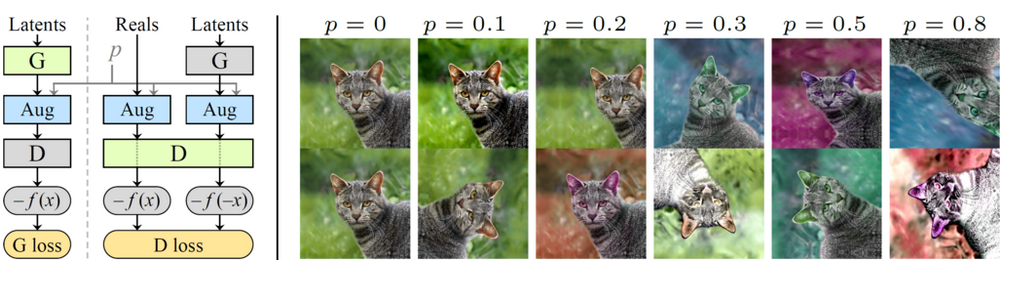}
\caption{Different augmentations applied with a certain probability. Note more than one augmentation can be applied per image, as is the case for the high $p$ values \cite{karras_progressive_2018}. Source: \cite{karras_training_2020}.}
\label{fig:progressive_arch_ada}
\end{figure}

This implementation of a \ac{GAN} is based on the StyleGAN paper \cite{karras_progressive_2018}, adding techniques that allow the model training to be more stable and require less data. The data augmentation techniques used in StyleGAN2 \cite{karras_training_2020} involve probabilistically modifying the input data with certain transforms to stabilize the training. A diagram and examples of augmented images is shown in figure \ref{fig:progressive_arch_ada}.

\begin{figure}[bth]
\centering
\includegraphics[width=0.8\linewidth]{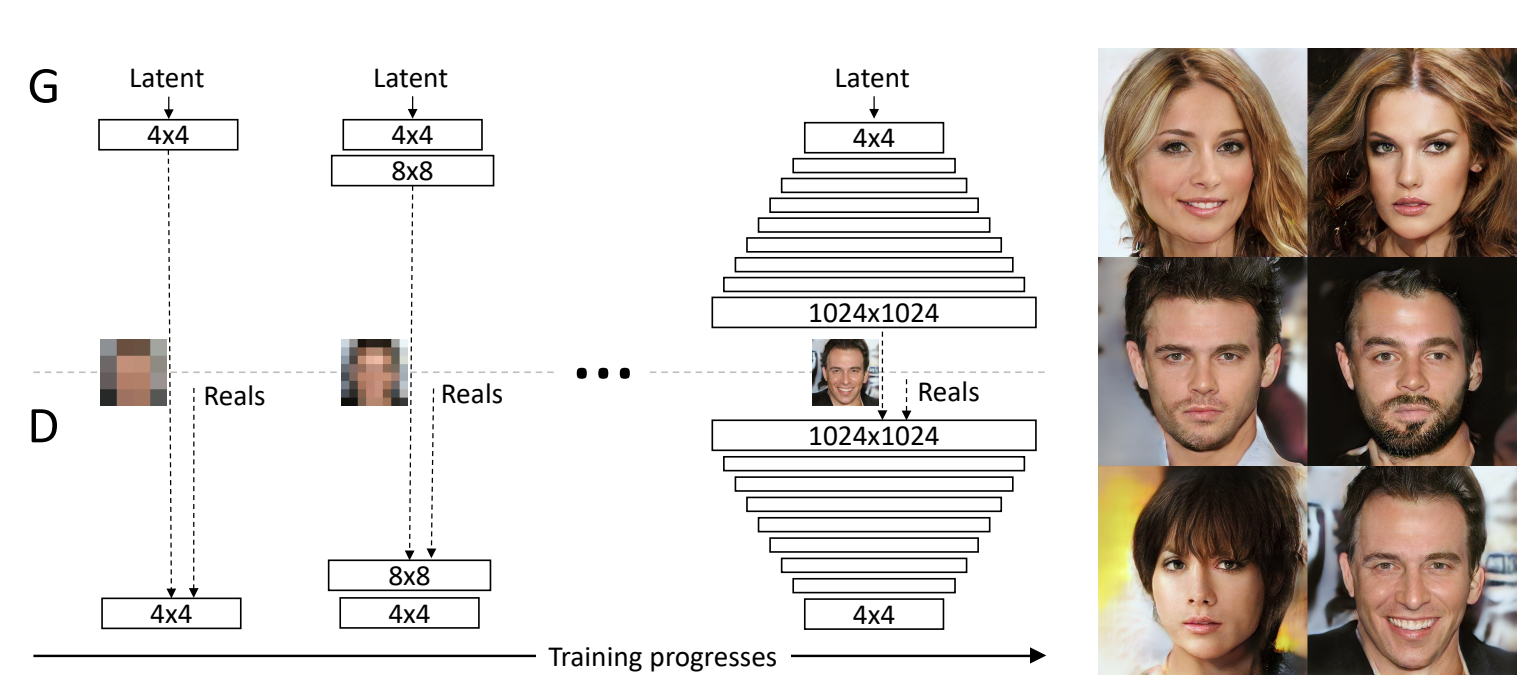}
\caption{Progressive architecture illustrated with results from the Celeba dataset\textsuperscript{A}. The 'G' Represents the generator and the 'D' the Discriminator. Source: \cite{karras_progressive_2018}.}
\small\textsuperscript{A} https://mmlab.ie.cuhk.edu.hk/projects/CelebA.html
\label{fig:progressive_arch_1}
\end{figure}

The core of the \ac{GAN} however, is its progressive architecture detailed in the first paper by the NVIDIA researchers, where it will progressively add/train layers to both the generator and discriminator, in an attempt to capture the overall features of the input data and work its way to the fine details. This mainly also allows for generation of higher resolution images without loosing quality. In figure \ref{fig:progressive_arch_1} you can observe the progressive structure of the \ac{GAN} and examples of generated images.

\section{Structure of the document}

The document is structured in the following parts:

\begin{itemize}
    \item\textbf{Theoretical Framework}: Chapter \ref{ch:theoretical} introduces 2 types of networks; \ac{NN} and \acp{CNN}, covering their building blocks and the mechanisms that allow them learn. The specifics of \acp{GAN} will also be covered, like their adversarial nature and the 2 networks that traditionally make them up (the Generator and the Discriminator). Benefits, limitations and challenges of different approaches of building \acp{GAN} are discussed as well.
    
    \item\textbf{Setup}: Chapter \ref{ch:Setup} discusses the essential programs installed in throughout the research conducted for this report. It also contains the most important Python libraries used in for the training of the \acp{GAN} and generation of images, and the libraries and \ac{API} used to download the images that were to be styled.
    
    \item\textbf{Introductory \ac{GAN}}: Chapter \ref{ch:KerasGAN} discusses and analyzes an introductory \ac{GAN} architecture that is used to generate black and white handwritten digits at 24x24 resolution. 
    The architecture is shown with diagrams to illustrate a basic \ac{GAN} architecture and use it as a base for the models used afterwards.
    It is trained and the results shown at different epochs, to see the learning progression of the model.
    
    \item\textbf{\ac{DCGAN}}: Chapter \ref{ch:CelebaDCGAN} analyzes the architecture of a \ac{DCGAN} that was originally trained on the CelebA\footnote{\url{https://mmlab.ie.cuhk.edu.hk/projects/CelebA.html}} dataset at a resoultion of 64x64 pixels. For this chapter, the model is trained with the Album covers dataset\footnote{\url{https://www.kaggle.com/datasets/greg115/album-covers-images}}.
    The architecture is shown with diagrams to illustrate the differences between the introductory \ac{GAN}.
    Modifications to the normalization of the images and the initialization of the weights are done to the base model, and the results are compared between all the different variations, both the generated images and the loss graphs.
    
    \item\textbf{StyleGAN2}: Chapter \ref{ch:StyleGAN2} analyzes and discusses the STyleGAN2 architecture. All the relevant concepts like \ac{ADA}, the progressive architecture of the model, and the style based generator architecture are covered in depth. The model is trained using the same Album covers dataset\footnotemark[\value{footnote}].The loss and \ac{FID} graphs are shown and the training process discussed. Lastly, both styled and un-styled images are generated using the trained model, and the results discussed.
    
    \item\textbf{Conculsion}: Chapter \ref{ch:Conclusion} discusses the goal and objectives of the research, and to what extent they have been achieved. It also discusses the temporal cost of conducting this research project and future lines of work.
\end{itemize}

%% file: Chapters/01Theoretical.tex
\chapter{Theoretical Framework}\label{ch:theoretical}

\section{Neural Networks}

This chapter will introduce different types \acf{NN}.

\subsection{Classical Neural Networks}\label{classical_NN}

The name \ac{NN} comes from the fact that it is inspired from how biological brains work, they consist of several neurons and connections between them. They are the basic building blocks of neural networks. Any particular neuron holds a value, typically from 1.0 to 0.0. This number is know as the neuron's activation \cite{3blue1brown_but_2017}.

\begin{figure}[bth]
\centering
\includegraphics[width=8cm]{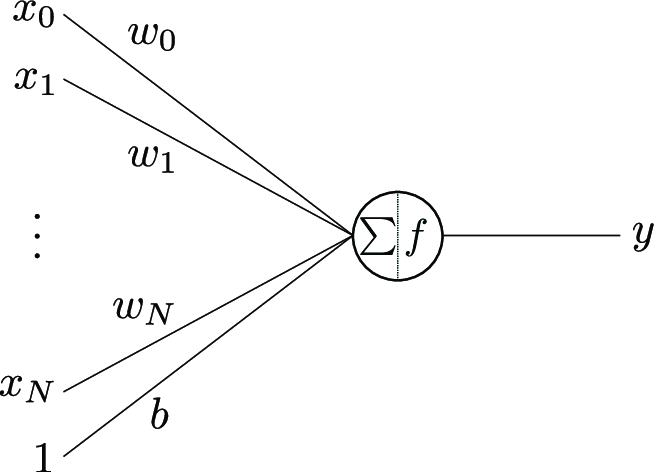}
\caption{Example neuron. Source: \cite{ioannou_structural_2017}.}
\label{fig:Example_Neuron}
\end{figure}

A neuron's activation depends on the weighted sum of the inputs from the other neurons it is connected to with their respective weights, and potentially a bias value. Once the weighted sum is calculated, it is fed through an activation function, that will output the neuron's activation \cite{3blue1brown_but_2017}. This is illustrated in figure \ref{fig:Example_Neuron}.

\begin{figure}[bth]
\centering
\includegraphics[width=10cm]{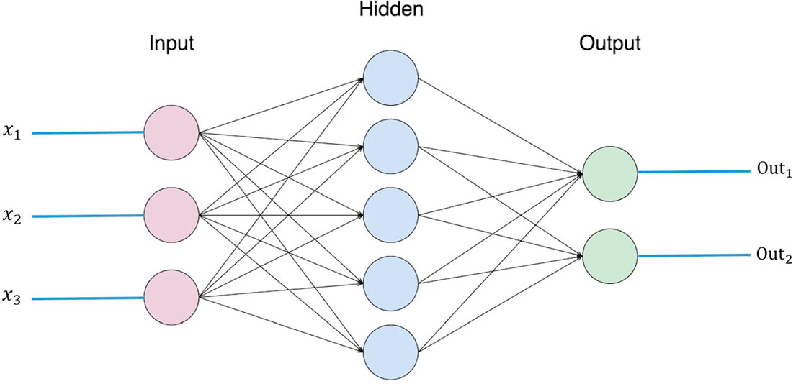}
\caption{Multi-layer neural network. Source: \cite{raissi_parameter_2019}.}
\label{fig:multi_layer_neural_network}
\end{figure}

The neurons form layers, and these layers are connected through the connections with their neurons, as seen in Figure \ref{fig:multi_layer_neural_network}.

Figure \ref{fig:multi_layer_neural_network} shows the 3 types of layers:

\begin{itemize}
    \item\texttt{Input layer}: Takes the input data, and feeds it to the next layer as is.
    
    \item\texttt{Hidden layer(s)}: These layers will perform the computations mentioned previously, and will output to the next layer. These computations finish when the input has been fed through all the hidden layers and arrives at the output layer.
    
    \item\texttt{Output layer}: Where the final computation of the network is located.
    \end{itemize}

Note that the number of layers and neurons in each will depend on the goal that the network wants to achieve. For binary classification tasks, the last layer will most likely be a single neuron, with a value from 0 to 1 depending on what the models classified the input as. The output could also be many neurons, that could represent an image.

\subsection{Activation Functions}
This section will the most common activation functions, and what their function is regarding \ac{NN}.

As mentioned in section \ref{classical_NN}, a neuron's activation is the value contained within it. If a neuron has n inputs $x_1 , x_2 , ... x_n$ then the output or activation of a neuron is $a = g ( w_1 x_1 + w_2 x_2 + w_3 x_3 + ... w_n x_n + b)$ \cite{noauthor_activation_2022}. The activation of all neurons is what will determine the network's state at a given moment. All the functions presented have their strengths and weaknesses, and should be treated as tools when designing a neural network.

\subsubsection{Sigmoid}

\begin{figure}[H]
\centering
\includegraphics[width=.6\linewidth]{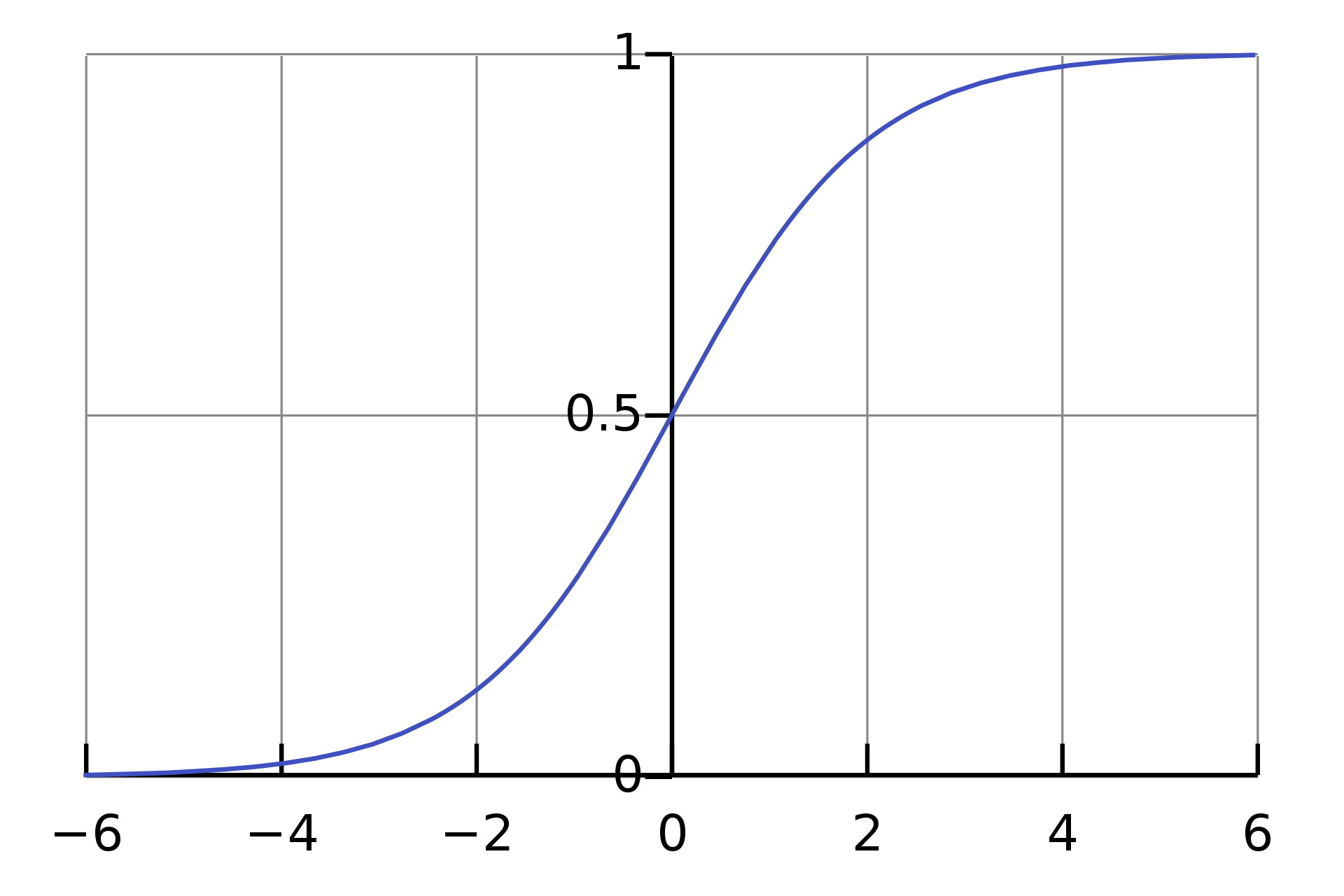}
\caption{Sigmoid function plotted. Source: \cite{noauthor_activation_2022}.}
\label{fig:sigmoid}
\end{figure}

Typically used for the output neuron/s and when the network's task is classification. Notice that the output is $[0,1]$. Sigmoid is useful for final layer when the network should give an answer between 0 and 1, but it can cause the network to suffer vanishing gradients (covered in section \ref{gradients}). Figure \ref{fig:sigmoid} shows the graph of this function and equation \ref{eq:sigmoid_math} is the mathematical representation.

\begin{eqfloat}
\myfloatalign
\begin{equation}
S(x)={\frac {1}{1+e^{-x}}}={\frac {e^{x}}{e^{x}+1}}=1-S(-x)
\label{eq:sigmoid_math}
\end{equation}
\caption{Sigmoid mathematical expression.}
\end{eqfloat}

\subsubsection{Hyperbolic tangent}
The \ac{Tanh} is different than the Sigmoid function in a key aspect, which is its output range which is $[-1,1]$ as opposed to $[0,1]$. \ac{Tanh} is useful to prevent problems like exploding and vanishing gradients since its output range is $[-1,1]$. The benefits and reasoning of when to use on or the other are discussed further in section \ref{improvements_tips}.

\begin{figure}[H]
\centering
\includegraphics[width=.55\linewidth]{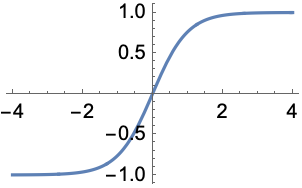}
\caption{\ac{Tanh} function. Source: \cite{noauthor_tanhwolfram_nodate}.}
\label{fig:tanh}
\vspace{-15px}
\end{figure}

\begin{eqfloat}
\myfloatalign
\begin{equation}
\tanh(x)={\frac {e^{x}-e^{-x}}{e^{x}+e^{-x}}}
\label{eq:hyperbolic_tan_expression}
\end{equation}
\caption{\ac{Tanh} expression.}
\end{eqfloat}

\subsubsection{Rectified Linear Activation Function}

The \ac{ReL} function is a linear activation function as opposed to the 2 previous activation functions. It is the most commonly used in neural networks due to it not penalizing positive weights as much as other functions, preventing vanishing gradients (see section \ref{gradients}). Another benefit is its sparsity, since \ac{ReL} will cause some neurons to be 0, when in the same case Sigmoid and \ac{Tanh} will contain values close to 0 but not 0, resulting in a dense representation \cite{daemonmaker_answer_2014}.

\begin{figure}[H]
\centering
\includegraphics[width=.5\linewidth]{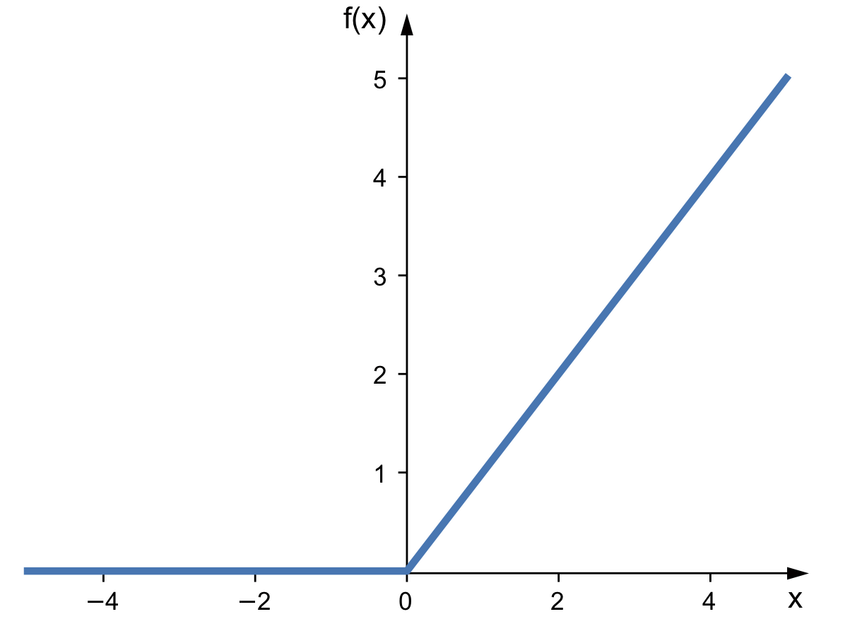}
\caption{\ac{ReL} function. Source: \cite{matougui_k-mer_2019}.}
\label{fig:rel}
\end{figure}

\begin{eqfloat}
\myfloatalign
\begin{equation}
{\begin{aligned}&ReL(x)={\begin{cases}0&{\text{if }}x\leq 0\\x&{\text{if }}x>0\end{cases}}\\{}\end{aligned}}
\label{eq:rel_math}
\end{equation}
\caption{\ac{ReL} mathematical expression.}
\end{eqfloat}

A neuron (or unit) that implements this activation function is referred to as a \ac{ReLU} for short \cite{brownlee_gentle_2019-2}.

\begin{figure}[H]
\centering
\includegraphics[width=.85\linewidth]{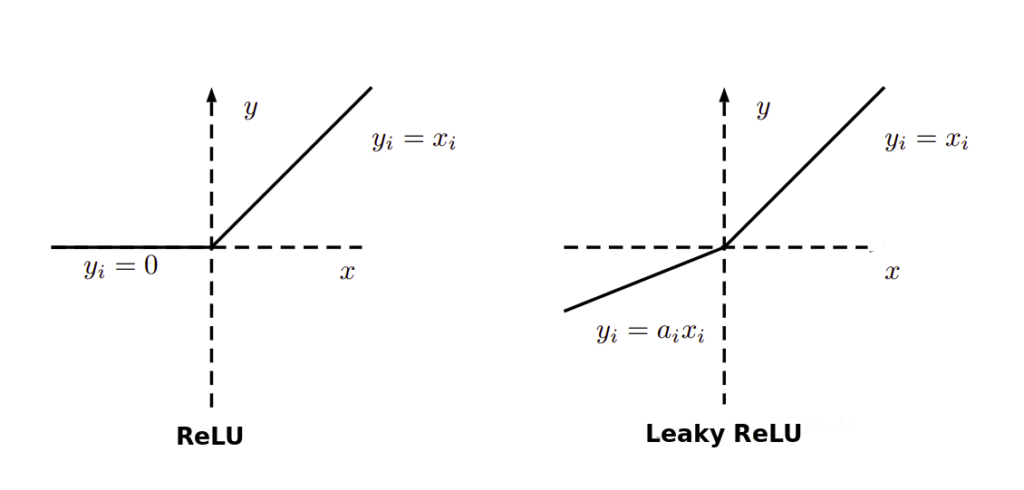}
\caption{Leaky \ac{ReLU} function. Source: \cite{noauthor_papers_nodate}.}
\label{fig:leaky_relu}
\end{figure}

This activation function also has a variant called leaky \ac{ReLU}. For the positive domain it has the same graph, however, for the negative domain, it has a slight slope, where a parameter $a$ is used to specify this slope. Figure \ref{fig:leaky_relu} shows this clearly.

\begin{eqfloat}
\myfloatalign
\begin{equation}
{\begin{aligned}&Leaky ReLU(x)={\begin{cases}ax&{\text{if }}x\leq 0\\x&{\text{if }}x>0\end{cases}}\\{}\end{aligned}}
\label{eq:leaky_relu_math}
\end{equation}
\caption{Leaky \ac{ReLU} mathematical expression.}
\end{eqfloat}

\subsection{How do networks "learn"?}

When the term learning is used, it refers to the adjusting of the weights and biases of the neurons in the network in order to achieve the desired result for given input data. To adjust these, there are several methods \cite{3blue1brown_gradient_2017}. The following explanation will consider the network to be a feed-forward network using \ac{GD} as the optimization algorithm.
    
The main steps in the training process for a network to learn are the following:

\begin{itemize}
    \item Start with some data, and look at the output, which will most likely be totally unrelated to the desired output.
    
    \item Take the given output, and compare it with a desired output. This is done with the use of an error function, where the more different the output is from the desired one, the further from 0 the number.
    
    \item Update the parameters (weights and biases of the neurons) so that feeding data through the network again will yield an output closer to the desired one, therefore minimizing the cost function.
    
    \item Repeat the process until N iterations are performed or the results are satisfying enough given a threshold value.
\end{itemize}

\begin{figure}[!hb]
\centering
\includegraphics[width=8cm]{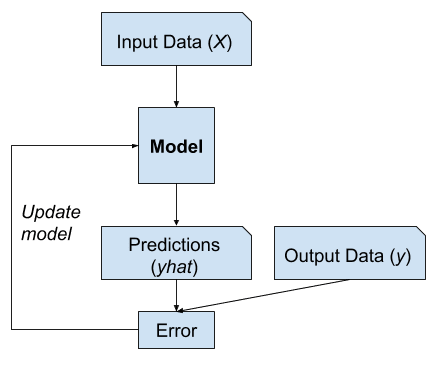}
\caption{Example of a network being trained and the steps performed. Source: \cite{brownlee_gentle_2019}.}
\label{fig:training_NN}
\end{figure}

Figure \ref{fig:training_NN} illustrates the previous points.

This type of learning is called supervised learning.

\subsection{Supervised vs. Unsupervised Learning}\label{supervised_unsupervised}

Supervised learning is when models are trained by updating their weights. It is an active process of learning where the output of the model will be compared against a desired value, and the model will adjust its weights to attempt to map its input data $x$ to output data $y$ that will equal or closely resemble the example data it is trained against \cite{brownlee_supervised_2016}.

Unsupervised learning is when there is no correction to the model. The algorithm is left to its own devices to present underlying structure or distribution in the data \cite{brownlee_supervised_2016}.

A model is constructed by extracting or summarizing the patterns in the input data. There is no correction of the model, as the model is not predicting anything. This is a form of unsupervised learning \cite{brownlee_gentle_2019}.

Generative models could be created using both supervised and unsupervised learning. However, the models that will be looked at in this report will all use forms of supervised learning.

\subsection{Forward pass/propagation}
The forward pass consists in feeding data through the network. Then the output is compared with the desired output with the use of a loss/error function. Cost functions will yield a high number if the network is performing poorly and a number close to 0 if the network is performing well.

A cost function can be thought as a function that takes the biases and weights of the neurons in the network as an input, and outputs a number based on how good they are.

Doing this for the whole training data of the network and averaging the result will give the total cost of the network.

Common loss functions include the \ac{MSE} and \ac{BCE} loss functions.

\begin{eqfloat}
\myfloatalign
\begin{equation}
{\begin{aligned} \sum_{i=1}^{D}(x_i-y_i)^2 \end{aligned}}
\label{eq:MSE}
\end{equation}
\caption{\ac{MSE} mathematical expression.}
\end{eqfloat}

In the \ac{MSE} loss function shown in equation \ref{eq:MSE}, $x_i$ is the current/output data, and $y_i$ is the desired output. It is used when you believe that your target data, conditioned on the input, is normally distributed around a mean value, and when it’s important to penalize outliers more \cite{noauthor_mean_nodate}.

\begin{eqfloat}
\myfloatalign
\begin{equation}
{\begin{aligned} -{(y\log(p) + (1 - y)\log(1 - p))} \end{aligned}}
\label{eq:BCE}
\end{equation}
\caption{\ac{BCE} mathematical expression.}
\end{eqfloat}

In the \ac{BCE} loss function shown in equation \ref{eq:BCE}, $p$ is the current/output data, and $y$ is the desired output (0 or 1) binary indicator. It is useful when the model should output one of 2 labels \cite{noauthor_classical_nodate}.

There is a generalization of the \ac{BCE} loss function when the model has to perform multi-class (more than 2 classes) classification tasks, know as \ac{CCE}. The mathematical expression is shown in equation \ref{eq:CCE}.

\begin{eqfloat}
\myfloatalign
\begin{equation}
{\begin{aligned} -\sum_{c=1}^My_{o,c}\log(p_{o,c}) \end{aligned}}
\label{eq:CCE}
\end{equation}
\caption{\ac{CCE} mathematical expression.}
\end{eqfloat}

In the \ac{CCE} loss function shown in equation \ref{eq:CCE}, $M$ is the number of classes, $p$ is the current/output data and $y$ is the desired output (0 or 1) binary indicator \cite{noauthor_classical_nodate}.

\subsection{Backward pass/propagation}
Back propagation is the act of modifying the weights and biases based on the output of the cost function.

When the cost is computed, the output is compared with the desired one. From this the difference between the goal and the result can be measured. This will be in the form of a change in the activation of the neurons in the last layer. However, the only parameters that can be changed are the weights and biases of the network \cite{3blue1brown_gradient_2017}.

This means that what will need to be computed is what change in the weights and biases of the previous layer will bring the results in the current layer closer to what is desired, and so on for all the layers in the network, moving backwards. Once this process is performed, we will have a list of all the changes that should be performed to the weights and biases in order to move the network closer to the result. This is \ac{GD}, since the rate of steepest decline to the cost function is computed, and the inputs of the function changed to achieve this result.

This could also be performed by obtaining the derivative of the cost function at a certain point (these points being the weights and biases of the network), however this is not usually feasible, since the function will be immensely complicated given the amount of input variables.

This is done for all training examples and the list with the changes that should be performed to the weights and biases is averaged for all examples, yielding the direction of best overall "improvement" for the model.

This is very computationally intensive however, so what is typically done is to shuffle the input data and form batches with size considerable smaller than the input data size, calculate the gradient for these, and update the weights and biases after every batch instead of after the whole data set. This will of course not be the exact step in the direction of most improvement, but it is a step and significantly less computationally intensive. The technique of using batching for performing \ac{GD} is referred to as \ac{MBSGD}. If the program were to use only 1 input per pass, it would be \ac{SGD}, which is fastest to compute but the direction of each step will vary greatly.

\begin{figure}[bth]
\centering
\includegraphics[width=0.9\linewidth]{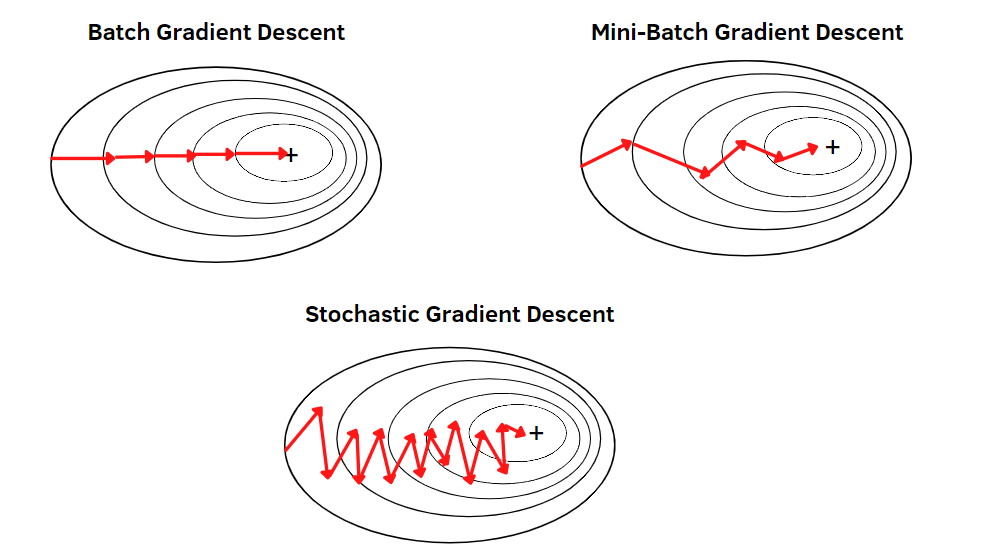}
\caption{Different types of gradient descent example. Source: \cite{laptrinhx_understanding_2021}.}
\label{fig:gradient_descent_examples}
\end{figure}

\pagebreak

\subsection{Convolutional Neural Networks}

\ac{CNN} use the same structure explained in the previous section, but they use convolutions and pooling.
    
\begin{itemize}
\item A convolution involves taking the input from one layer and passing it through a filter, which will output the convolved image. The goal of these filters is to pick up on patterns. Stride, padding and size will be talked about more in detail in later sections \cite{noauthor_convolutional_2022}.

\begin{figure}[bth]
\centering
\includegraphics[width=0.8\linewidth]{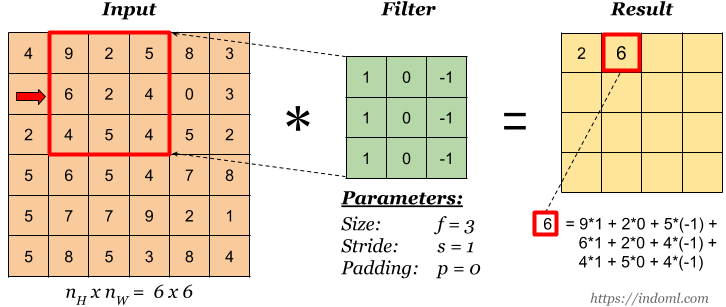}
\caption{Convolutions example. Source: \cite{noauthor_student_2018}.}
\label{fig:convolutions}
\end{figure}

\item A pooling layer has the goal of condensing the extracted features by either taking the maximum value of a certain region or the average value.

\begin{figure}[bth]
\centering
\includegraphics[width=0.62\linewidth]{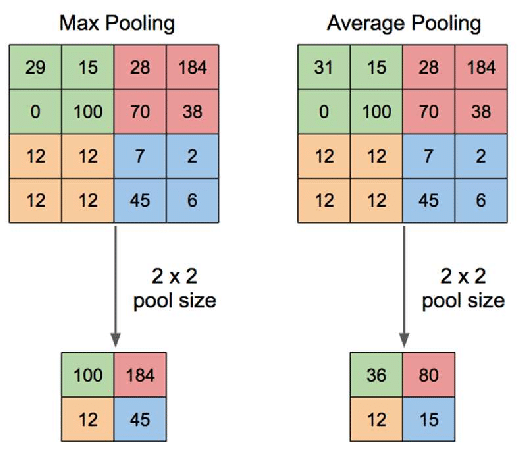}
\caption{Max pooling and average pooling illustrated. Source: \cite{yani_application_2019}.}
\label{fig:pooling}
\vspace{-20px}
\end{figure}

\end{itemize}

The filters are what is modified during the training process using the same mechanisms outlined before, forward and backward propagation. The neurons in the hidden layers of this type of networks typically have the same filters, in order to pick up on the same pattern no matter where it is located in the input. These types of networks are well suited to capture features of images \cite{mandal_cnn_2021}.

\section{Generative models}

\subsection{Discriminative vs. Generative Modeling}

When creating and training a model, the goal could be to either predict a class label, or to generate or create new examples in the input distribution. These models are classed as discriminative and generative models respectively.

A good generative model will be able to generate new data that is not just plausible but indistinguishable from real data.

\subsection{Generative Adversarial Networks}

\acp{GAN} are a type of generative network architecture. Generally they are built with 2 networks, a generator and a discriminator.

\begin{itemize}
\item\texttt{\ac{G}}: Generates new plausible examples from the input.

\item\texttt{\ac{D}}: Classifies the generated examples as real (from the input) or fake (generated).
\end{itemize}

The generator will draw from an input distribution, and map this distribution to an output through training. Generally, this is in the form of a vector containing randomly sampled values from a normal distribution of mean $0$ and standard deviation $1$. This vector is also known as the latent space vector \cite{brownlee_gentle_2019}.

\subsection{Why Adversarial?}

The term adversarial is used since the 2 networks are competing against each other. The generator is trying to fool the discriminator, and the discriminator is trying to become better at recognizing fakes.

\begin{displayquote}
\textquote{We can think of the generator as being like a counterfeiter, trying to make fake money, and the discriminator as being like police, trying to allow legitimate money and catch counterfeit money. To succeed in this game, the counterfeiter must learn to make money that is indistinguishable from genuine money, and the generator network must learn to create samples that are drawn from the same distribution as the training data} \cite{brownlee_gentle_2019}.

This is called a zero-sum-game, since if one side performs well, it is rewarded, or no change is performed to model parameters, while the other side is punished, or no change is performed to model parameters. There is no \textquote{winning} for both sides.

\textquote{At a limit, the generator generates perfect replicas from the input domain every time, and the discriminator cannot tell the difference and predicts “unsure” (e.g. 50\% for real and fake) in every case. This is just an example of an idealized case} \cite{brownlee_gentle_2019}.
\end {displayquote}

In practice, it is not needed to arrive at this point to create a useful generator model.

\begin{figure}[!hb]
\centering
\includegraphics[width=0.65\linewidth]{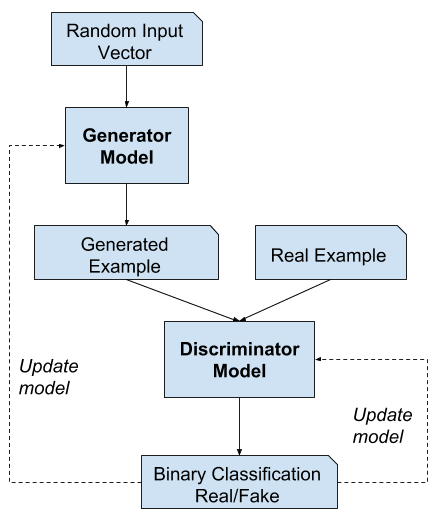}
\caption{GAN architecture illustrated. Source: \cite{brownlee_gentle_2019}.}
\label{fig:GAN Architecture}
\end{figure}

\subsection{Deep Convolutional GAN}

\acp{DCGAN} are \acp{GAN} that explicitly use convolutional layers in the discriminator and convolutional-transpose layers in the generator. This type of architecture laid the ground-work for this project, as it was the first one used to learn and experiment with \acp{GAN} themselves.

\section{Challenges and tips for training GANs}\label{challenges_and_tips}

\acp{GAN} bring their own set of issues to the already existing challenges with training \ac{ML} models. The reason more challenges are faced is because 2 models are being trained simultaneously instead of a single one. All of these common problems are areas of active research, and the solutions to them sometimes depend on the context. However, there are some best practices that are generally accepted by the research community \cite{brownlee_tips_2019}.

\subsection{Failure modes of GANs}\label{failure_modes}

The following section discusses the challenges that are faced when training \acp{GAN}.

\subsubsection{Mode Collapse}
Mode collapse is when the generator is only able to produce an output of a single or very limited subset of the training data. In other words, the mapping from latent space to output space is very limited. Mode collapse will manifest in the generator producing the very similar or the same image for different input vectors \cite{noauthor_gan_nodate}.

\begin{figure}[bth]
\centering
\includegraphics[width=0.92\linewidth]{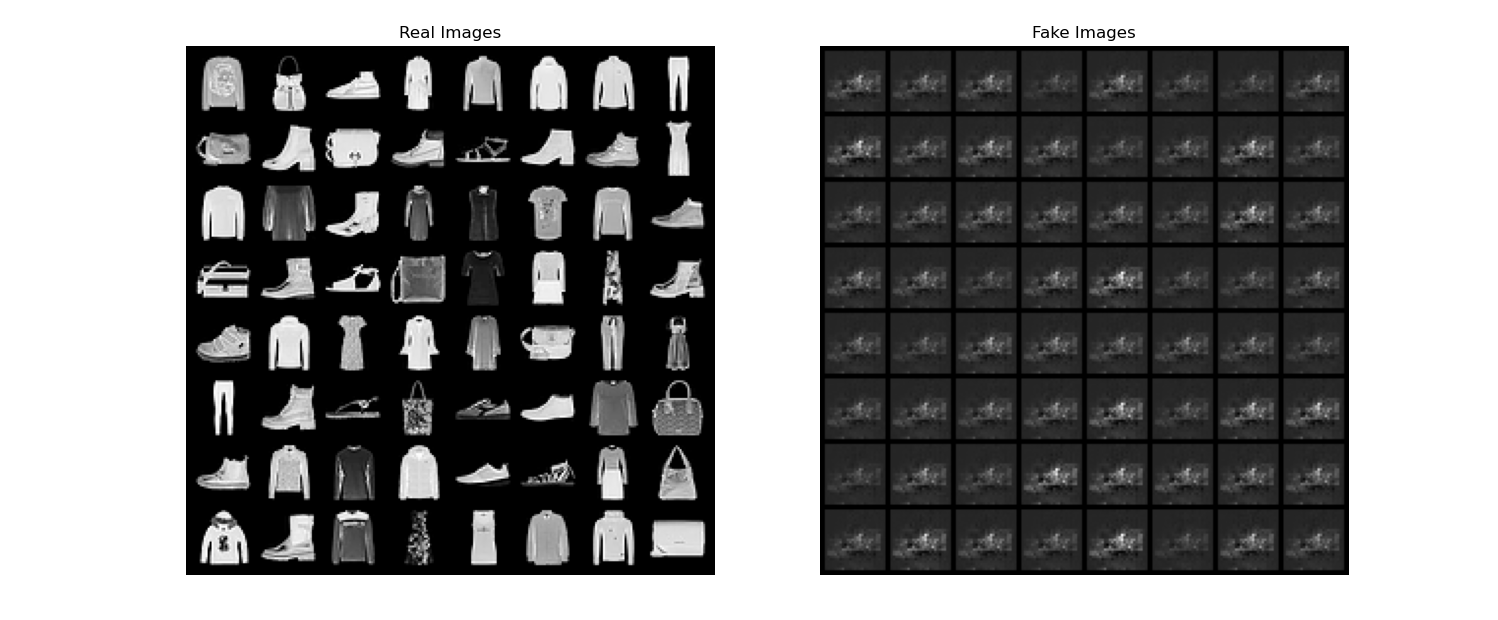}
\caption{Mode collapse encountered when testing different hyper-parameters.}
\label{fig:Mode Collapse}
\end{figure}

\subsubsection{Vanishing and exploding gradients}\label{gradients}
The vanishing gradients phenomenon occurs when during the weight adjustment (back-propagation) step, the weights get close or equal to zero, leaving the initial layers' weights almost or completely unchanged. Exploding gradients conversely means that during this step, the weights become very large, affecting the initial layers greatly and causing the \ac{GD} to diverge \cite{noauthor_vanishing_2021}.

\subsubsection{Failure to Converge} \label{failure_to_converge}
Due to the nature of \acp{GAN}, there will be a point where the generator has become good enough to fool the discriminator (barring any other failure mode), making the discriminator effectively guess if the picture generated is fake or not (50\% discriminator accuracy). This becomes a problem since the generator will be trained on random feedback, and its results might be hindered \cite{noauthor_gan_nodate}.

Model convergence can be detected when reviewing the learning curves of the \ac{GAN}.

\begin{figure}[bth]
\centering
\includegraphics[width=0.75\linewidth]{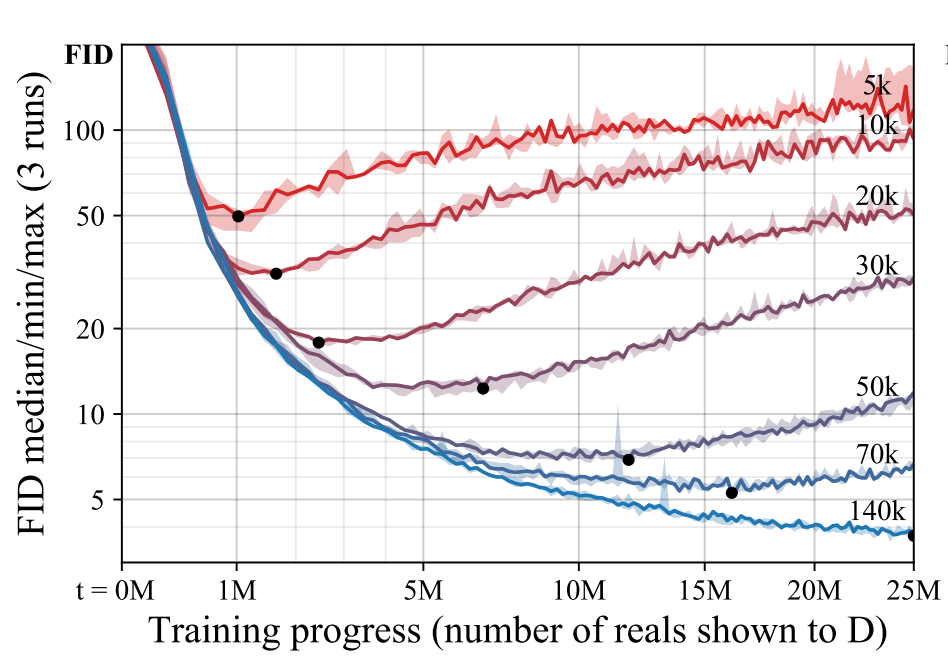}
\caption{Convergence points, the black dots show the points where the generator performance starts to deteriorate. Source: \cite{karras_training_2020}.}
\label{fig:Convergence}
\end{figure}

\pagebreak

\subsection{Improvements and training tips}\label{improvements_tips}

The following section discusses how to overcome or compensate for the shortcomings discussed in the previous section \ref{failure_modes}.

\subsubsection{Initializing weights}
This refers to the initialization of each neuron's activation. Initializing them avoids the gradients vanishing or exploding gradients problem \cite{dellinger_weight_2019}. For a \ac{DCGAN}, the most effective way is to sample randomly from a zero-centered Gaussian distribution with a standard deviation of 0.02 \cite{brownlee_how_2019-1}.

\subsubsection{Scaling pixel values}
This refers to both scaling the input images and using an activation function at the end of the generator that also outputs a scaled value image. Good practice is to use the \ac{Tanh} activation function at the output of the generator, whose output range is $[-1, 1]$. The range of values for the scaled images should be the same for the output of the generator and the scaled data.

The main idea is that if the weights are only positive like in the case of the Sigmoid activation function, the wights that feed into a node of the next layer will all be positive, making it harder for the weight to change direction since all of the weights are "pulling" it toward the positive side \cite{ekoulier_answer_2018}. Furthermore, not only are the input values always in the same range but so are the output values, hence the discriminator will be fed data with the same format regardless if it is fake or real, stabilizing the training.

\subsubsection{Batch normalization}
\ac{BN} standardizes the output of a particular layer to have a normal distribution with mean $0$ and \ac{STD} of $1$. This is done to regularize the training, avoids overfitting and speeds up training\cite{brownlee_gentle_2019-1}.

However, the input of the discriminator and the output of the generator, have shown to decrease network performance and stability when \ac{BN} is used \cite{brownlee_tips_2019}.

\subsubsection{Dropout}
Dropout reduces over-fitting by probabilistically ignoring some neuron inputs\cite{brownlee_how_2018}. This has the effect of modifying the other neurons in the layer to take on more responsibility, and achieves a sparse representation of each layer, which in turn makes the network learn a sparse representation of the input data, which reduces over-fitting\cite{srivastava_dropout_2014}.

It is important to note that this method is not necessary, and sometimes detrimental when used together with batch normalization \cite{brownlee_gentle_2019-1}.

\subsubsection{Noisy labels and label smoothing}
Implementing noisy labels involves inserting some fake images to the real batch of input data and vice-versa for the fake batch, usually with a ratio of 5\% inserted data and 95\% original data\cite{brownlee_gentle_2019-1}. Smoothing labels mean that images are not judged or taken as either fully real or fully fake, but rather it is a spectrum. A label could have a value from 0.7 and 1.2 if it real for example \cite{chintala_how_2022}.

\subsection{How to evaluate GAN results}

There are 2 common quantitative methods of evaluating \ac{GAN} performance.

\begin{itemize}
    \item\texttt{\ac{FID}}: Estimates the quality of the generated images by \textquote{evaluating confidence of the conditional class predictions for each synthetic image (quality) and the integral of the marginal probability of the predicted classes (diversity)}\cite{brownlee_how_2019-2}. This uses a pre-trained model to extract the features. As of writing this report, the current one used is the \href{https://en.wikipedia.org/wiki/Inceptionv3}{inception v3 model}.
    
    \item\texttt{\ac{IS}}: Also uses the same pre-trained model to make the predictions. It measures variety and individual quality.
\end{itemize}

Another method is to look at the generator loss vs the discriminator loss. Ideally, the generator loss starts at a high value and becomes lower, while the discriminator loss remains constantly at a low number. This means the generator has learnt from the training and can feed the discriminator better fakes, but not enough so that the discriminator is totally fooled. As mentioned in \ref{failure_to_converge}, when the generator becomes good at producing fakes while the discriminator hasn't kept up, the generator will not receive any extra information and start being trained on random guesses.

The problem with these metrics is that they do not consider spatial relationships, which is a major factor in evaluating the model's performance. They need to be used together with qualitative evaluation, since the objectives of these models are usually to produce data that humans would perceive as real, such as human faces or cats.

%% file: Chapters/02Setup.tex
\chapter{Setup and Programs used}\label{ch:Setup}

The following chapter introduces all the programs and libraries used to set up the repeatable environment for running and training the \ac{GAN} models chosen, along with other tasks necessary such as downloading and processing images and plotting results.

\section{Windows 10 + Ubuntu 20.04.4 with 5.10 LTS kernel}
Since the main development computer was running the Windows 10 \ac{OS}, it was more convenient to leverage the benefits that \ac{WSL} offered, by enabling a Linux distribution to be ran on top of Windows 10 (almost) seamlessly. The goal was to use one underlying OS to avoid dual booting and the time impact it has, and also avoid GPU pass through problems with \acp{VM}, along with the performance impact. Linux was preferred over windows for development due simply to familiarity and comfort, since all the libraries and programs could be installed on Windows 10 without a problem. In fact, Docker desktop was used since it was easier to manage containers and images and see the error logs with a \ac{GUI}.

Anaconda for windows was originally used but \ac{WSL} proved to be faster to develop on and Anaconda was slower to start thus added friction to the development process.

Other development computers were used to train models, which were running \ac{LTS} versions of Ubuntu 20.04 \ac{LTS}, which were accessed remotely to train models for long periods of time, and because their more powerful \acp{GPU}.

\section{Docker}
Docker was used to build development ready environments, such as the one in StyleGAN2\footnote{\url{https://github.com/NVlabs/stylegan2-ada-pytorch}} GitHub repository. This meant that if a model needed a particular version of a program or library, there was no need to install it on the development computer. Instead, downloading or building the image with the provided Dockerfile would provide the environment necessary for the training of the models and other research tasks necessary like plotting results.

\section{Kaggle}
\href{https://www.kaggle.com/}{Kaggle} is a website that is host to thousands of datasets, such as famous ones like the \ac{MNIST} and the CelebA datasets. It also offers users a place to host self made datasets and Jupyter notebooks. It was used to download the Album covers dataset\footnote{\url{https://www.kaggle.com/datasets/greg115/album-covers-images}}.

\section{Spotify API}
The Spotify \ac{API}\footnote{\url{https://developer.spotify.com/documentation/web-api/}} was used in order to obtain album cover images from a given playlist. This made it easy to download the images that were to be styled, since an existing playlist with songs of albums that wanted to be used could be specified and the album covers downloaded. The possibility of creating custom playlists also exists, to download custom album cover images to be styled.

\section{Python}

Python was the development language for anything involved with the models themselves. It was partly used in data acquisition as well, together with \ac{JS}.

Python version: 3.8.10

The following are the main libraries used for creating and training the models and for visualization.

\subsection{PyTorch}
\textquote{PyTorch is an optimized tensor library for deep learning using \acp{GPU} and \acp{CPU}.} \cite{noauthor_pytorch_nodate}. It is primarily developed by \ac{FAIR} lab and is open-source \cite{noauthor_pytorch_2022}. This library was used for training and modifying the \ac{DCGAN}, and because it facilitates use of \ac{GPU} acceleration to speed up training. It is also the library used for the implementation of StyleGAN2 used.

\subsection{TensorFlow + Keras}
Keras is a library that provides high level wrappers for \ac{TF} functions in Python, which is an open-sourced end-to-end platform, a library for multiple machine learning tasks \cite{noauthor_keras_2020}. The introductory \ac{GAN} used is implemented in Keras. StyleGAN2 has an implementation made with Keras but the PyTorch implementation results in faster training in most cases \cite{noauthor_nvlabsstylegan2-ada-pytorch_2022}.

Keras version: 2.6.0
\ac{TF} version: 2.8.0

\subsection{TensorBoard}
TensorBoard is a visualization toolkit for TensorFlow, however it can be used for model performance tracking and visualization for models made with other libraries such as PyTorch. In fact, \href{https://github.com/NVlabs/stylegan2-ada-pytorch}{StyleGAN2}'s authors used it to generate statistics and the model is implemented in PyTorch.

TensorBoard version: 2.8.0

\subsection{NumPy}
NumPy is a Python library that provides a multidimensional array object, various derived objects (such as masked arrays and matrices), and an assortment of routines for fast operations on arrays, including mathematical, logical, shape manipulation, sorting, selecting, basic linear algebra, basic statistical operations, random simulation and much more \cite{noauthor_numpy_nodate}.

NumPy version: 1.22.3

\subsection{Spotipy}
Spotipy is a wrapper library to facilitate making \ac{API} calls to the Spotify endpoints. It provides helper objects and methods that were used to streamline the process of obtaining album covers from Spotify playlists.

\section{Jupyter Notebook}
Jupyter notebook aids development of python applications by allowing the code to be split in cells and executed individually. This is especially useful during development of \ac{AI} models since there are some parts of the code that will need to be run individually several times, such as loading the data to test out different transforms.

Since the whole code is not run every single time, one can focus development on different blocks of the overall project better. It also allows for markdown to be added as cells between code cells, to provide a more detailed and potentially visual explanation to each cell of code.

Jupyter notebook version: 6.4.11

\section{CUDA}
CUDA is a parallel computing platform and programming model developed by NVIDIA for general computing on NVIDIA \acp{GPU} \cite{noauthor_cuda_2017}. \ac{AI} applications benefit from using the \ac{GPU} for several reasons, but mainly it is because \acp{GPU} can perform many simultaneous operations on their thousands of cores, parallelizing the work many more times than a \ac{CPU} with a handful of cores \cite{noauthor_gpu_nodate}.

Another benefit is memory bandwidth, speed and size. If the dataset is very large and can fit on GPU memory, since this memory is close to the \ac{GPU} cores and is faster and of higher bandwidth than system memory, the retrieval process of data is much faster.

Since all the computers used for development had NVIDIA \acp{GPU}, it was also the most versatile option.

CUDA version: 11.6

%% file: Chapters/IntroGAN.tex
\chapter{Introductory GAN}\label{ch:KerasGAN}

The goal of analyzing and running the tutorial code mentioned is to see the results it yields, check the performance of the model, both in how much time it took to train and the results, and understand the architecture of this \ac{GAN} and why this architecture works for the dataset used. The results look at different training intervals and how the models has progressed with the increasing number of epochs, and potential improvements that could be made.

\section{Architecture}

\begin{figure}[bth]
        \myfloatalign
        {\includegraphics[width=1\linewidth]{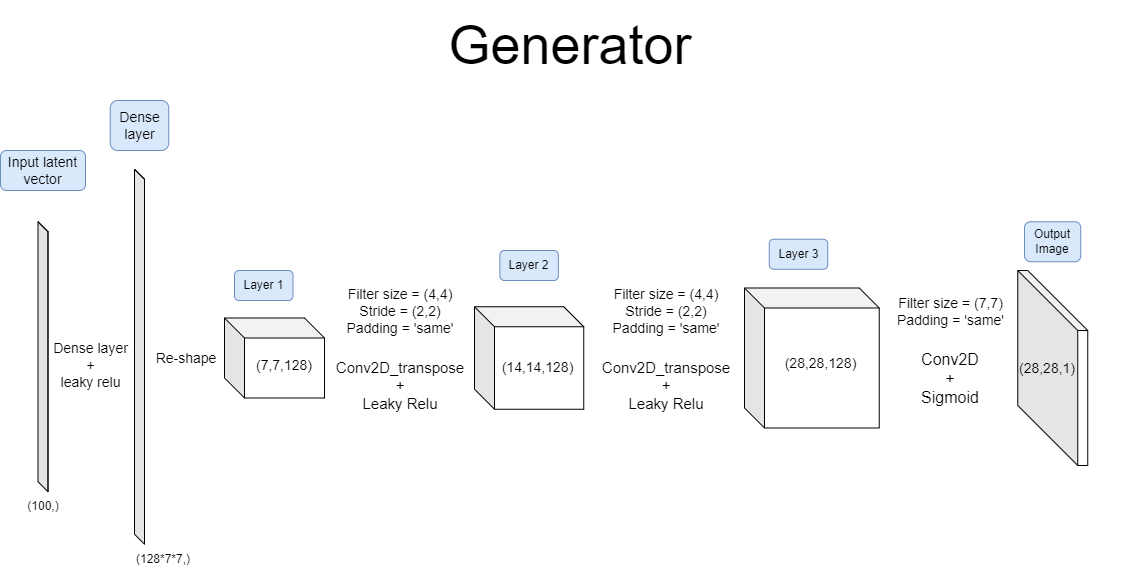}}
        \caption{Introductory \ac{GAN} architecture of the Generator.}
        \label{fig:keras_arch_g}
\end{figure}

\begin{figure}[bth]
        \myfloatalign
        {\includegraphics[width=0.85\linewidth]{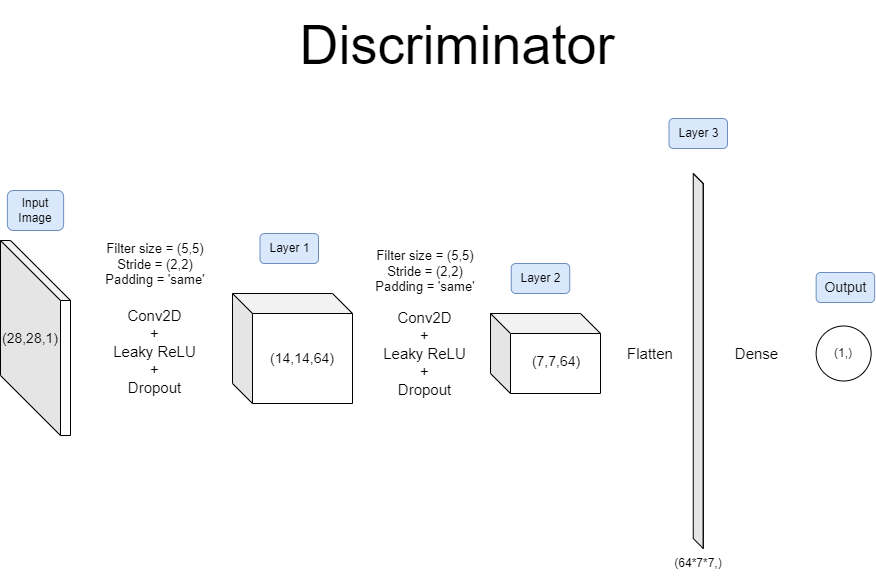}}
        \caption{Introductory \ac{GAN} architecture of the Discriminator.}
        \label{fig:keras_arch_d}
\end{figure}

The following figures show both the generator and the discriminator, which will then in turn be combined as per figure \ref{fig:GAN Architecture}. Figure \ref{fig:keras_arch_g} shows the generator and figure \ref{fig:keras_arch_d} the discriminator architecture.

\pagebreak

\section{Dataset}

\begin{figure}[bth]
        \myfloatalign
        \subfloat[The first 25 digits of the \ac{MNIST} dataset.]
        {\includegraphics[width=.43\linewidth]{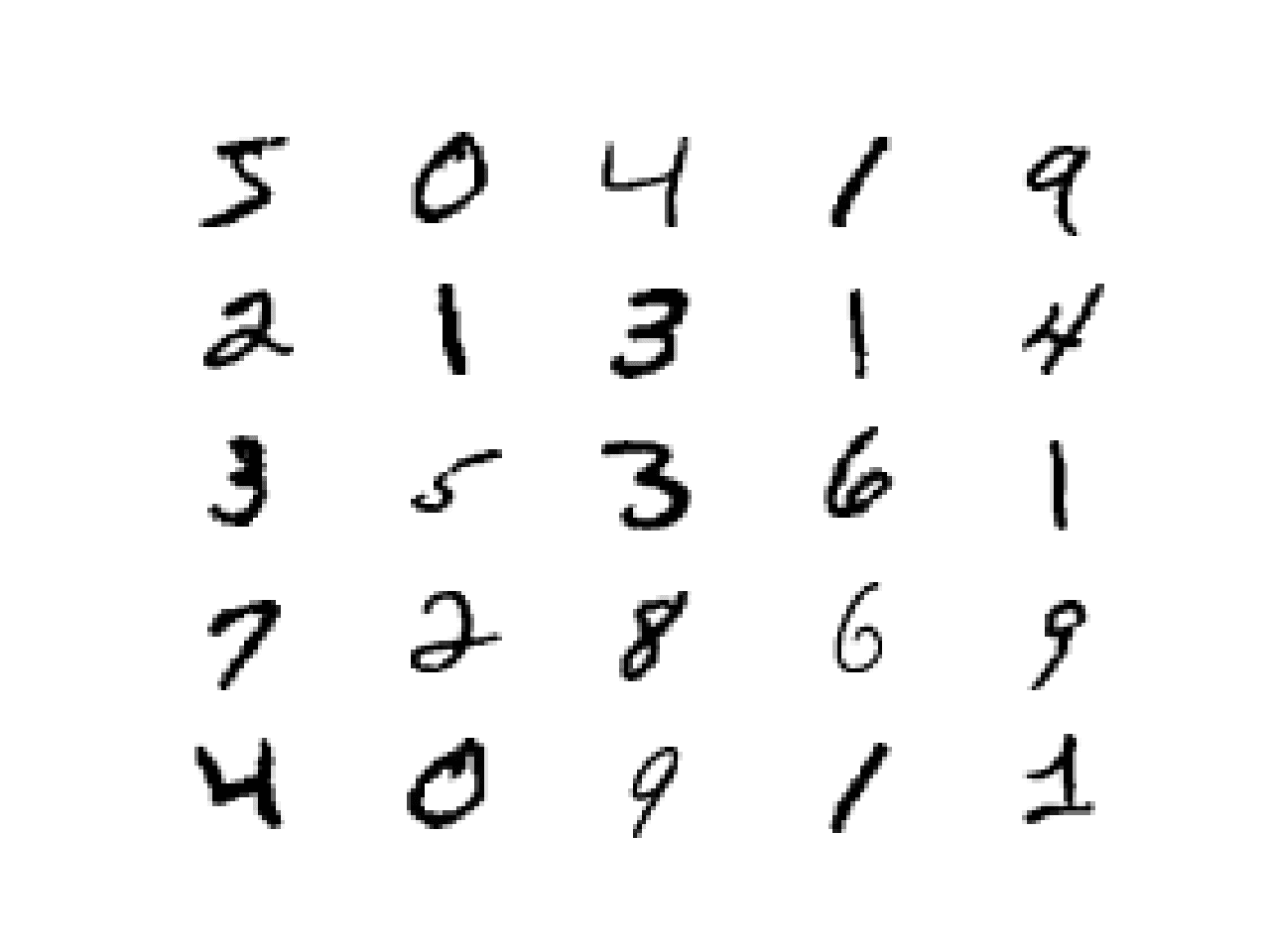}} \quad
        \subfloat[Example of ambiguous digits in the dataset.]
        {\includegraphics[width=.43\linewidth]{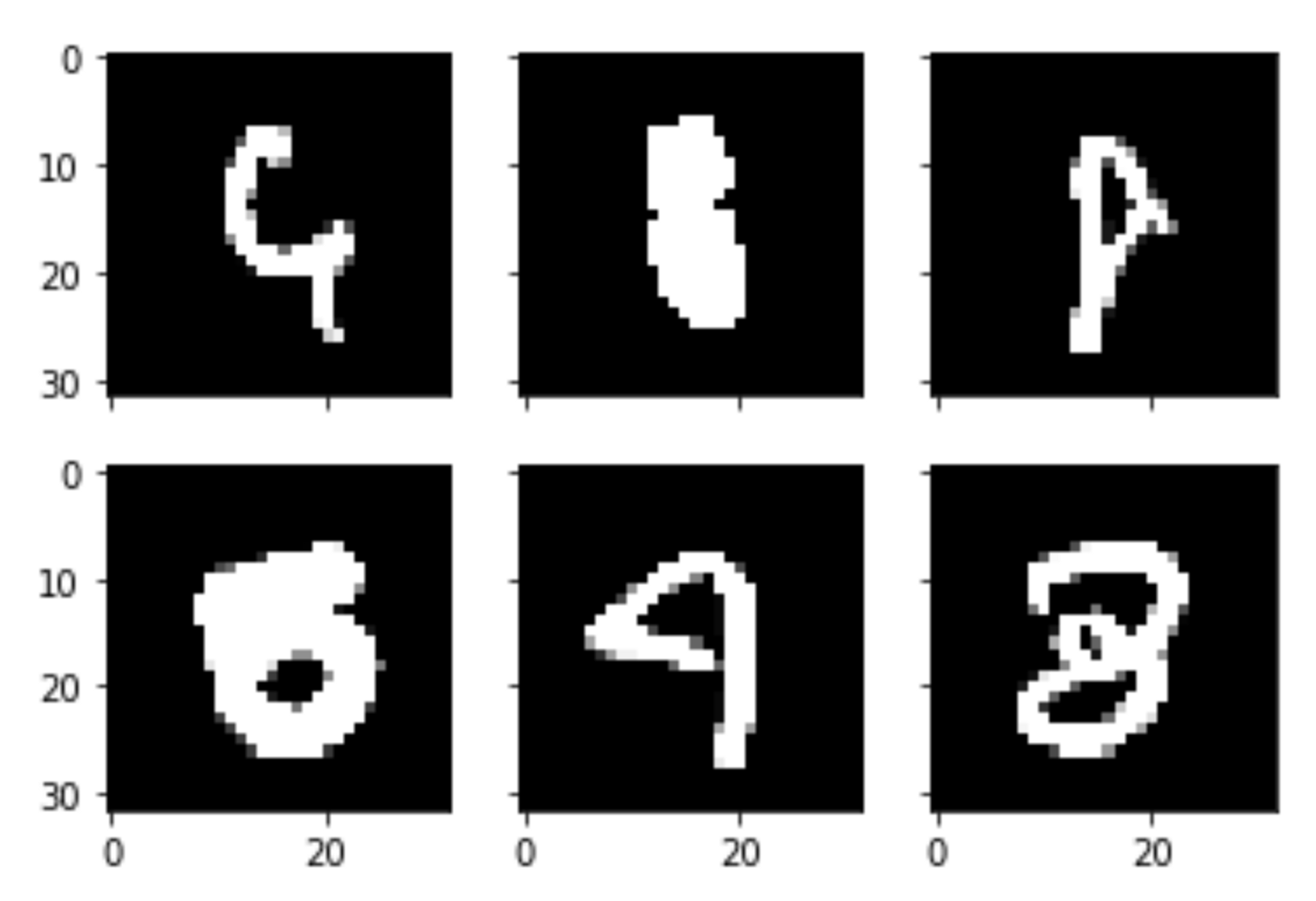}} \\
        \caption{Examples from the \ac{MNIST} dataset.}
        \label{fig:mnist_digits_example}
\end{figure}

The \ac{MNIST} dataset was used for this first \ac{GAN}. It is composed of 60k images of 28x28 grey-scale handwritten digits. The results and the performance of the \ac{GAN} should therefore be easy to evaluate qualitatively by humans since handwritten digits will are easily recognizable. The also dataset contains some digits that could be understood as 2 different numbers, since not all digits humans write are perfect. Some examples of digits are shown in figure \ref{fig:mnist_digits_example}.

\pagebreak

\section{Training}

The training process for this \ac{GAN} is as follows; For every epoch it will iterate through $Total\_images / Batch size$, since the model is trained with mini-batch \ac{GD}. The values are shown in table \ref{tab:hyperparameters_keras} Every batch will be split in half, 1 half real images and 1 half fake images, and they will be used to train (forward pass + backward pass) the discriminator. Once trained, in the same epoch, the generator will generate (forward pass) $Batch size$ number of fake images, and the weights will be adjusted based on the discriminators error (backward pass).

The hyperparameters used for the model training can be seen in table \ref{tab:hyperparameters_keras}. They are unchanged from the original code, since the idea is to learn about \ac{GAN} architectures, and tuning will be performed on the following \acp{GAN}. \ac{BCE} loss is used since the discriminator will tell between 2 possible options (real or fake) how close it is to one or the other i.e. how fake or real the image it receives is.

\begin{table}[bth]
    \myfloatalign
  \begin{tabularx}{\textwidth}{Xll} \toprule
    \tableheadline{Parameter} & \tableheadline{Value} \\ \midrule
    Total images & $60000$ \\
    Batch size & $256$ \\
    Image dimensions & $28x28x1$ \\
    Learning rate & $0.0002$ \\
    Adam's beta & $0.5$ \\
    Latent vector elements & $100$ \\
    Training epochs & $100$ \\
    Leaky \ac{ReLU} slope & $0.2$ \\
    Weight initializer & glorot\_uniform (default) \\
    Pixel value scale of training images & NO \\
    Batch normalization & NO \\
    One-side label smoothing & NO \\
    Noisy Labels & NO \\
    Dropout & $0.4$ (Discriminator) \\
    Progressive growing & NO \\
    Loss & Binary cross entropy \\
    \bottomrule
  \end{tabularx}
  \caption{Hyperparameters used to train the introductory model.}
  \label{tab:hyperparameters_keras}
  \vspace{-15px}
\end{table}

\section{Results}
During the training, the code outputs the images generated at intervals of 10 epochs.

The generated images at 10 epochs resemble handwritten digits but have very thin lines in some cases that break up the number, and imperfections outside the number itself. The overall shape of the digits is correct however.

\begin{figure}[bth]
        \myfloatalign
        \subfloat[Results after 10 epochs.]
        {\includegraphics[width=.61\linewidth]{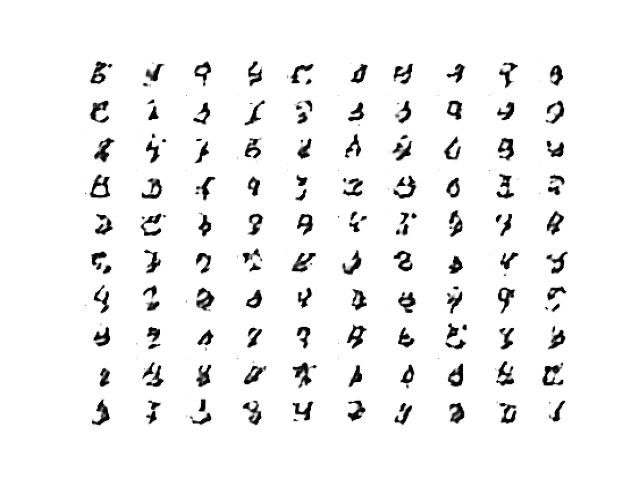}} \quad
        \subfloat[Results after 30 epochs.]
        {\includegraphics[width=.61\linewidth]{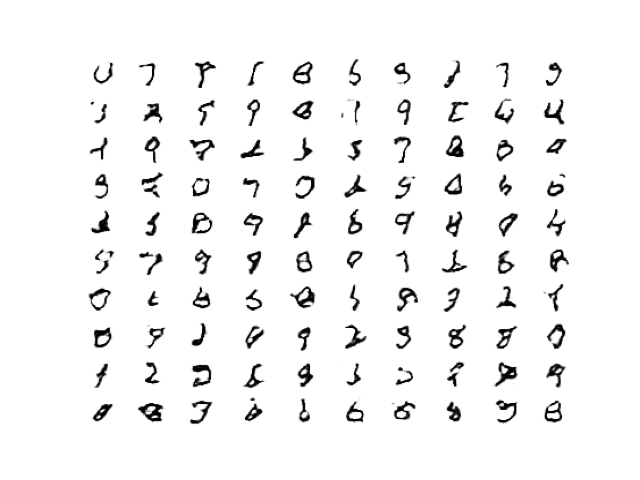}}
        \caption{Training results at 10 and 30 epochs of the introductory \ac{GAN}.}
        \label{fig:keras_10_30_e}
\end{figure}

Looking at results from 30 epochs of training, they are much cleaner, without as many of the imperfections. Both results can be compared in figure \ref{fig:keras_10_30_e}.

\begin{figure}[bth]
        \myfloatalign
        {\includegraphics[width=1\linewidth]{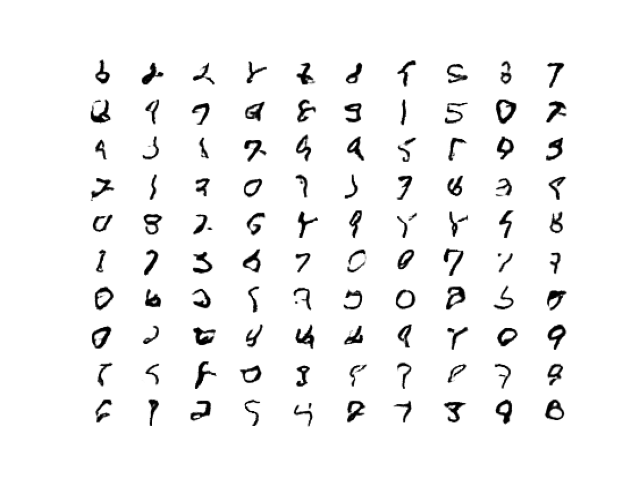}}
        \caption{Training results at 100 epochs of the introductory \ac{GAN}.}
        \label{fig:keras_100_e}
\end{figure}

Finally, at 100 epochs (figure \ref{fig:keras_100_e}), most numbers have all the features of the original dataset. It seems that the model's weak point is generating images where these features are cohesive, meaning that the shape of the overall number is correct but the strokes are not properly joined together.

\section{Conclusion}

There are some improvements that can be made to this network. For example, Sigmoid function is used for both the discriminator and the generator. If \ac{Tanh} was used as the activation function for the generator, the training would be more stable. Other improvements would be to use batch normalization, along with the other improvements mentioned in \ref{improvements_tips}, which have not been implemented in this model.

With that, these improvements were applied, along other techniques, to the following \ac{DCGAN}.

%% file: Chapters/CelebrityFacesDCGAN.tex
\chapter{DCGAN}\label{ch:CelebaDCGAN}

For this chapter, the \ac{DCGAN} from the PyTorch tutorial pages\footnote{https://pytorch.org/tutorials/beginner/dcgan\_faces\_tutorial.html} will be analyzed, modified and evaluated for the different setups. This \ac{GAN} was selected since it is composed of more layers that make it capable of generating human faces, at a higher resolution and with more channels than the previous \ac{GAN}.

The main benefit is that it is more robust for picking up on the features that make up a face than the previous \ac{GAN}. The hypothesis then became that adding more layers would make any \ac{GAN} pick up on all features of the input data no matter how complex, but this of course was not the case.

The following is the process followed to use this \ac{GAN}, what was done in attempts to make it generate better album covers and a look at its underlying architecture.

\section{Architecture}

\begin{figure}[H]
        \myfloatalign
        {
        \includegraphics[width=1\linewidth]{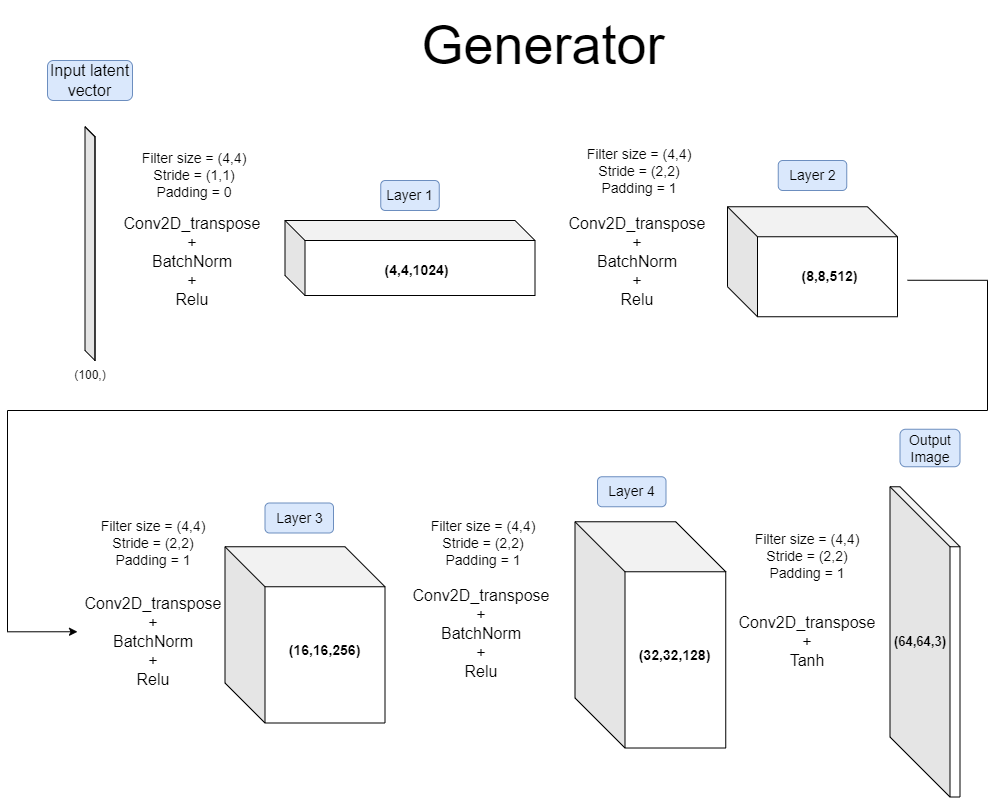}}
        \caption{\ac{DCGAN} Architecture of the Generator.}
        \label{fig:dcgan_g}
\end{figure}

\begin{figure}[H]
        \myfloatalign
        {
        \includegraphics[width=0.9\linewidth]{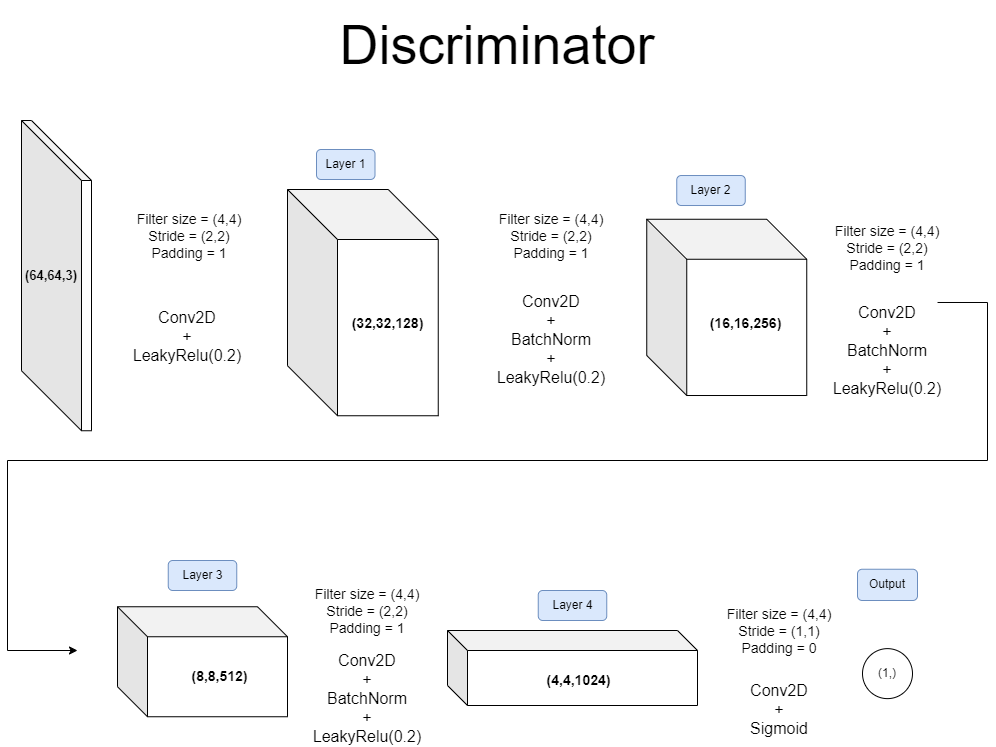}} 
        \caption{\ac{DCGAN} Architecture of the Discriminator.}
        \label{fig:dcgan_d}
\end{figure}

The following figures show both the generator and the discriminator, which will then in turn be combined as per figure \ref{fig:GAN Architecture}. Figure \ref{fig:dcgan_g} shows the generator and figure \ref{fig:dcgan_d} shows the discriminator.

Note that for this \ac{GAN} there are more layers than the previous one, batch normalization is used and the shapes of the layers are different. Not only do the layers have different sizes versus the previous model, but the hidden layer dimensions are symmetrical.

Furthermore, note that many of the optimizations and training tips mentioned in section \ref{improvements_tips} have been applied; batch normalization, \ac{Tanh} activation at the output of the generator and \ac{ReLU} used between hidden layers.

\section{Dataset}\label{dataset_dcgan}

The album cover dataset was used for the experiments. To generate credible album covers, the model would need to extract the features from the input dataset, so this broad album cover dataset\footnote{https://www.kaggle.com/datasets/greg115/album-covers-images} was chosen as the base. it contains 80K album covers from all genres.

\section{Improvements}\label{improvements_dcgan}

The following section discusses the improvements that were done to the existing code, contrasting the original code with the modified one, and providing the rationale behind the modifications.

\subsection{Data Normalization}

The following code was used to transform and normalize the data.

\begin{figure}[H]
        \myfloatalign
        {
        \includegraphics[width=1\linewidth]{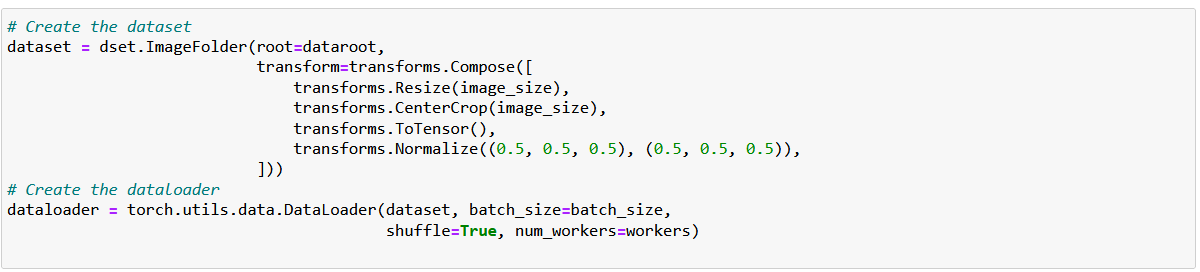}}
        \caption{Normalizing input data.}
        \label{fig:normalization_default}
\end{figure}

The transformation consists of several steps like resizing the image to the desired resolution, but the most notable is the normalization step. The normalize function takes the mean and \ac{STD} of the input data as parameters, of all 3 \ac{RGB} channels. In the tutorial code it is hard-coded since the values of the Celba dataset are close to that.

\begin{figure}[H]
        \myfloatalign
        {\label{fig:mean_std_pre}
        \includegraphics[width=1\linewidth]{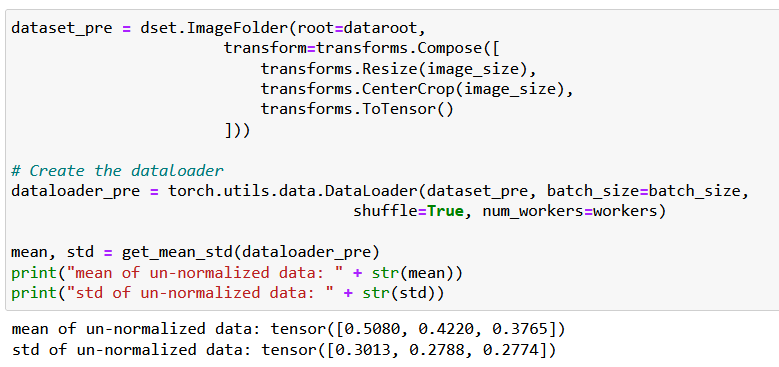}}
        \caption{Mean and \ac{STD} values of all 3 \ac{RGB} channels before code optimization.}
\end{figure}

To attempt to improve results and make the normalization step work for any input dataset, another normalization strategy was implemented as well. Instead of choosing 0,5 for both the mean and \ac{STD}, the mean of the different channels of the images in the dataset was pre-calculated, then these values were used to normalize the data, in order to achieve as close to a normally distributed dataset as possible.

\begin{figure}[H]
        \myfloatalign
        {\label{fig:mean_std_custom}
        \includegraphics[width=1\linewidth]{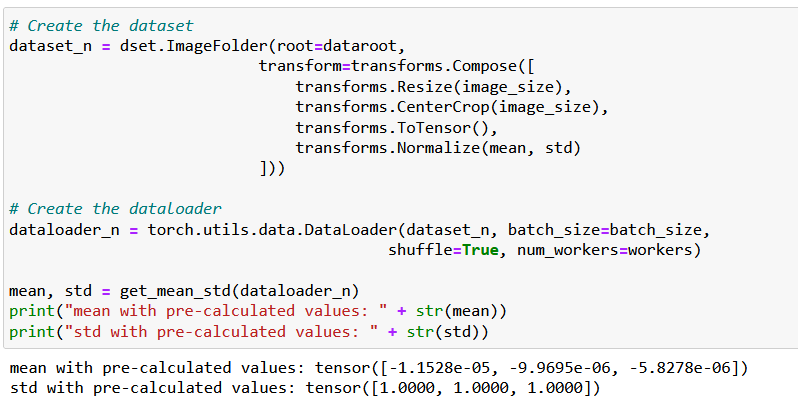}}
        \caption{Mean and \ac{STD} values of all 3 \ac{RGB} channels after code optimization.}
\end{figure}

Note that now the mean and \ac{STD} of the data is (close to) 0 and 1 respectively. The results this yielded in the output data will be discussed in the results section \ref{results_dcgan}.

\begin{lstlisting}[language=Python,
                   caption=Function used for calculating the mean and \ac{STD} values.]
# Function for getting the mean and std of the dataset channels
def get_mean_std(loader):
    # var[X] = E[X**2] - E[X]**2
    channels_sum, channels_sqrd_sum, num_batches = 0, 0, 0

    for data, _ in loader:
        channels_sum += torch.mean(data, dim=[0, 2, 3])
        channels_sqrd_sum += torch.mean(data ** 2, dim=[0, 2, 3])
        num_batches += 1

    mean = channels_sum / num_batches
    std = (channels_sqrd_sum / num_batches - mean ** 2) ** 0.5

    return mean, std
\end{lstlisting}

\subsection{Weight Initialization}

The other place for optimization was with the weight initialization. The tutorial code uses the following function for initializing the weights.

\label{default_init}
\begin{lstlisting}[language=Python,
                   caption=Default weight initialization code.]
def weights_init(m):
    classname = m.__class__.__name__
    if classname.find('Conv') != -1:
        nn.init.normal_(m.weight.data, 0.0, 0.02)
    elif classname.find('BatchNorm') != -1:
        nn.init.normal_(m.weight.data, 1.0, 0.02)
        nn.init.constant_(m.bias.data, 0)
\end{lstlisting}

If the operation is a convolution, it initialized the weights with a random distribution of mean 0 and \ac{STD} of 0,2, and if it is a batch normalization, then the mean will be 1 with \ac{STD} of 0,2 and bias of 0. \footnote{This might seem to counter the architecture diagrams, since they depict a layer as the result of the operations. It is illustrated this way since the convolutional and convolutional transpose layers are the ones that change the data shape.}

Instead, other weight initialization functions were made in an attempt to initialize weights that better fit each of the layers. For convolutional and convolutional transpose layers that use \ac{ReLU} as their activation function, the weights are initialized following the Kaiming (or He) initialization. It is a zero-centered Gaussian normal distribution with standard deviation of $\sqrt{2/{n}_{l}}$. $n_{l}$ is the inputs to the node. The full exoression can be seen in \ref{eq:He_init_math}.

\begin{eqfloat}
\myfloatalign
\begin{equation}
w_{l} \sim \mathcal{N}\left(0,  2/n_{l}\right)
\label{eq:He_init_math}
\end{equation}
\caption{He initialization distribution.}
\end{eqfloat}

For layers that did not use \ac{ReLU}, the normalized Xavier initialization method was used instead. This was useful for the last layers in the network since for the generator uses \ac{Tanh} and the discriminator uses Sigmoid for the activation functions right before the output layer. It is a uniform distribution described in equation \ref{eq:norm_xavier_init_math}.

\begin{eqfloat}
\myfloatalign
\begin{equation}
w_{l} \sim \mathcal{U} \left(\sqrt6/\sqrt{n_{l} + m_{l}}\right)
\label{eq:norm_xavier_init_math}
\end{equation}
\caption{Normalized Xavier initialization distribution.}
\end{eqfloat}

where $n_{l}$ is the number of inputs to a node (e.g. number of nodes in the previous layer) and $m_{l}$ is the number of outputs from the layer (e.g. number of nodes in the current layer) \cite{brownlee_weight_2021}.


From the original code you can see that it iterates through the layers defined in the generator or discriminator. In order to initialize the layers with the modified weights, the model code needs to be adjusted to make the layers attributes of the class, so they can be modified individually.

\begin{figure}[bht]
        \myfloatalign
        \subfloat[The original code.]
        {\label{fig:og_code_g}
        \includegraphics[width=.45\linewidth]{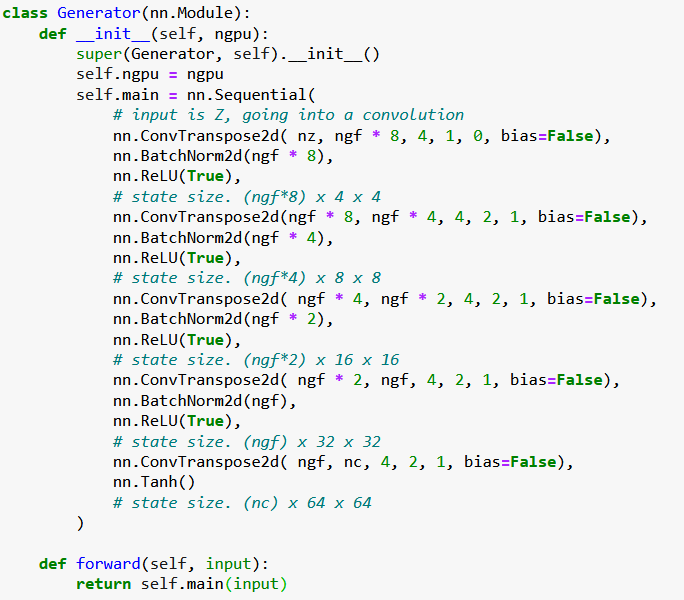}} \quad
        \subfloat[The modified code for initializing each layer individually.]
        {\label{fig:modified_code_g}
         \includegraphics[width=.45\linewidth]{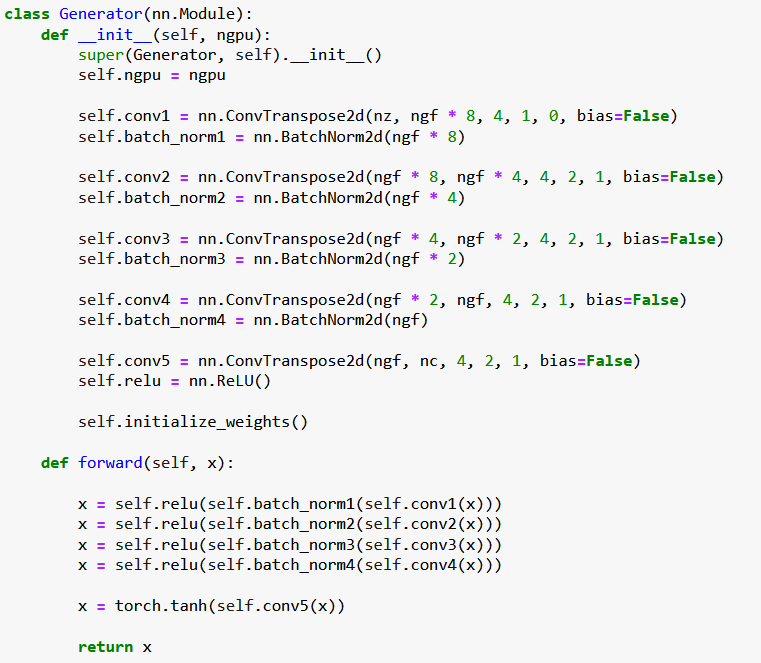}} \\
         \caption{Comparison between the original code and the optimizations.}
\end{figure}

Note that the architecture has not changed.

\label{custom_init}
\begin{lstlisting}[language=Python, caption=Custom weight initialization code.]
def initialize_weights(self):
    for m in self.modules():
        if isinstance(m, nn.Conv2d):
            nn.init.kaiming_normal_(m.weight, nonlinearity='relu')

            if m.bias is not None:
                nn.init.constant_(m.bias, 0)
        
        elif isinstance(m, nn.BatchNorm2d):
            nn.init.constant_(m.weight, 1)
            nn.init.constant_(m.bias, 0)
            
    #initialize last layer with weights more suited for Tanh and Sigmoid
    nn.init.xavier_normal_(self.conv5.weight)
\end{lstlisting}

\pagebreak

\section{Training}

The code was executed with the jupyter notebook provided by the web page of this \ac{GAN}, testing the different combinations specified in the previous section \ref{improvements_dcgan}, changing the generator and the discriminator code that was run and the loading and normalization of the dataset.

The hyperparameters used for training the model are shown in table \ref{tab:hyperparameters_dcgan}.

\begin{table}[bth]
    \myfloatalign
  \begin{tabularx}{\textwidth}{Xll} \toprule
    \tableheadline{Parameter} & \tableheadline{Value} \\ \midrule
    Total images & $79734$ \\
    Batch size & $128$ \\
    Batches/epoch & $623$ \\
    Image dimensions & $64x64x3$ \\
    Learning rate & $0.0002$ \\
    Adam's beta & $0.5$ \\
    Latent vector elements & $100$ \\
    Training epochs & ${5,12}$ \\
    Leaky \ac{ReLU} slope & $0.2$ (Discriminator) \\
    Weights initializer & \ref{default_init}{default} , \ref{custom_init}{custom} \\
    Batch normalization & YES \\
    Pixel value scale of training images & YES \\
    One-side label & smoothing NO \\
    Noisy Labels & NO \\
    Dropout & NO \\
    Progressive growing & NO \\
    Loss & Binary cross entropy \\
    \bottomrule
  \end{tabularx}
    \caption{Hyperparameters used to train the DCGAN model.}
  \label{tab:hyperparameters_dcgan}
\end{table}

\pagebreak

\section{Results}\label{results_dcgan}

This section discussed the results obtained from the different variations of the modifications used to train the model. 

\subsection{Unmodified Model Results}

The results of the album dataset with default parameters are the following;

\begin{figure}[H]
        \myfloatalign
        \subfloat[Album dataset loss plot over the training process.]
        {\label{fig:album_loss_default_12e}
        \includegraphics[width=0.9\linewidth]{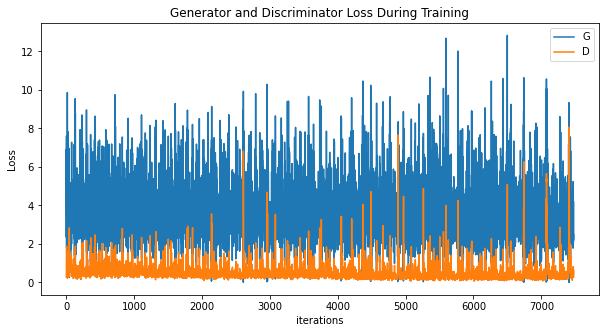}}\quad
        \subfloat[Album dataset Generated images vs. real images]
        {\label{fig:album_results_default_12e}
        \includegraphics[width=0.97\linewidth]{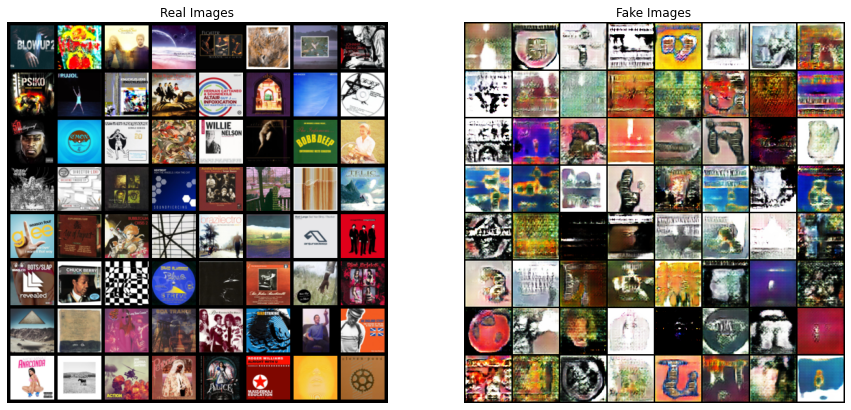}} \\
        \caption{Unmodified model training statistics and results.}
        \vspace{-20px}
\end{figure}

Note that the training seems to diverge and not improve after a certain point. For about the first 2000 iterations the generator loss has a downward trend but then diverges. This could indicate that the discriminator is no longer providing valuable feedback to the generator.

In order to improve the results, the modifications discussed in section \ref{improvements_dcgan}.

\subsection{Modified Normalization Results}
Here are the results with the modified data normalization technique.

\begin{figure}[H]
        \myfloatalign
        \subfloat[Modified normalization loss plot over the training process.]
        {\label{fig:album_loss_custom_norm_12e}
        \includegraphics[width=0.88\linewidth]{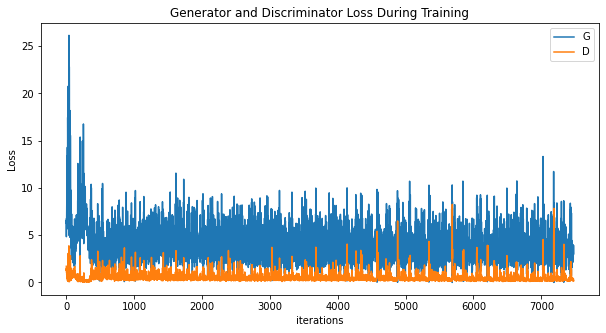}} \quad
        \subfloat[Modified normalization generated images vs. real images.]
        {\label{fig:album_results_custom_norm_12e}
        \includegraphics[width=0.95\linewidth]{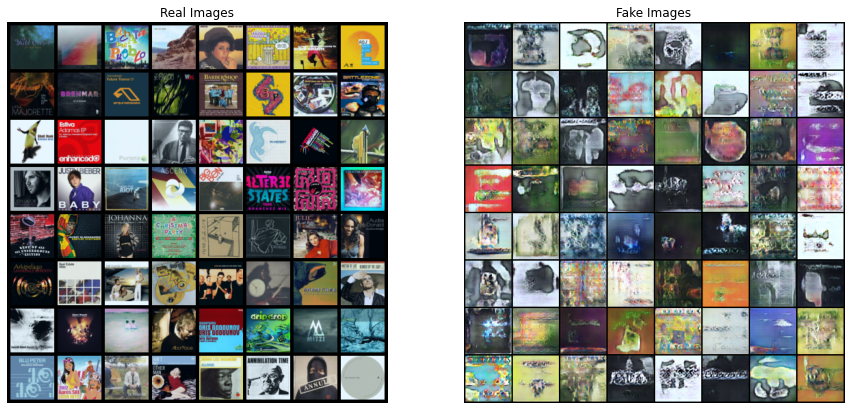}} \\
        \caption{Modified image normalization training statistics and results.}
        \vspace{-25px}
\end{figure}

\pagebreak

\subsection{Modified Weight Initialization Results}
Here are the results with the modified weight initialization technique.

\begin{figure}[bth]
        \myfloatalign
        \subfloat[Modified weight initialization loss plot over the training process.]
        {\label{fig:album_loss_custom_weight_12e}
        \includegraphics[width=0.9\linewidth]{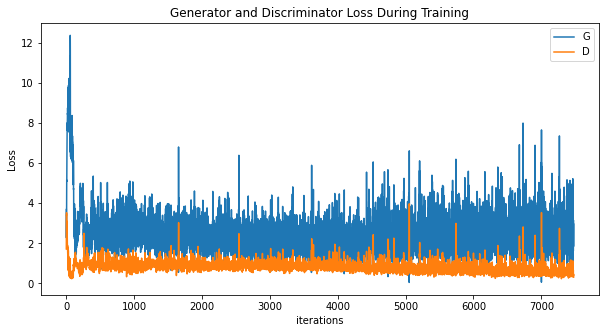}} \quad
        \subfloat[Modified weight initialization generated images vs. real images.]
        {\label{fig:album_results_custom_weight_12e}
        \includegraphics[width=0.95\linewidth]{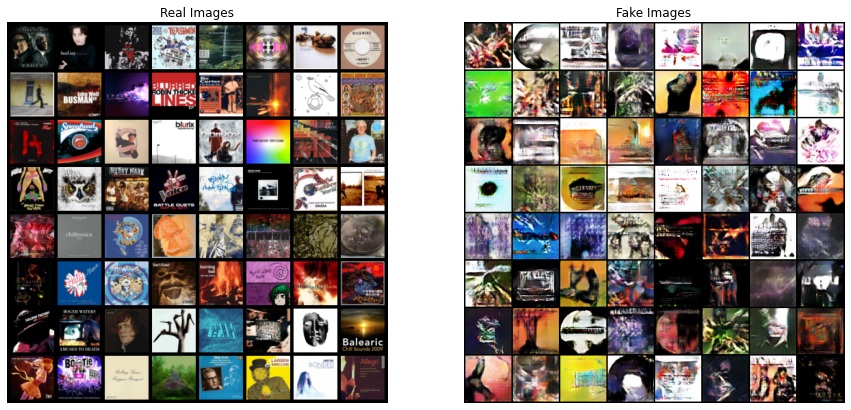}} \\
        \caption{Modified weight initialization model training statistics and results.}
\end{figure}

\pagebreak

\subsection{Modified Weight Initialization + Normalization Results}
\begin{figure}[bth]
        \myfloatalign
        \subfloat[Loss plot over the training process, model with both modifications applied.]
        {
        \includegraphics[width=0.8\linewidth]{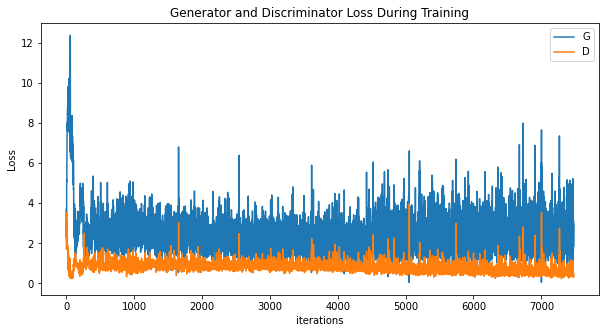}} \quad
        \subfloat[Generated images vs. real images, model with both modifications applied.]
        {
        \includegraphics[width=0.85\linewidth]{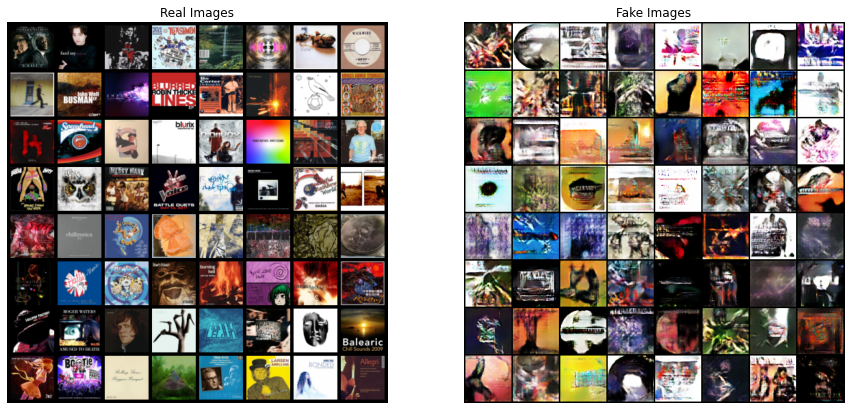}} \\
        \caption{Fully improved model training statistics and results.}\label{fig:album_results_both_12e}
        \vspace{-10px}
\end{figure}

The combination of both previous improvements is shown in figure \ref{fig:album_results_both_12e}.

\section{Conclusion}\label{modified_dcgan_conclusion}

Analyzing all the results, there seems to be little appreciable difference, especially in the results. Not only are the results of this \ac{GAN} not perfect, but adding layers to this \ac{GAN} was attempted to generate higher resolution images, but the training time skyrocketed and the results became worse. However, the loss graph of both improvements applied seems to stabilize the discriminator loss. This can be seen by the lower spread in the loss numbers for the discriminator loss, and the lower peaks versus the non-modified model. This is the case at least up until around 4000 iterations where the generator loss begins to climb again, indicating the discriminator is no longer providing meaningful feedback (mode collapse). This would confirm that normalizing the data and initializing the weights taking into account the architecture of the \ac{GAN} does stabilize training.

The generated images themselves however are what is being analyzed as the final result. 

Comparing the results from the unmodified model we can see that the difference in the extracted features is not significant. The models that generated these images were saved and 9 individual images were generated to see the results more in detail.

\begin{figure}[bth]
        \myfloatalign
        \subfloat[Default model fakes.]
        {\label{fig:fake9_default}
        \includegraphics[width=.45\linewidth]{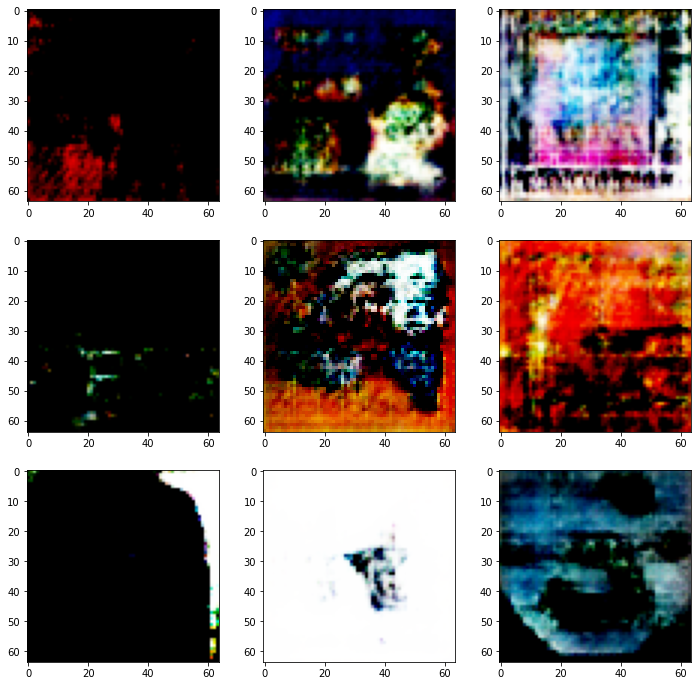}}
        \quad
        \subfloat[Modified model fakes.]
        {\label{fig:fake9_modified}
        \includegraphics[width=.45\linewidth]{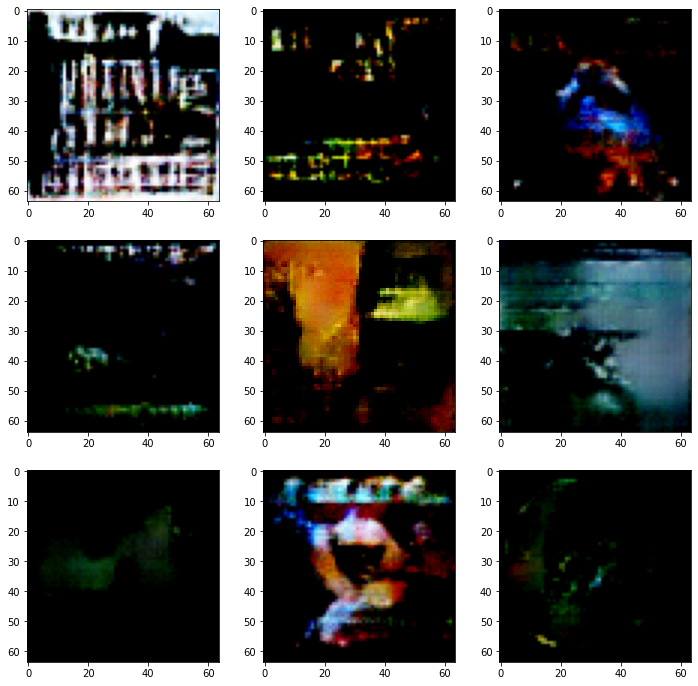}} \\
        \caption{Fully improved model curated images.}
\end{figure}

There might be an argument in favour of the modified model from these specimens since it does seem to have more detail on some of the features captured, however generated images from both models look very similar in their composure.

With this \ac{GAN}, it is difficult to make the model generate images that capture the fine and overall features of the album covers in the input dataset. Furthermore, this architecture has a scaling problem. If a higher resolution would be desired, for example 128x128, either more layers need to be added to both the generator and the discriminator, or the dimensions of the layers need to be changed. In the first case, the model training time sky-rockets, and in the second case, the feature extraction will be worsened since there is more data being fed into the model but the model will be unable to capture it.

%% file: Chapters/StyleGAN2.tex
\chapter{StyleGAN2}\label{ch:StyleGAN2}

After the conclusions from experimenting with the last model, it became more clear that StyleGAN2\footnote{\url{https://github.com/NVlabs/stylegan2-ada-pytorch}} \cite{noauthor_nvlabsstylegan2-ada-pytorch_2022} was a perfect fit for the task, since it uses data augmentation to prevent overfitting and a progressive architecture to capture the general and specific features of the input data and generate high resolution images. It also uses a style based generator, which means that the latent vector is mapped into a style mapping instead of being used directly to generate the image like a traditional \ac{GAN}.

\section{Architecture}

This section discusses architecture elements that are most relevant for styling images, enabling the \ac{GAN} faster training times and allowing the use of smaller datasets as opposed to previous model architectures.

\subsection{Progressive architecture}
The architecture of this \ac{GAN} is based on the same concept of having several convolutional layers as the previous \ac{GAN}. The difference is these layers are not trained all at once, and instead are progressively trained.

\begin{figure}[H]
        \myfloatalign
        {\includegraphics[width=0.8\linewidth]{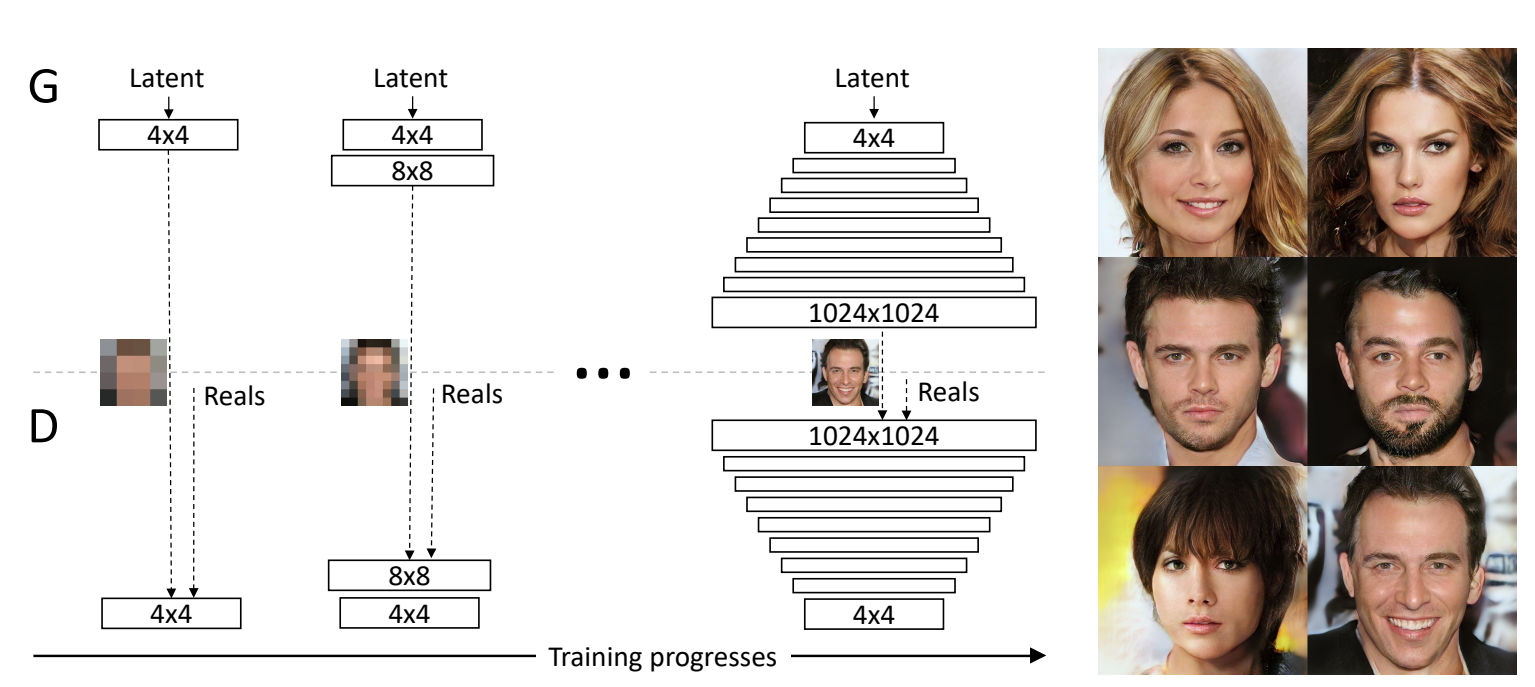}}
        \caption{Progressive architecture illustrated. Source: \cite{karras_progressive_2018}.}
        \label{fig:progressive_arch}
\end{figure}

\begin{figure}[H]
        \myfloatalign
        {\includegraphics[width=0.9\linewidth]{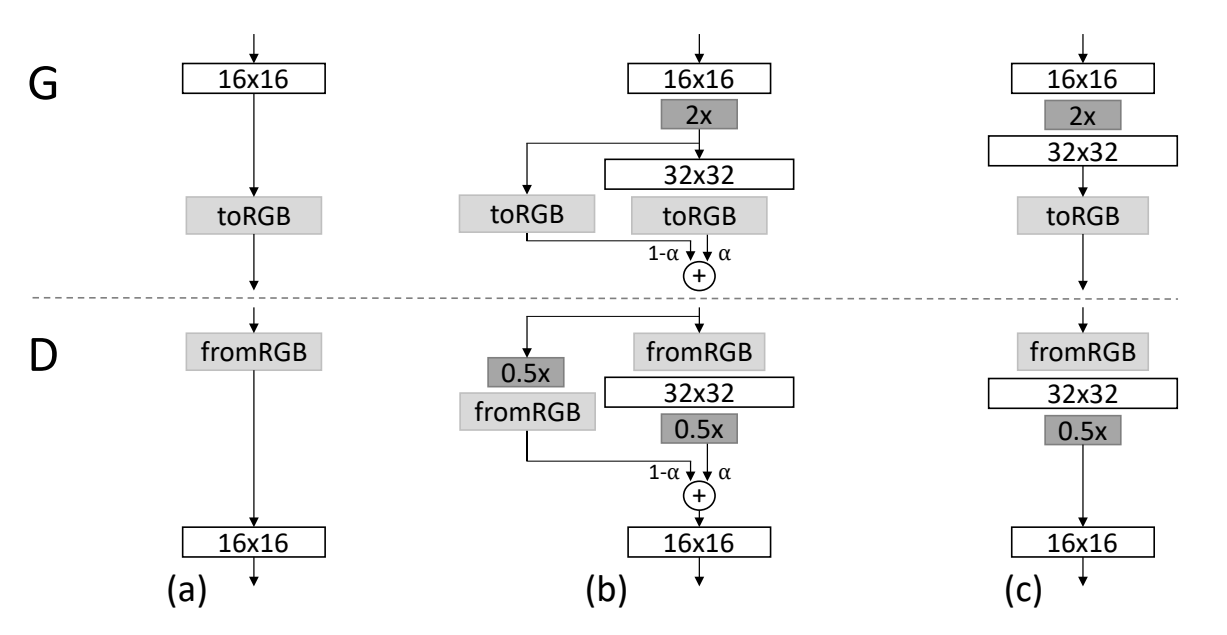}}
        \caption{Example of fading in new layers. The resolution is doubled with each new layer and the toRGB represents a layer that projects feature vectors to RGB colors and fromRGB does the reverse; both use 1 × 1 convolutions \cite{karras_progressive_2018}. Source: \cite{karras_progressive_2018}.}
        \label{fig:progressive_arch_detailed}
\end{figure}

Progressively training the layers involves by starting with the lowest convolutional layers (4x4 layers in figure \ref{fig:progressive_arch}), training them and then transitioning to training the layer that follows. The transition is smooth, so once the transition happens, it happens by gradually phasing out the previous layer and fading in the new one with a changing weight of $\alpha$ \cite{karras_progressive_2018}. The transition process is illustrated in figure \ref{fig:progressive_arch_detailed}.

The real images that are fed to the discriminator are down scaled accordingly to the output resolution of the generator for a particular step in the training process \cite{karras_progressive_2018}.

This enables the \ac{GAN} to train much faster, since most of the training iterations are done at lower resolutions \cite{karras_progressive_2018}.

\subsection{Style Mapping}

\begin{figure}[H]
        \myfloatalign
        {\includegraphics[width=0.6\linewidth]{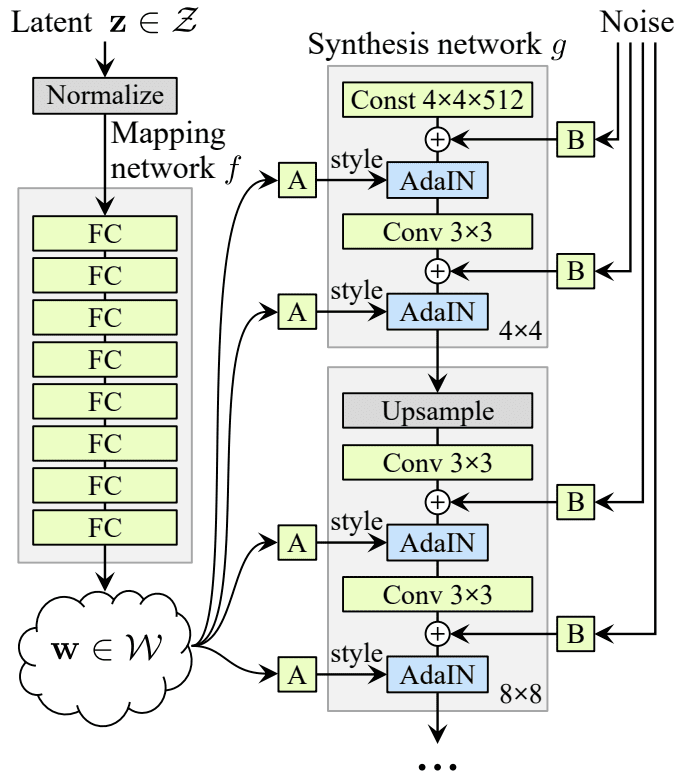}}
        \caption{Style based architecture. Source: \cite{karras_style-based_2019}.}
        \label{fig:style_arch}
\end{figure}

As mentioned, this model architecture makes the use of styles in the generator to generate images. Figure \ref{fig:style_arch} displays the style based architecture.

Instead of directly using the latent vector $Z$ to generate images, it is fed through a mapping network that will output as many styles as there are progressive layers times 2 (if there are 7 convolutional layers then there will be 14 style vectors). A particular style vector is then transformed and incorporated into each block of the generator model after the convolutional layers via an operation called adaptive instance normalization or AdaIN \cite{brownlee_gentle_2019-3}. This involves first standardizing the output of the feature map (output of the convolutions) to a standard Gaussian, then adding the style vector as a bias term.

\begin{eqfloat}
\myfloatalign
\begin{equation}
AdaIN( x_{i} ,y) \ =\ y_{s}{}_{,}{}_{i} \ \frac{x_{i} \ -\ \mu ( x_{i})}{\sigma ( x_{i})} \ +\ y_{b}{}_{,}{}_{i}
\label{eq:ada_in}
\end{equation}
\caption{AdaIN mathematical representation \cite{karras_style-based_2019}.}
\end{eqfloat}

\subsection{Adaptive Discriminator Augmentation}

The progressive architecture improvement leads to drastically reduced training times and allows the network to capture the input features of the data with more precision. However, Karras et al. \cite{karras_training_2020} improved on the results of this \ac{GAN} by augmenting the images that . both the generator \textit{and} the discriminator see. Image augmentation means to apply a transformation to the images, for example a 45 degree rotation. 

Why would this work if neither the discriminator nor the generator see what an actual un-transformed real or fake image looks like? Essentially, in a normal situation, the discriminator is comparing the generated data (distribution $X$) with the real data (distribution $Y$) and seeing how well they match. In the case that both are augmented, the discriminator will compare transformed generated data (distribution $TX$) with transformed real data (distribution $TY$). If a point is reached where $TX = TY$, then theoretically $X$ should equal $Y$ if the transformation is invertible. The researchers found that this leads to the generator being able to produce images like the input distribution.

\begin{figure}[bth]
        \myfloatalign
        {
        \includegraphics[width=0.75\linewidth]{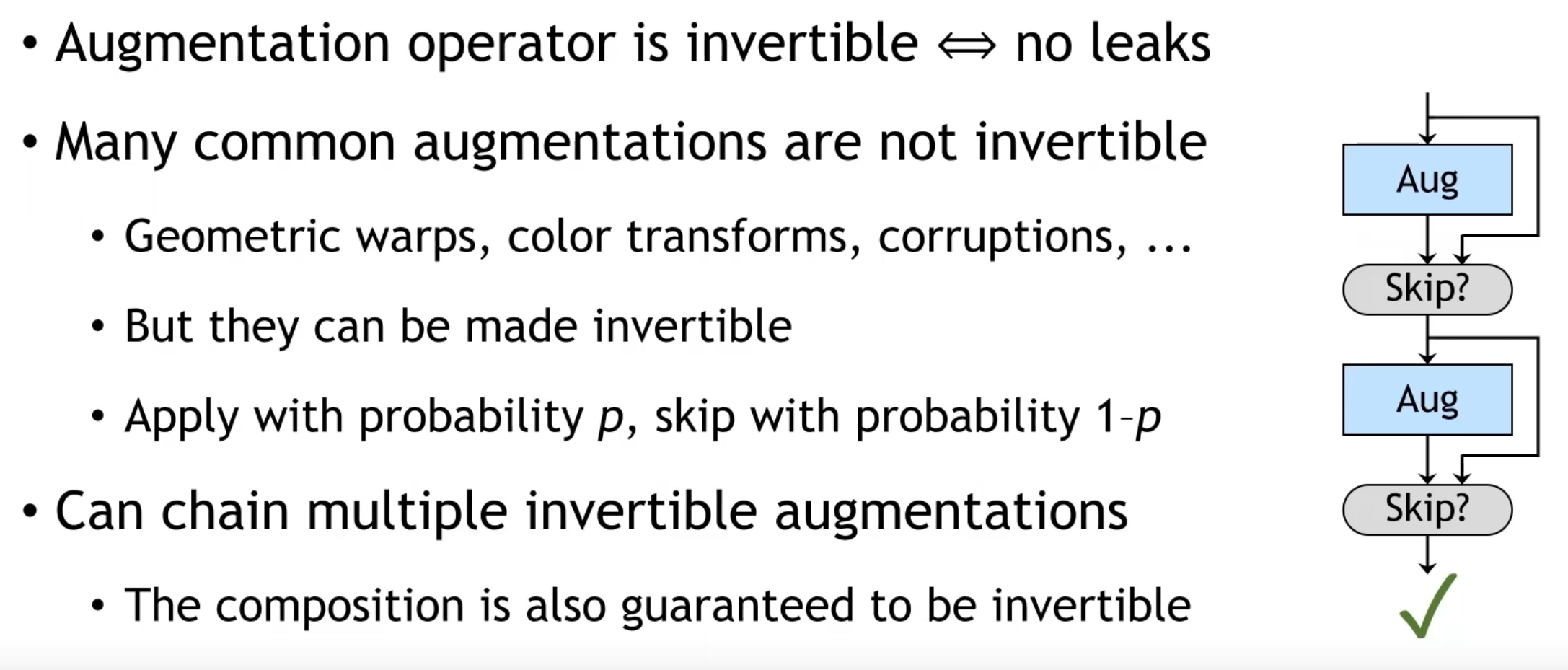}}
        \caption{Findings from researchers Karras et al. \cite{karras_training_2020} regarding augmentations. Source: \cite{finnish_center_for_artificial_intelligence_fcai_tero_2021}.}
        \label{fig:ada_pipeline}
\end{figure}

Furthermore, the augmentations were done adaptively for these models. Karras et al. \cite{karras_training_2020} also found that small datasets benefited from the augmentations but in large datasets it became harmful. They solved this problem by making the value of $p$ adaptive.

\begin{figure}[bth]
        \myfloatalign
        {
        \includegraphics[width=0.9\linewidth]{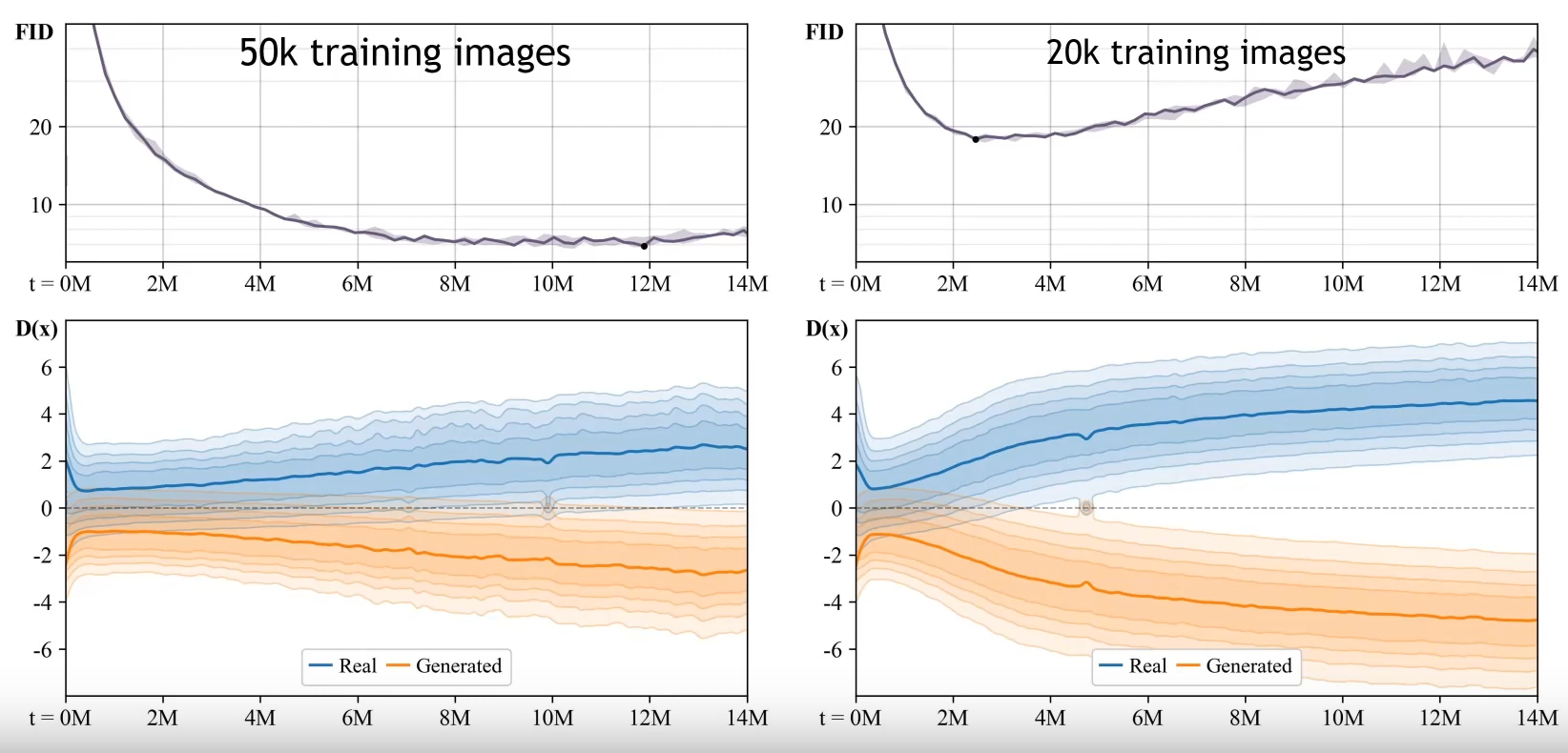}}
        \caption{Number of real and fake predictions by the discriminator. The drifting overlap between distributions shows that the discriminator gets better at telling fakes apart from reals, and this causes mode collapse. Source: \cite{finnish_center_for_artificial_intelligence_fcai_tero_2021}.}
        \label{fig:distribution_discriminator}
\end{figure}

Since the discriminator will begin to tell the real and fake images apart with more confidence as time goes on, the distributions of the predicted values of the generator (so if input image $x$ is fake or real) will drift apart more \cite{finnish_center_for_artificial_intelligence_fcai_tero_2021}. This is illustrated in figure \ref{fig:distribution_discriminator}.

The point of divergence in training seemingly coincides with the point where the distributions start drifting apart. So the Adaptive tuning will measure this overlap to either augment more or less. The measure is the average sign of the discriminator output. 

\begin{figure}[bth]
\myfloatalign
$ r_t = \mathbb{E}( sign(D_{train}) ) $
\caption{Average sign of the discriminator output.}
\label{fig:rt_calc}
\end{figure}

If it is too high, augment more, if it is too low augment less. This technique is called \ac{ADA} \cite{finnish_center_for_artificial_intelligence_fcai_tero_2021}.

Karras et al. \cite{karras_training_2020} found a target value of $r_t = 0.6$ that works equally well for all datasets tested and they used that..

This is actually the key to avoiding overfitting, as it adapts to the results and modifies the augmentation probability value $p$ to keep the training from diverging.

\subsection{Bidirectional GAN}

Another important feature of this model is it's ability also map input images into the latent space of the model. The Generator has a 'Mapping' module that is also trained. This module works similarly to a \ac{VAE}. A \ac{VAE} consists in both a decoder and an encoder, that are trained simultaneously. The encoder has the mission of compressing the data, and the decoder of decompressing it \cite{noauthor_variational_2022}.

\begin{figure}[bth]
        \myfloatalign
        {\label{fig:VAE}
        \includegraphics[width=0.65\linewidth]{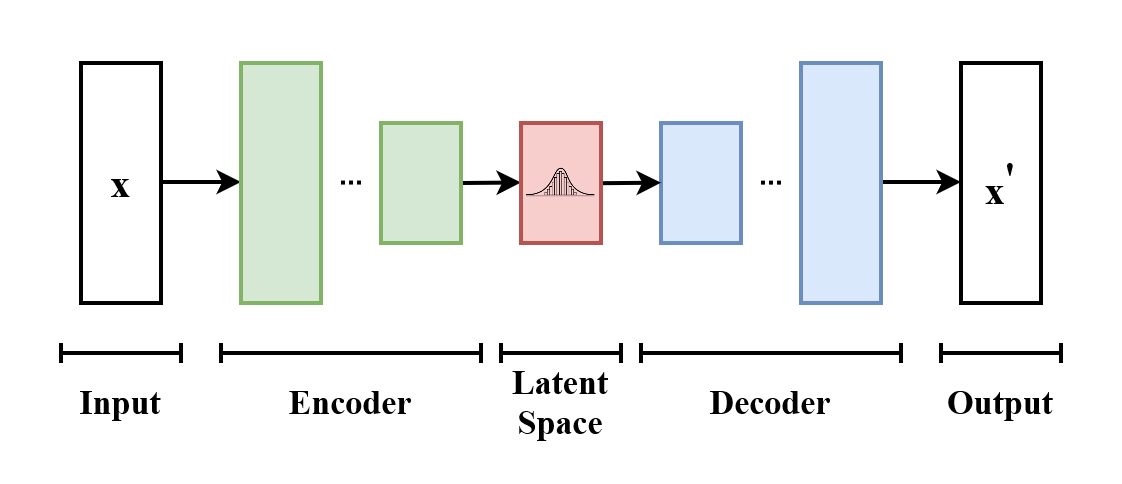}}
        \caption{Basic \ac{VAE}. Source: \cite{noauthor_variational_2022}.}
\end{figure}

The similarity is that the Mapping module contained in the generator of the StyleGAN2 code behaves like an encoder, mapping the input image into the latent space of the model that has been trained.

\section{Setup}\label{stylegan_setup}

The following covers all the setup needed and the code used to generate images. The results of using this code can be found in the results section \ref{styleGAN2_results}.

\subsection{Dataset}

The first step was to download the \ac{GAN} from the \href{https://github.com/NVlabs/stylegan2-ada-pytorch}{StyleGAN2} GitHub repository. Once downloaded, the album archive dataset needed to be prepared in order to be used for training.

The following command prepares the input data with the format required by the training script and it also serves to clean up the data if there are images of different resolutions by setting them all to the same resolution with the --width/height options.

\begin{lstlisting}[language=Bash, caption=Dataset generation script]
python dataset_tool.py --source=/workspace/shared/archive \
--dest=/workspace/shared/archive256 --width=256 --height=256
\end{lstlisting}

Furthermore, with this \ac{GAN} the specific genre images could be generated and the results evaluated better since it generates higher quality images. So for the specific genre data, the following script was used to download many album covers of a specific genre.

The hypothesis is this will train a model that will know what a generic album looks like, to then be tailored to different genres by feeding it latent space vectors from the images that are genre specific.

For the specific data, the Spotify \ac{API} was queried for playlists of the query specified.

\begin{lstlisting}[language=Python, caption=Code for downloading images from playlists given a query.]
playlist_ids = []

# Get a bunch of playlist ids
RANGE=5

for i in range(RANGE):
    results = sp.search(QUERY, limit=50, type='playlist')
    
    while 1:
        for idx, item in enumerate(results['playlists']['items']):
            playlist_ids.append(item['id'])
        
        if results['playlists']['next'] is None:
            break
        try:
            results = sp.next(results['playlists'])
        except:
            print("exception, continuing to next playlist.")
            break
\end{lstlisting}

From this list of playlist ids obtained from the search results, the program then retrieved all the songs contained within, and saved the album id's to a python dictionary, the key being the unique album id, and the value the image download \ac{URL}. By using a Python dictionary duplicate albums were avoided.

\begin{lstlisting}[language=Python, caption=Save album ids and \acp{URL} to a dictionary for downloading them.]
albums = {}

for i in range(len(playlist_ids)):
    results = sp.playlist(playlist_ids[i])
    print(results)
    for idx, item in enumerate(results['tracks']['items']):
        if item['track'] is None or len(item['track']['album']['images']) < 2:
            continue
        albums[item['track']['album']['id']] = item['track']['album']['images'][1]['url']
\end{lstlisting}

These steps were taken since the music \acp{API} found did not provide direct endpoints for querying for albums searching by genre.

\begin{lstlisting}[language=Python, caption=Code for downloading the images.]
def download(album):
    urllib.request.urlretrieve(album[1], FOLDER + album[0] + EXTENSION)

#Convert to list of tuples so the file can be named after the unique id
list_albums = [(k, v) for k, v in albums.items()]    

PROCESSES = multiprocessing.cpu_count() - 1

start = time.time()

with multiprocessing.Pool(PROCESSES) as p:
        p.map_async(
            download,
            list_albums
        )
        # clean up
        p.close()
        p.join()

print(f"Time taken = {time.time() - start:.10f}")

\end{lstlisting}

This code downloaded thousands of images. However, this turned out to be a problem since generating latent space vectors of thousands of images was very time consuming, in the order of days. This script was therefore further refined to take a single playlist (or more but they needed to be explicitly specified) with a reduced number of songs (around 100) of unique albums, that the user could freely make and these images would then be downloaded, the vectors generated, and the images generated using those vectors as a base. This new script also gave more flexibility since the user was free to add whatever they felt to a playlist and see the results.

\begin{lstlisting}[language=Python, caption=Updated script for getting the album ids given one or many playlist ids.]
albums = {}
# Spotify ID(s) of the playlists
playlist_ids = ['4uj6bJgjBtW7r91ognROc7']

for i in range(len(playlist_ids)):
    results = sp.playlist_items(playlist_ids[i], fields="items(track(album(id,images))),next")

    #print(results)
    while 1:
        for idx, item in enumerate(results['items']):
            if item['track'] is None or item['track']['album'] is None or len(item['track']['album']['images']) < 2:
                continue
            albums[item['track']['album']['id']] = item['track']['album']['images'][1]['url']
            
        if results['next'] is None:
            break
        try:
            results = sp.next(results)
        except:
            print("exception, continuing to next playlist.")
            break
\end{lstlisting}

This code is also helpful in that it gathers the album info in the same loop as the playlists, unlike before where they were separate loops.

\subsection{Latent vector generation}

In order to feed these images to the network as "inspiration", they needed to be the same shape as the input data (256x256 in this case) and then converted to latent space vectors. This was done with the trained model and helper scripts from the \href{https://github.com/NVlabs/stylegan2-ada-pytorch}{StyleGAN2} GitHub repository.

Since for this task, the images simply needed to be the same size, the dataset\_tool.py script was omitted and a simpler method of converting the images to the desired resolution was used.

\begin{lstlisting}[language=Python, caption=Code for reshaping images to the input resolution.]
# Re-shape images to the resolution used by the GAN
for file in os.listdir(SPOTIFY_IMAGE_FOLDER):
    im = Image.open(os.path.join(SPOTIFY_IMAGE_FOLDER, file))
    if im.mode == 'L':
        print(file)
        im = im.convert(mode='RGB')
    im = im.resize((RESOLUTION_GAN, RESOLUTION_GAN), Image.Resampling.LANCZOS)
    im.save(os.path.join(DATASET_IMAGE_DIR, file), format="png")
\end{lstlisting}

Then the projector.py script generated latent space vectors of a single image latent space vector. Since many images needed to be processed, a python script was made in order to run this python script on all the images in a directory.

\begin{lstlisting}[language=Python, caption=Latent vector generation code. This code is the part that takes the longest.]
index = 0
# Convert all images to vectors and output them to the specified vector folder
for subdir, dirs, files in os.walk(DATASET_IMAGE_DIR):
    for file in files:
        if not file.endswith(EXTENSION):
            continue
        output = VECTORS_DIR + str(index)
        index +=1
        command = 'python3 ' + BASE_SCRIPTS_DIR + 'projector.py' + \
        ' --outdir=' + output + ' --target=' + os.path.join(subdir, file) + \
        ' --network=' + NETWORK_DIR + ' --save-video=false' + ' --num-steps=600'
        os.system(command)
\end{lstlisting}

The resulting vectors are saved with a name of 'projected\_w.npz'. Their shape is $(14,512)$. This represents 14 style elements and 512 elements for the latent vector elements. 

To load the resulting vectors, the following command was used.

\begin{lstlisting}[language=Python, caption=Code for loading the generated vectors into the program.]
vectors = []
for subdir, dirs, files in os.walk(VECTORS_DIR):
    for file in files:
        if not file.endswith('.npz'):
            continue
        print(os.path.join(subdir, file))
        data = np.load(os.path.join(subdir, file))['w']
        vectors.append(data)
\end{lstlisting}

\section{Training}

Once the dataset has been configured, the training can be started with the following command:

\begin{lstlisting}[language=Bash, caption=Command to start training.]
python train.py --outdir=/workspace/shared/training-runs --data=/workspace/shared/archive256/ --gpus=1
\end{lstlisting}

There are several parameters that can be specified, like whether to calculate the metrics during the training or not, but they were left as defaults.

The training was stopped after 3d 10h 36m on a machine with a NVIDIA Quadro RTX 5000. See the section \ref{styleGAN2_results} for more details.

The specifications of the NVIDIA Quadro RTX 5000 GPU used are shown on table \ref{tab:quadro_stats}.

\begin{table}[bth]
    \myfloatalign
    \begin{tabularx}{\textwidth}{Xll} \toprule
        CUDA Parallel-Processing Cores &	3,072 \\
        NVIDIA Tensor Cores &	384 \\
        NVIDIA RT Cores &	48 \\
        GPU Memory &	16 GB GDDR6 \\
        RTX-OPS &	62T \\
        Rays Cast &	8 Giga Rays/Sec \\
        FP32 Performance &	11.2 TFLOPS \\
        Graphics Bus &	PCI Express 3.0 x 16 \\
        NVLink &	Yes \\
        Display Connectors &	DP 1.4 (4), VirtualLink (1) \\
        Form Factor &	4.4" (H) x 10.5" (L) Dual Slot \\
        VR Ready &	Yes \\
        \bottomrule
    \end{tabularx}
    \caption{NVIDIA Quadro RTX 5000 specifications.}
    \label{tab:quadro_stats}
\end{table}

\begin{figure}[H]
        \myfloatalign
        {\includegraphics[width=1\linewidth]{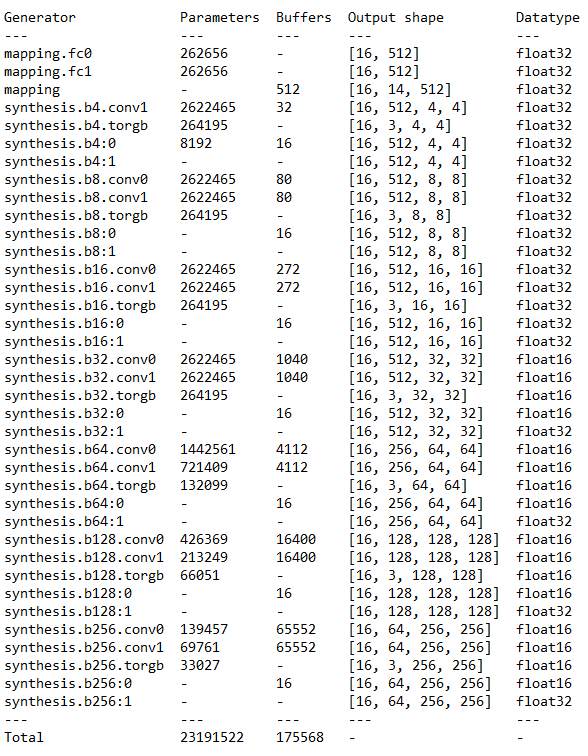}}
        \caption{Generator StyleGAN2 architecture.}
        \label{fig:generator_stylegan}
\end{figure}

\begin{figure}[H]
        \myfloatalign
        {\includegraphics[width=1\linewidth]{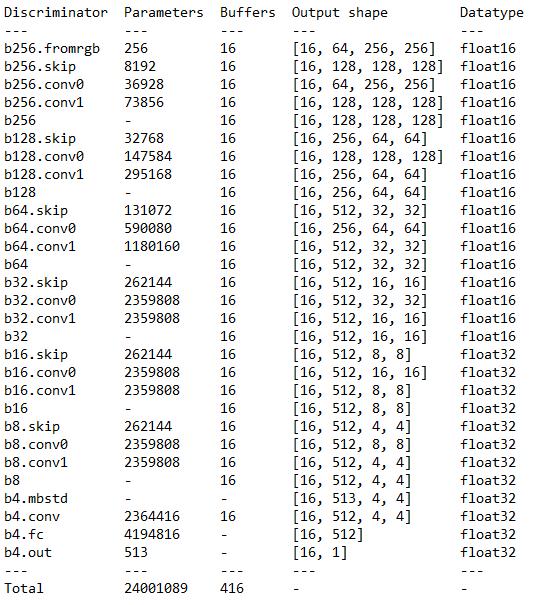}}
        \caption{Discriminator StyleGAN2 architecture.}
        \label{fig:discriminator_stylegan}
\end{figure}

The layers that were configured by the training script are shown in figure \ref{fig:generator_stylegan} and \ref{fig:discriminator_stylegan}.

\pagebreak

Table \ref{tab:hyperparameters_stylegan2} shows the hyperparameters used for the StyleGAN2 training.

\begin{table}[H]
    \myfloatalign
  \begin{tabularx}{\textwidth}{Xll} \toprule
    \tableheadline{Parameter} & \tableheadline{Value} \\ \midrule
    Total images & $79734$ \\
    Batch size & $16$ \\
    Image dimensions & $256x256x3$ \\
    Learning rate & see \ref{eq_lr} below \\
    Adam params. & $\beta1 = 0, \beta2 = 0.99, \epsilon=10^{-8}$ \\
    Latent vector elements & $512$ \\
    Reals shown to the discriminator (KIMG) & $6280.0$ \\
    \ac{ADA} target & $0.6$ \\
    Leaky \ac{ReLU} slope & $0.2$ \\
    Weight initializer & Normal random distribution \\
    Pixel value scale of training images & YES \\
    Batch normalization & NO (see \ref{pixelwise_norm} below) \\
    One-side label & smoothing NO \\
    Noisy Labels & NO \\
    Dropout & NO \\
    Progressive growing & YES \\
    Loss & see \ref{WGAN-GP} below \\
    \bottomrule
  \end{tabularx}
  \caption{Hyperparameters used to train the StyleGAN2 model.}
  \label{tab:hyperparameters_stylegan2}
\end{table}

The following is a list with some unique methods for parameter initialization used for this \ac{GAN}.

\begin{enumerate}
\item\texttt{Equalized Learning Rate}: Weights are initialized to the aforementioned normal random distribution $w_{l} \sim \mathcal{N}\left(0,1\right)$, and are then scaled at runtime such that $w_i = w_i/c$, where $w_i$ are the weights and $c$ is the per-layer normalization constant from He’s initializer \cite{karras_progressive_2018}\cite{he_delving_2015}.\label{eq_lr}

\item\texttt{Pixelwise feature vector normalization in the generator}: The feature vector in each pixel is normalized to unit length in the generator after each convolutional layer using a variant of local response localization \cite{karras_progressive_2018} \cite{krizhevsky_imagenet_2012}.\label{pixelwise_norm}

\item\texttt{WGAN-GP loss}: Wasserstein \ac{GAN} + Gradient Penalty \cite{gulrajani_improved_2017}, a \ac{GAN} that uses Wasserstein loss and a gradient penalty.\label{WGAN-GP}
\end{enumerate}

\section{StyleGAN2 Results}\label{styleGAN2_results}

This section discusses the training statistics, un-styled and styled generated images results, their features and how they are mixed. The loss and \ac{FID} graphs are taken from the Tensorboard metrics generated during the model training.

\subsection{Training statistics}

\begin{figure}[H]
\centering
\includegraphics[width=1\linewidth]{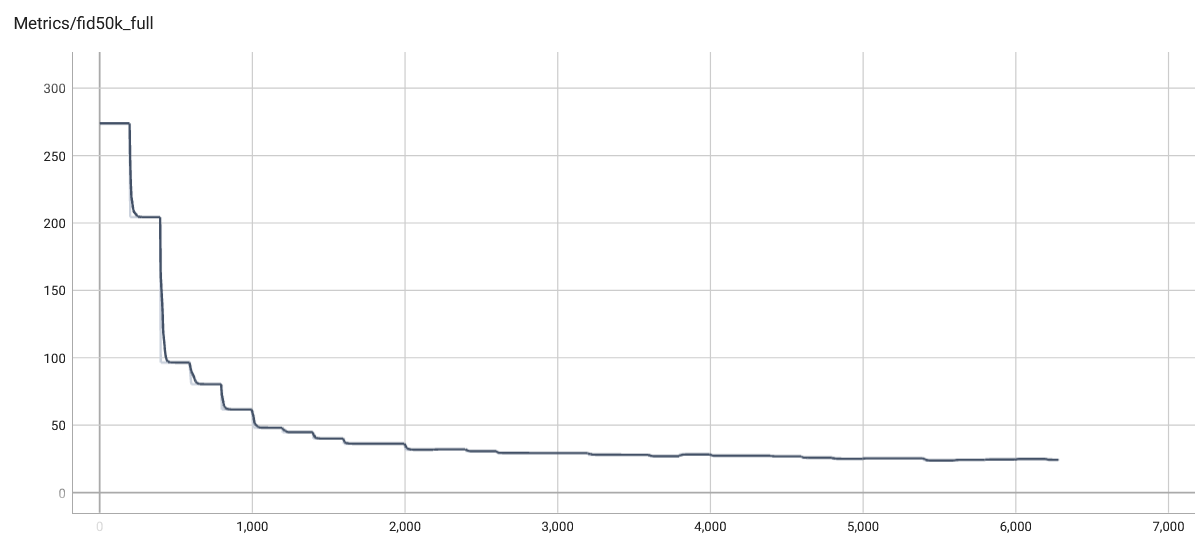}
\caption{The \ac{FID} plotted over time (minutes).}
\label{fig:fid_stylegan2}
\end{figure}

The \ac{FID} starts stagnating at around 2000-3000 \ac{KIMG}. This indicates the model is no longer learning to generate better images. An important note is that if the \ac{ADA} technique was not used, the results would begin to worsen (the training starts to diverge) instead of stay flat at around 2000-3000 \ac{KIMG}, indicating that the \ac{ADA} technique of the StyleGAN2 model is beneficial for this dataset and application as well. The model used for this section is at the 6200 \ac{KIMG} checkpoint. The \ac{FID} at that point is 24.28.

\begin{figure}[H]
        \myfloatalign
        \subfloat[The generator loss plotted over time (minutes).]
        {\includegraphics[width=.4\linewidth]{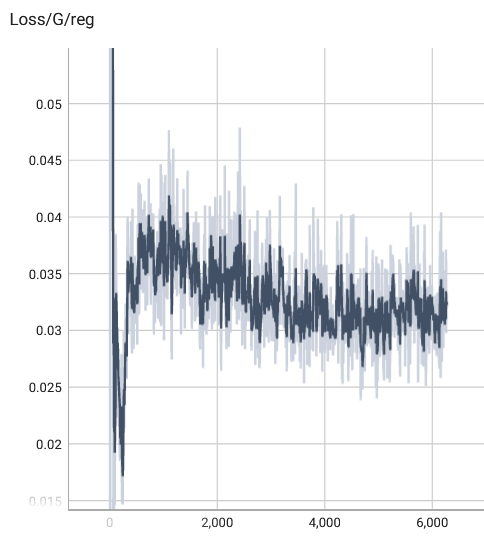}} 
        \quad
        \subfloat[The discriminator loss plotted over time (minutes).]
        {\includegraphics[width=.5\linewidth]{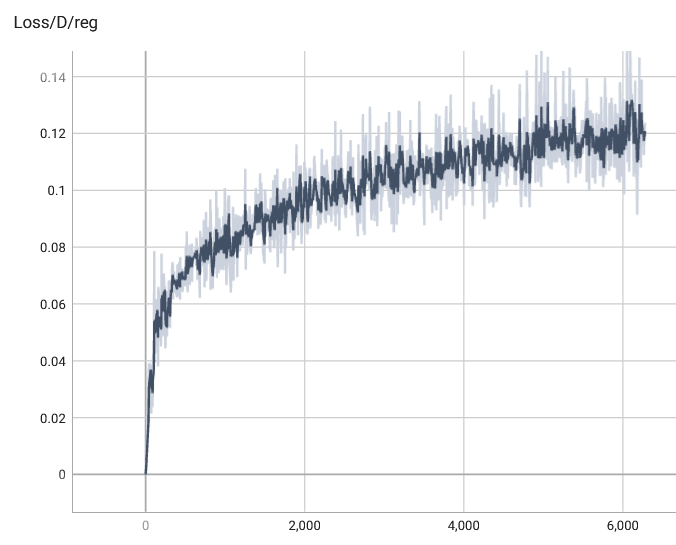}}
        \caption{Discriminator and Generator loss over time.}
        \label{fig:loss_stylegan2}
\end{figure}

In figure \ref{fig:loss_stylegan2} the discriminator loss becomes bigger with time since the generator becomes better at generating convincing fakes, hence why the generator's loss becomes less with time.

Once trained and the network pickle has been saved, it can be loaded into python scripts.

\begin{lstlisting}[language=Python, caption=Loading the network.]
with dnnlib.util.open_url(NETWORK_DIR) as f:
    Gs = legacy.load_network_pkl(f)['G_ema'].to(device) # type: ignore
\end{lstlisting}

Images can be generated using the generate.py script. The following command was used to generate 6 images, each with a different seed. Note the truncation parameter, this parameter simply re-samples values for the latent vector $z$ if they fall outside a specific value \cite{marchesi_megapixel_2017}.

\begin{figure}[H]
        \myfloatalign
        \subfloat[Seed = 600]
        {\label{fig:seed600}
        \includegraphics[width=.35\linewidth]{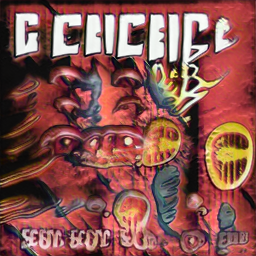}} \quad
        \subfloat[Seed = 601]
        {\label{fig:seed601}
        \includegraphics[width=.35\linewidth]{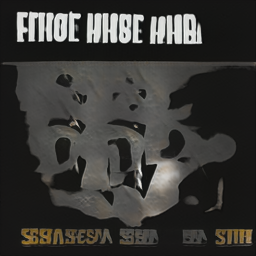}} \\
        \subfloat[Seed = 602]
        {\label{fig:seed602}
        \includegraphics[width=.35\linewidth]{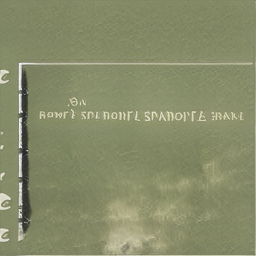}} \quad
        \subfloat[Seed = 603]
        {\label{fig:seed603}
        \includegraphics[width=.35\linewidth]{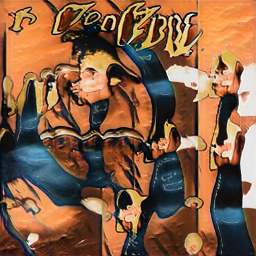}} \\
        \subfloat[Seed = 604]
        {\label{fig:seed604}
        \includegraphics[width=.35\linewidth]{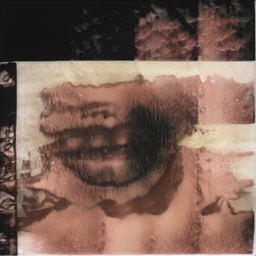}} \quad
        \subfloat[Seed = 605]
        {\label{fig:seed605}
        \includegraphics[width=.35\linewidth]{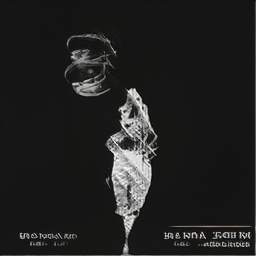}} \\
        \caption{Example of images generated by the model.}\label{fig:unmodified_images}
\end{figure}

The following command was used to generate un-styled images seen in figure \ref{fig:unmodified_images}. Seeds were used so that these results can be replicated.

\begin{lstlisting}[language=Bash, caption=Generate 6 album covers.]
python3 generate.py --outdir=/workspace/shared/results/ --trunc=0.7 --seeds=600-605 --network=/workspace/shared/StyleGAN2/stylegan2-ada-pytorch/00002-archive256-auto1/network-snapshot-006200.pkl
\end{lstlisting}

\subsection{Un-styled images}

\subsection{Styled image generation}

The latent space vectors that were generated by the projector.py script. Once generated, they were averaged. The averaged vector was used as the input for the generate.py script. The resulting image was one of the best and most promising so far.

Initially a large amount of images was used to generate a rock album cover. It involved using the Python code mentioned in \ref{stylegan_setup} to get 4283 images which all were included in rock themed Spotify playlists.

\begin{lstlisting}[language=Python, caption=Vector average code.]
average = np.mean(vectors, axis=0)

output = RESULTS_DIR
vector = RESULTS_DIR + f'test_projected_w.npz'

command = 'python3 ' + BASE_SCRIPTS_DIR + 'generate.py' + \
' --outdir=' + output + ' --projected-w=' + vector + ' --network=' + NETWORK_DIR

os.system(command)
\end{lstlisting}

\begin{figure}[H]
        \myfloatalign
        {\label{fig:rock_first}
        \includegraphics[width=.65\linewidth]{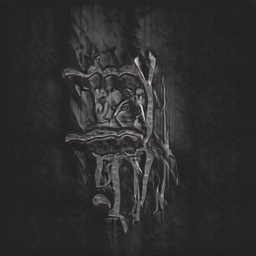}}
        \caption{First custom image generated by the \ac{GAN}.}
        \vspace{-25px}
\end{figure}

However, as mentioned in section \ref{stylegan_setup}, there was a problem with this approach. Generating the vectors from 4283 images took about 2 days and all the vector data took ~20GB of space. Therefore, another method was needed to generate specific images, in which custom playlist(s) created by the user with song from the desired albums were used.

The following examples use a Hotline Miami game soundtrack playlist\footnote{\url{https://open.spotify.com/playlist/4uj6bJgjBtW7r91ognROc7}} for the data. It contains 49 unique albums from the original soundtrack of the game Hotline Miami \cite{noauthor_hotline_2022}. It was used since the styles of the covers are similar and I wanted to replicate it in a generated cover while validating the idea.

\begin{figure}[H]
        \myfloatalign
        {\includegraphics[width=.6\linewidth]{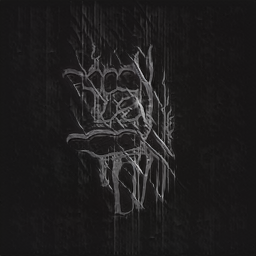}}
        \caption{Generated image with Hotline Miami latent vectors averaged.}
        \label{fig:rock_miami}
\end{figure}

After acquiring that data, generating the vectors and averaging them, the result was what is shown in figure \ref{fig:rock_miami}.

This was initially concerning, since not only did it look almost identical to the generated rock album cover, but it also does not look like any of the albums used, which meant at least 2 things. Firstly, this method of combining the images to form one with the features of all of them was not going to work, and second, the fact that the first album cover generated looked promising was simply by chance.

\begin{figure}[H]
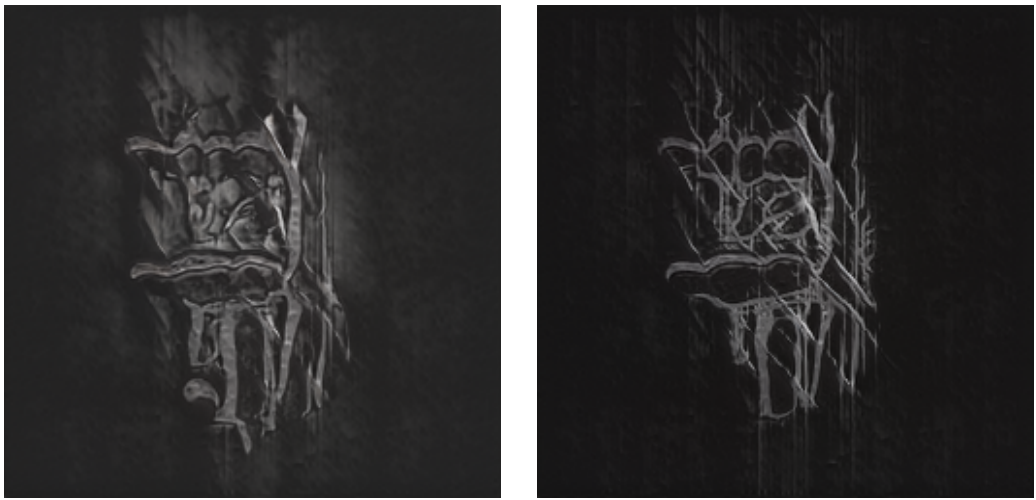

        \myfloatalign
        \subfloat[First custom image generated by the \ac{GAN}.]
        {\includegraphics[width=.45\linewidth]{gfx/results/first_rock_attempt.png}} 
        \quad
        \subfloat[Generated image with Hotline Miami latent vectors averaged.]
        {
        \includegraphics[width=.45\linewidth]{gfx/results/Hotline_miami_avg.png}}
        \caption{Average vector images.}
        \vspace{-15px}
        \label{fig:rock_miami_and_first}
\end{figure}

Since this method of styling the images did not work, I moved on to the other 2 mothods available for styling images intermediate vector interpolation and style mixing. The first thing attempted was to interpolate between 2 latent space vectors, and see if in this case the result would be a mix of the 2 input projected albums.

The linear interpolation and style mixing code were made with the help of code provided by Jae Won Choi \cite{md_these_2021}.

\begin{lstlisting}[language=Python, caption=Linear interpolation code.]
# LINEAR INTERPOLATION BETWEEN 2 IMAGES
IMAGE_A_VECTOR_PATH = '/workspace/shared/vectors_hotline_miami/48/projected_w.npz'
IMAGE_B_VECTOR_PATH = '/workspace/shared/vectors_hotline_miami/1/projected_w.npz'

point_A = np.load(IMAGE_A_VECTOR_PATH)['w']
point_B = np.load(IMAGE_B_VECTOR_PATH)['w']
lam = 0.5 # LINEAR INTERPOLATION

# Generate a transition between the images
# Number of steps
DIVISION = 50

images_array = []
interpolated_dir = RESULTS_DIR + "interpolated"
os.makedirs(interpolated_dir, exist_ok=True)

steps = np.arange(0.0, 1.0, 1/DIVISION)

for idx, i in enumerate(steps):
    lam = i # LINEAR INTERPOLATION
    print(lam)
    
    inter = lam*point_B+(1-lam)*point_A

    inter_img = Gs.synthesis(torch.from_numpy(inter).to(device))
    inter_img = (inter_img.permute(0, 2, 3, 1) * 127.5 + 128).clamp(0, 255).to(torch.uint8)
    images_array.append(inter_img[0].cpu().numpy())
    
    PIL.Image.fromarray(images_array[idx], 'RGB').save(f'{interpolated_dir}/{idx}.png')
\end{lstlisting}

\begin{figure}[H]
        \myfloatalign
        \subfloat[Image A projection.]
        {
        \includegraphics[width=.47\linewidth]{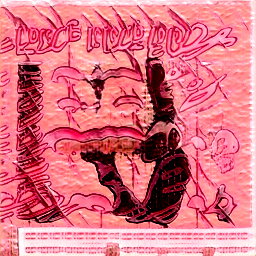}}
        \quad
        \subfloat[Image B projection.]
        {
        \includegraphics[width=.47\linewidth]{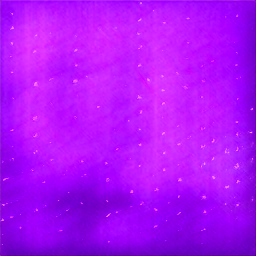}}
        \caption{Projection of the source images chosen.}
        \label{fig:interpolation_source}
\end{figure}

So 2 of the projected images were picked (figure \ref{fig:interpolation_source}) and interpolated between as a proof of concept.

\begin{figure}[H]
        \myfloatalign
        {\includegraphics[width=0.82\linewidth]{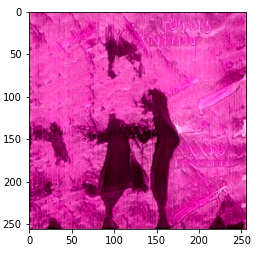}}
        \caption{Interpolation results.}
        \label{fig:interpolation}
        \vspace{-15px}
\end{figure}

The results shown in figure \ref{fig:interpolation} now not only proved promising but also the combination had features from both projected albums.

Both the color and the contents of the cover itself are being interpolated. This is because apart from interpolating between the 512 latent space elements that are the base of each image, we are also interpolating between the styles of both images. There is also the possibility to interpolate between more than 2 vectors, however this is left for future work.

The last step was to attempt to style mix images with the style vectors of the projected images.

\begin{lstlisting}[language=Python, caption=Style mixing code.]
mixed = {}
for idx in range(RANGE):
    for t in ids:
        mix = np.copy(dlatents[idx])
        mix[0][:7] = dlatents[t][0][:7] ## YOU CAN CHANGE WHICH LAYERS TO MIX
        miximg = Gs.synthesis(torch.from_numpy(mix).to(device))
        miximg = (miximg.permute(0, 2, 3, 1) * 127.5 + 128).clamp(0, 255).to(torch.uint8)
        mixed[(idx,t)] = miximg[0].cpu().numpy()
\end{lstlisting}

The most important piece of this last code is the style mixing line:

\begin{lstlisting}[language=Python, caption=Code to replace some styles of one vector to the other.]
mix[0][:7] = dlatents[t][0][:7] ## YOU CAN CHANGE WHICH LAYERS TO MIX
\end{lstlisting}

This code inserts the style information from one image's vector into the other, hence the style mixing name. The mix array then contains the mixed image vector, that then is fed to the generator model in order to generate the mixed image itself.

\begin{figure}[H]
        \myfloatalign
        \includegraphics[width=.95\linewidth]{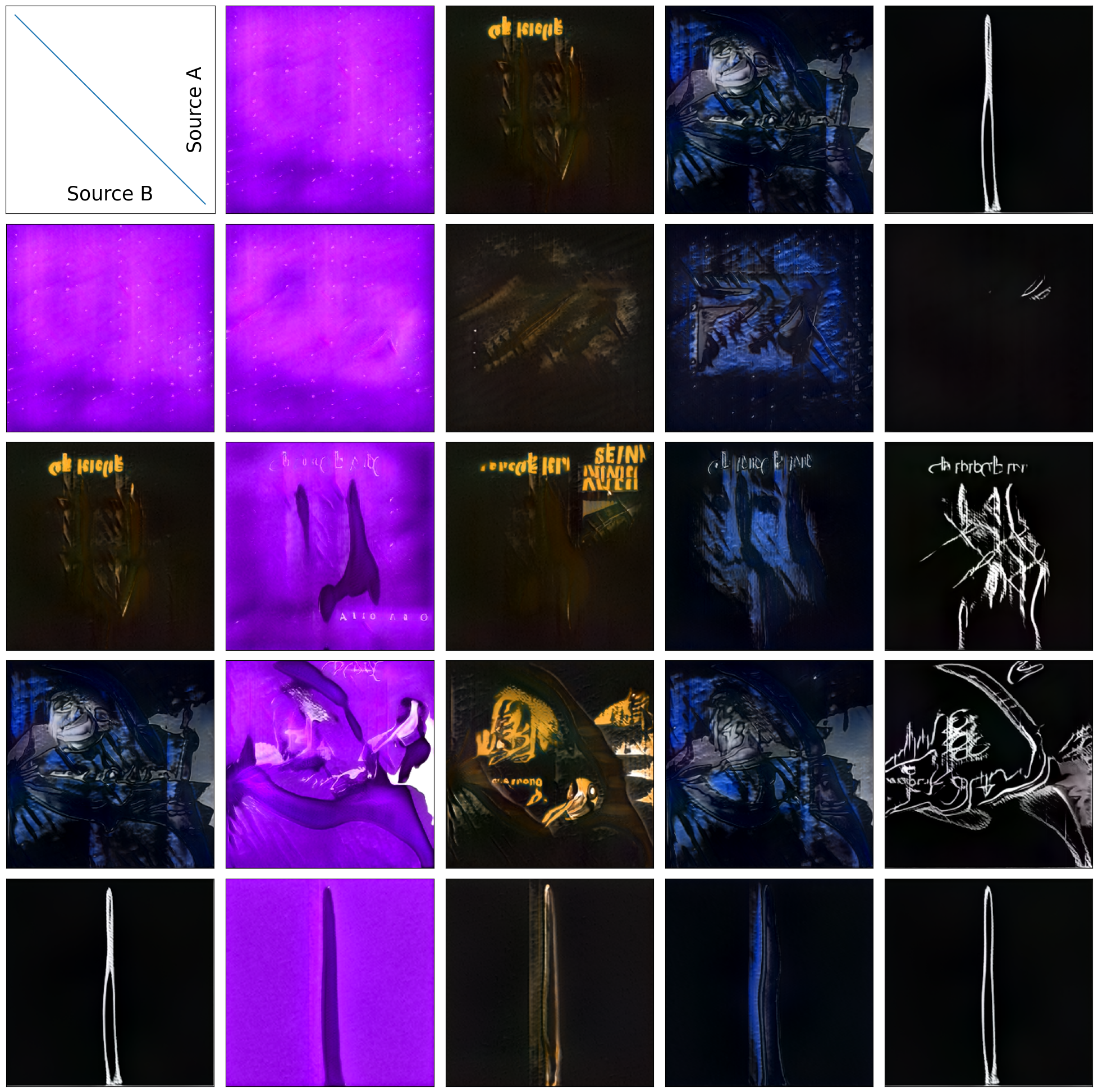}
        \caption{Style mixing the first 7 styles of some projected images.}
        \label{fig:style_mixing_styles}
        \vspace{-20px}
\end{figure}

From figure \ref{fig:style_mixing_styles} you can see the different projected album covers and how they mix with one another. For this image the first 7 styles were mixed while the rest were left the same.

\begin{figure}[H]
        \myfloatalign
        {
        \includegraphics[width=1\linewidth]{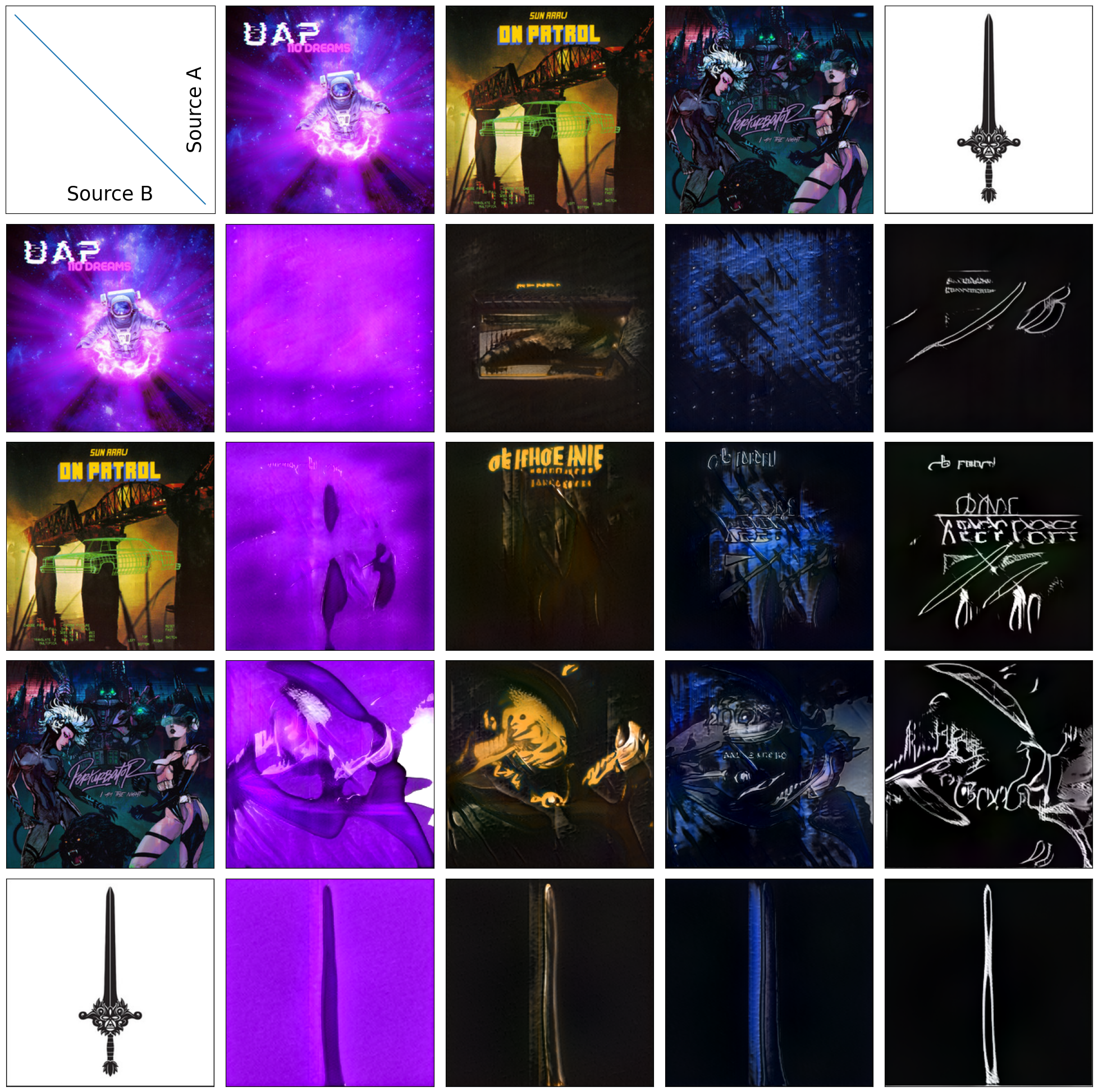}}
        \caption{Style mixing the first 7 styles, with the original images instead of their projections.}
        \label{fig:style_mixing_og}
        \vspace{-20px}
\end{figure}

Note that the source images are the projected images, not the original album covers. This is because the projection to the latent space of the model is not perfect. As a reference, figure \ref{fig:style_mixing_og} displays the source images, not the projections of those images on the vector space of the model. The most notable characteristic of these images is that, when mixing half of the styles (7 in this case), the resulting image gets it's color from source A and its shape from source B. This is a great example of style mixing since most images in this grid show this behaviour.

\begin{figure}[H]
        \myfloatalign
        \subfloat[Mixing 12 styles.]
        {\includegraphics[width=.7\linewidth]{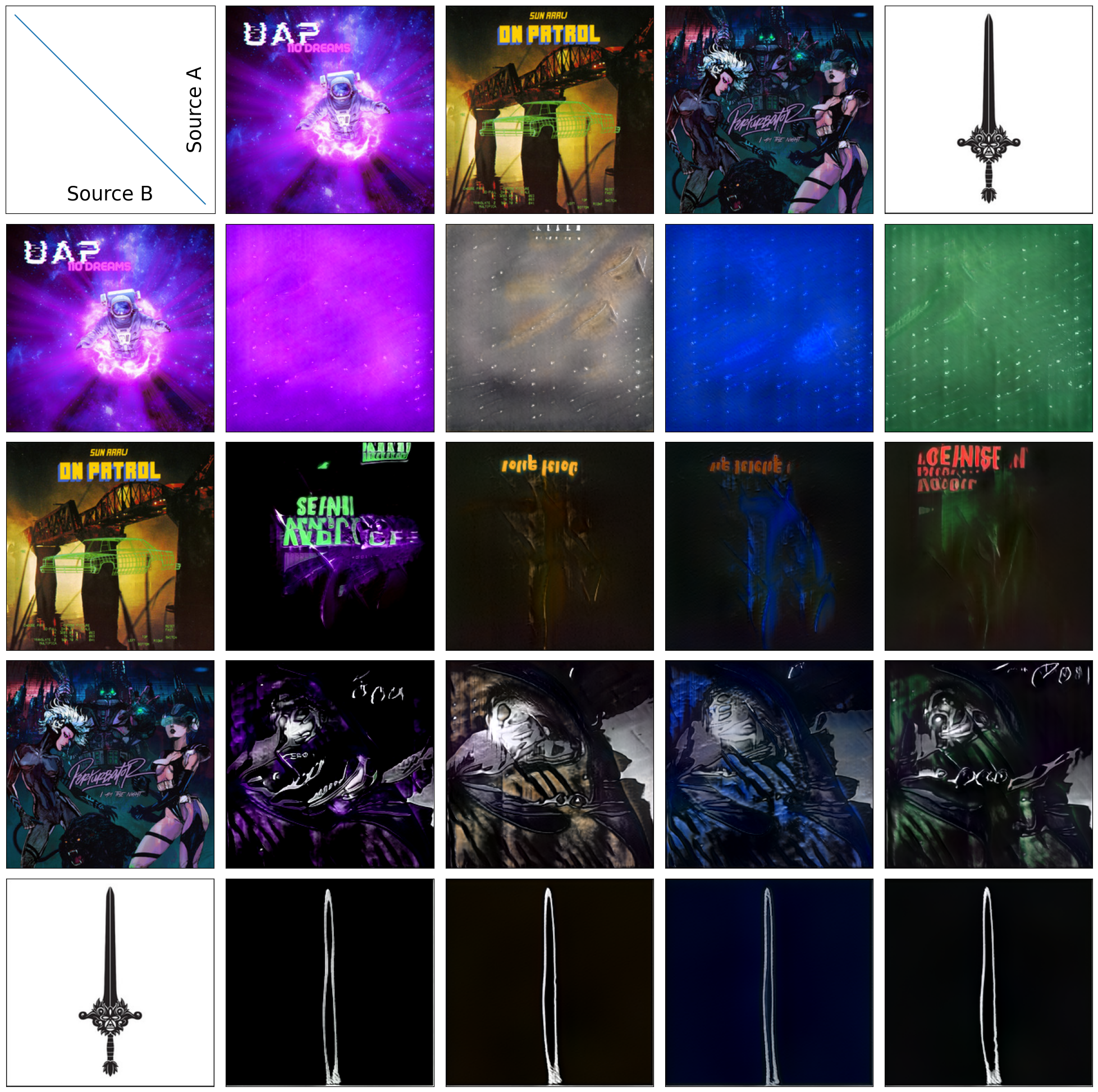}} 
        \quad
        \subfloat[Mixing 3 styles.]
        {\includegraphics[width=.7\linewidth]{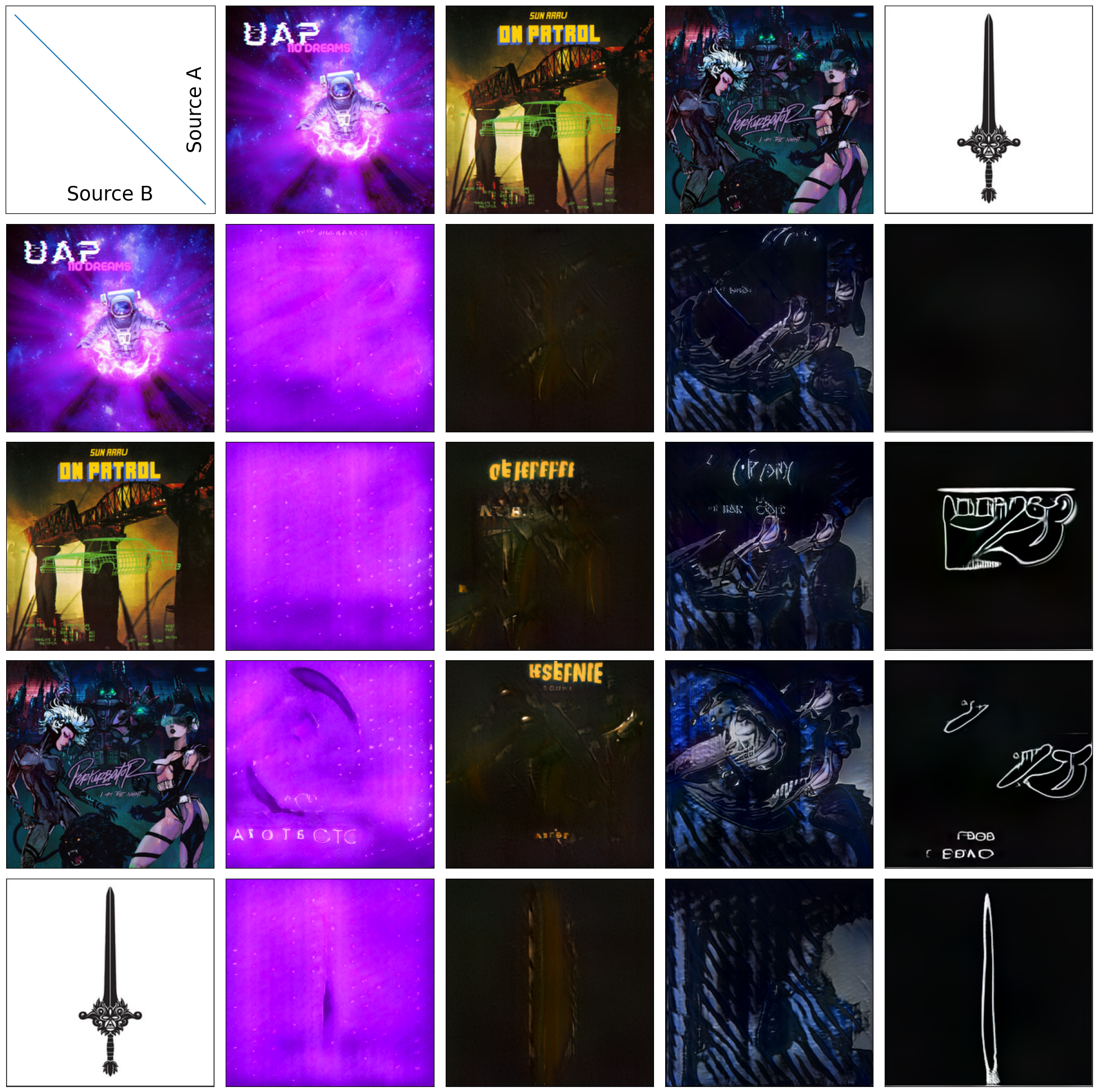}}
        \caption{Images generated when mixing a different number of styles.}
        \label{fig:12_3_styles}
        \vspace{-20px}
\end{figure}

It is also possible to mix a different number of styles. To test this, 3 and 12 styles were mixed as well. The results are shown in figure \ref{fig:12_3_styles}.

\subsection{Conclusion}

The results from this \ac{GAN} beyond satisfactory. Not only is then model able to capture the overall features of the input dataset but also the fine details. Furthermore, since the generated images are of higher resolution, these features can be appreciated better when looking at the covers.

The architecture that allows this model to style images was not only covered in depth, but used practically to generate styled images. Other methods of styling images were also attempted, like interpolating between 2 latent vectors. Since the interpolated vector exists at a middle point in the latent space, the generated image contains features from both, but there is no distinction in the styles, they all become mixed. With modifying the styles of the style vectors generated, individual features can be changed, like the color, with minimal disruption to other features like the shapes in the image.

The one downside is that this requires a comparatively long process after the model has been trained, since the latent vectors of the images need to be calculated before they are used, though this is mitigated with the fact that not as many images need to be downloaded.

%% file: Chapters/05Conclusion.tex
\chapter{Conclusion}\label{ch:Conclusion}

\section{Conclusion}

The purpose of this section is to determine to what extent the goal initially laid out in section \ref{goals} has been achieved.

The primary goal of this dissertation was to understand how the StyleGAN2 architecture works by generating album cover art. By documenting the analysis of 3 different types of \ac{GAN} models, training them, generating images with them, applying modifications and styling images, it is clear that this goal was achieved.

The other objectives accomplished are as follows;

\begin{enumerate}

\item{\textbf{Set up a repeatable development and training environment by using a custom Docker container}}

This objective was achieved by setting up a Docker container and installing the latest NVIDIA CUDA drivers for the 5.10 LTS Linux kernel. The Docker container was based on the Dockerfile provided in the StyleGAN2 GitHub page\footnote{\url{https://github.com/NVlabs/stylegan2-ada-pytorch/blob/main/Dockerfile}}. Necessary libraries were added for the other tasks needed as they became relevant, like performing API calls and plotting results.

\item{\textbf{Introducing theoretical framework of the model by providing explanations of the necessary concepts}}

This objective was achieved by first explaining 2 types of networks, \ac{NN} and \acp{CNN}. Then, generative models and their base architecture were documented. Finally, challenges of training \acp{GAN}, and how to overcome them, were discussed in section \ref{challenges_and_tips}.

\item{\textbf{Familiarize myself with existing \ac{GAN} architectures by analyzing and experimenting with 3 different \ac{GAN} models}}: 

This objective was achieved by looking at 3 different \ac{GAN} models.

    \begin{enumerate}
       \item{\textbf{Introductory GAN}}
       
       This preliminary model was used to document what a basic \ac{GAN} architecture's contents are in a practical sense. Images of handwritten digits of 24x24 pixels were generated with this \ac{GAN}, and the results were analyzed at at different epoch intervals. Analyzing these results demonstrated that an increased number of epochs does not linearly correlate with increased quality in the resulting images.
       
       \item{\textbf{DCGAN}}
       
       After analyzing the Keras \ac{GAN}, the next \ac{GAN} analyzed was the Celebrity faces \ac{DCGAN}. Celebrity faces \ac{DCGAN} was used since it uses \ac{RGB} images as the input and it generated higher resolution images (64x64 pixels) using celebrity faces as the input dataset, which have more features than handwritten digits. This model was trained instead with the Album covers dataset\footnote{https://www.kaggle.com/datasets/greg115/album-covers-images} for the purposes of this report. Furthermore, it was modified to see the effects of some optimization techniques discussed. Analyzing the generated images and the loss graphs demonstrated that the modified weight initialization and image normalization I implemented proved to stabilize the training, as seen in section \ref{modified_dcgan_conclusion}.
       
       \item{\textbf{StyleGAN2}}
       
       The final model used to generate the styled album covers, which improved understanding of style based architecture, by comparing and contrasting various code developed to style images.
       
    \end{enumerate}

\item{\textbf{Generate styled images by using the trained StyleGAN2 model}}

This objective was achieved through the experimentation done to reach the previous objectives, concretely, analyzing and experimenting with the StyleGAN2 architecture. Styled images generated with a different number of styles mixed to analyze the differences and the features that the model can distinguish between. Linear interpolation was also done between 2 latent space vectors, to see the differences between this method and style mixing.

\end{enumerate}

\section{Cost of the project}

This section displays the temporal cost of the project. The time spent on each phase of the project can be seen below:

\begin{itemize}
\item\texttt{Research}: Tasks related to the theoretical understanding of \ac{NN} and \acp{GAN}: \texttt{50h}

\item\texttt{Experimentation}: Tasks related to running code, evaluating results and tweaking model performance. \texttt{150h}

\item\texttt{Data manipulation}: Tasks related to acquiring the data and transforming it (projecting images to the latent space, using those vectors for style mixing or interpolation, etc.) \texttt{80h}

\item\texttt{Writing report}: Writing and revising the final documentation. \texttt{120h}

\begin{figure}[H]
        \myfloatalign
        {\label{fig:temporal_estimation}
        \includegraphics[width=.9\linewidth]{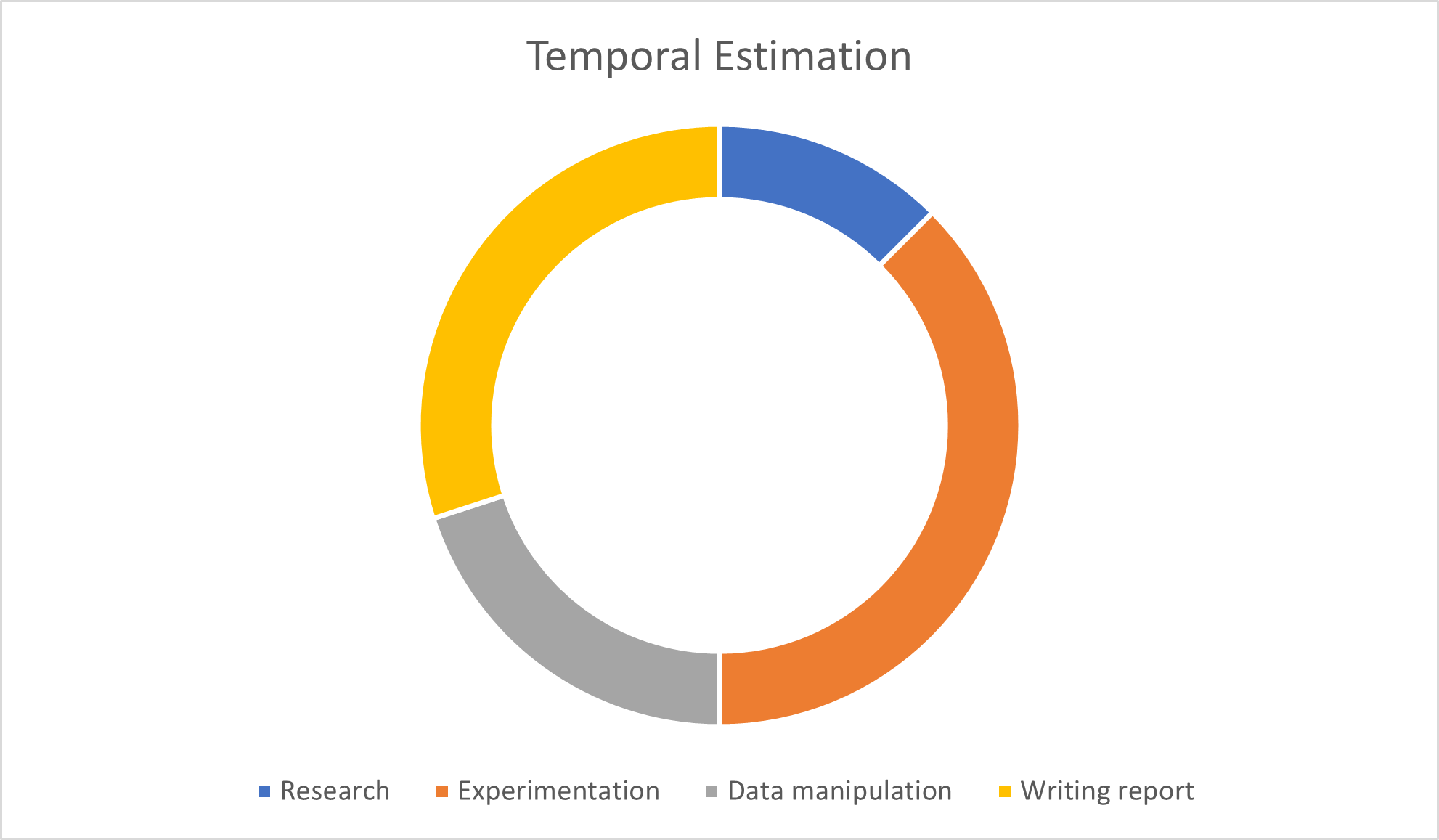}}
        \caption{Pie chart representing the ratio o f time spent on each task versus the total.}
\end{figure}

\end{itemize}

\section{Future work}

This section discusses future developments and potential improvements that could be performed to increase the quality of the generated images.

\begin{enumerate}
\item\texttt{\textbf{Using a different album dataset}}: The dataset used for training the StyleGAN2 was good in the sense that it contained a high amount of images, but the problem was that some of the album cover were vinyl discs. These had a round shape and I believe this did not help the model since it was trying to cover a very wide spectrum of images already, and having 2 formats of covers could confuse the model. The next step would be to train 2 models, one for circular images and one for square images.

\item\texttt{\textbf{Interpolate between more than 2 vectors}}: Since I wanted to explore and understand the architecture of StyleGAN2 more deeply, and wanted to combine the styles of images, I did not explore interpolating more than 2 latent space vectors. It would be beneficial to also see the results of the interpolation between more than 2 latent space vectors, since the results could provide more insight into the latent space of the model. Furthermore, interpolation between the style vectors themselves applying it to a single projected image would theoretically mean that the projected cover would have styles of many others. 

\item\texttt{\textbf{Publishing research}}: Since I trained and generated images with StyleGAN2, and my research led me to learn more deeply the intricacies of the architecture and also some practical knowledge on how to use the resulting model to generate latent space vectors of images and mix their styles, it would be beneficial to share these findings with the research community so that there is a reference report on style mixing with a model trained with album cover images.

\end{enumerate}

%% file: Chapters/Appendix.tex
\chapter{Appendix}
\section{Project tree structure}

The following folder is available to download via the following OneDrive link:

\url{https://lasalleuniversities-my.sharepoint.com/:f:/g/personal/felipe_perez_students_salle_url_edu/EnSWS2cmRoFDqha1juBPY_UBXwNq1yITGYe5p2VoUtCy8w?e=h6xpW5}

\begin{forest}
  for tree={
    font=\ttfamily,
    grow'=0,
    child anchor=west,
    parent anchor=south,
    anchor=west,
    calign=first,
    inner xsep=7pt,
    edge path={
      \noexpand\path [draw, \forestoption{edge}]
      (!u.south west) +(7.5pt,0) |- (.child anchor) pic {folder} \forestoption{edge label};
    },
    file/.style={
        edge path={\noexpand\path [draw, \forestoption{edge}]
        (!u.south west) +(7.5pt,0) |- (.child anchor) \forestoption{edge label};},
        inner xsep=2pt,font=\small\ttfamily
    },
    before typesetting nodes={
      if n=1
        {insert before={[,phantom]}}
        {}
    },
    fit=band,
    before computing xy={l=20pt},
    s sep=2pt,
  }
  [\textbf{TFG}
    [\textbf{Datasets}: Folder for datasets. Album covers dataset available at: \\[\url{https://www.kaggle.com/datasets/greg115/album-covers-images}, file]
    ]
    [\textbf{DCGAN}
        [\textbf{Models}]
        [\textbf{Results}]
        [dcgan.ipynb, file]
    ]
    [\textbf{Introductory GAN}
        [\textbf{Models}]
        [\textbf{Results}]
        [Intro.ipynb, file]
    ]
    [\textbf{StyleGAN2}
        [\textbf{Model}: Model files (network snapshot used + stats + fakes).]
        [\textbf{Results}: Folder for the results. Configurable in Final.ipynb]
        [\textbf{stylegan2-ada-pytorch}: Cloned StyleGAN2 repository.\\[(\url{https://github.com/NVlabs/stylegan2-ada-pytorch}), file]]
        [Final.ipynb, file]
    ]
    [\textbf{Docker}:
        [\textbf{PyTorch}: Dockerfile for running dcgan.ipynb and Final.ipynb notebooks.
            [Dockerfile, file]
        ]
        [\textbf{KerasTF}: Contains Dockerfile for running Intro.ipynb notebook.
            [Dockerfile, file]
        ]
        [container\_start\_cmds.txt: Commands for the Dockerfiles and containers., file]
    ]
]
\end{forest}

%% file: FrontBackmatter/Bibliography.tex
\manualmark
\markboth{\spacedlowsmallcaps{\bibname}}{\spacedlowsmallcaps{\bibname}} 
\refstepcounter{dummy}
\addtocontents{toc}{\protect\vspace{\beforebibskip}} 
\addcontentsline{toc}{chapter}{\tocEntry{\bibname}}
\label{app:bibliography}
\printbibliography

%% file: ClassicThesis.bib
@video{3blue1brown_gradient_2017,
	title = {Gradient descent, how neural networks learn {\textbar} Chapter 2, Deep learning},
	url = {https://www.youtube.com/watch?v=IHZwWFHWa-w},
	author = {{3Blue1Brown}},
	urldate = {2022-03-20},
	date = {2017-10-16},
}

@online{mandal_cnn_2021,
	title = {{CNN} for Deep Learning {\textbar} Convolutional Neural Networks},
	url = {https://www.analyticsvidhya.com/blog/2021/05/convolutional-neural-networks-cnn/},
	titleaddon = {Analytics Vidhya},
	author = {Mandal, Manav},
	urldate = {2022-03-21},
	date = {2021-05-01},
	langid = {english},
}

@online{noauthor_student_2018,
	title = {Student Notes: Convolutional Neural Networks ({CNN}) Introduction},
	url = {https://indoml.com/2018/03/07/student-notes-convolutional-neural-networks-cnn-introduction/},
	shorttitle = {Student Notes},
	titleaddon = {Belajar Pembelajaran Mesin Indonesia},
	urldate = {2022-03-21},
	date = {2018-03-07},
	langid = {indonesian},
}

@inreference{noauthor_convolutional_2022,
	title = {Convolutional neural network},
	rights = {Creative Commons Attribution-{ShareAlike} License},
	url = {https://en.wikipedia.org/w/index.php?title=Convolutional_neural_network&oldid=1075395384},
	booktitle = {Wikipedia},
	urldate = {2022-03-23},
	date = {2022-03-05},
	langid = {english},
	note = {Page Version {ID}: 1075395384},
}

@online{noauthor_mean_nodate,
	title = {Mean squared error loss function {\textbar} Peltarion Platform},
	url = {https://peltarion.com/knowledge-center/documentation/modeling-view/build-an-ai-model/loss-functions/mean-squared-error},
	titleaddon = {Peltarion},
	urldate = {2022-04-05},
	langid = {english},
}

@online{brownlee_gentle_2019,
	title = {A Gentle Introduction to Generative Adversarial Networks ({GANs})},
	url = {https://machinelearningmastery.com/what-are-generative-adversarial-networks-gans/},
	titleaddon = {Machine Learning Mastery},
	author = {Brownlee, Jason},
	urldate = {2022-04-05},
	date = {2019-06-16},
	langid = {american},
}

@video{3blue1brown_but_2017,
	title = {But what is a neural network? {\textbar} Chapter 1, Deep learning},
	url = {https://www.youtube.com/watch?v=aircAruvnKk},
	shorttitle = {But what is a neural network?},
	author = {{3Blue1Brown}},
	urldate = {2022-04-16},
	date = {2017-10-05},
}

@online{brownlee_tips_2019,
	title = {Tips for Training Stable Generative Adversarial Networks},
	url = {https://machinelearningmastery.com/how-to-train-stable-generative-adversarial-networks/},
	titleaddon = {Machine Learning Mastery},
	author = {Brownlee, Jason},
	urldate = {2022-04-16},
	date = {2019-06-18},
	langid = {american},
}

@article{raissi_parameter_2019,
	title = {On parameter estimation approaches for predicting disease transmission through optimization, deep learning and statistical inference methods},
	doi = {10.1080/23737867.2019.1676172},
	pages = {1--26},
	journaltitle = {Letters in Biomathematics},
	author = {Raissi, Maziar and Ramezani, Niloofar and Padmanabhan, Seshaiyer},
	date = {2019-10},
}

@thesis{ioannou_structural_2017,
	title = {Structural Priors in Deep Neural Networks},
	type = {phdthesis},
	author = {Ioannou, Yani},
	date = {2017-09},
	doi = {10.17863/CAM.26357},
}

@online{noauthor_gan_nodate,
	title = {{GAN} Training {\textbar} Generative Adversarial Networks},
	url = {https://developers.google.com/machine-learning/gan/training},
	titleaddon = {Google Developers},
	urldate = {2022-04-16},
	langid = {english},
}

@article{karras_training_2020,
	title = {Training Generative Adversarial Networks with Limited Data},
	url = {http://arxiv.org/abs/2006.06676},
	journaltitle = {{arXiv}:2006.06676 [cs, stat]},
	author = {Karras, Tero and Aittala, Miika and Hellsten, Janne and Laine, Samuli and Lehtinen, Jaakko and Aila, Timo},
	urldate = {2022-04-16},
	date = {2020-10-07},
	eprinttype = {arxiv},
	eprint = {2006.06676},
}

@article{karras_progressive_2018,
	title = {Progressive Growing of {GANs} for Improved Quality, Stability, and Variation},
	url = {http://arxiv.org/abs/1710.10196},
	journaltitle = {{arXiv}:1710.10196 [cs, stat]},
	author = {Karras, Tero and Aila, Timo and Laine, Samuli and Lehtinen, Jaakko},
	urldate = {2022-04-16},
	date = {2018-02-26},
	eprinttype = {arxiv},
	eprint = {1710.10196},
}

@online{dellinger_weight_2019,
	title = {Weight Initialization in Neural Networks: A Journey From the Basics to Kaiming},
	url = {https://towardsdatascience.com/weight-initialization-in-neural-networks-a-journey-from-the-basics-to-kaiming-954fb9b47c79},
	shorttitle = {Weight Initialization in Neural Networks},
	titleaddon = {Medium},
	author = {Dellinger, James},
	urldate = {2022-04-19},
	date = {2019-04-04},
	langid = {english},
}

@online{brownlee_how_2019-1,
	title = {How to Implement {GAN} Hacks in Keras to Train Stable Models},
	url = {https://machinelearningmastery.com/how-to-code-generative-adversarial-network-hacks/},
	titleaddon = {Machine Learning Mastery},
	author = {Brownlee, Jason},
	urldate = {2022-04-19},
	date = {2019-06-20},
	langid = {american},
}

@online{brownlee_gentle_2019-1,
	title = {A Gentle Introduction to Batch Normalization for Deep Neural Networks},
	url = {https://machinelearningmastery.com/batch-normalization-for-training-of-deep-neural-networks/},
	titleaddon = {Machine Learning Mastery},
	author = {Brownlee, Jason},
	urldate = {2022-04-19},
	date = {2019-01-15},
	langid = {american},
}

@article{srivastava_dropout_2014,
	title = {Dropout: A Simple Way to Prevent Neural Networks from Overfitting},
	volume = {15},
	issn = {1533-7928},
	url = {http://jmlr.org/papers/v15/srivastava14a.html},
	shorttitle = {Dropout},
	pages = {1929--1958},
	number = {56},
	journaltitle = {Journal of Machine Learning Research},
	author = {Srivastava, Nitish and Hinton, Geoffrey and Krizhevsky, Alex and Sutskever, Ilya and Salakhutdinov, Ruslan},
	urldate = {2022-04-19},
	date = {2014},
}

@online{brownlee_how_2018,
	title = {How to Reduce Overfitting With Dropout Regularization in Keras},
	url = {https://machinelearningmastery.com/how-to-reduce-overfitting-with-dropout-regularization-in-keras/},
	titleaddon = {Machine Learning Mastery},
	author = {Brownlee, Jason},
	urldate = {2022-04-19},
	date = {2018-12-04},
	langid = {american},
}

@software{chintala_how_2022,
	title = {How to Train a {GAN}? Tips and tricks to make {GANs} work},
	url = {https://github.com/soumith/ganhacks},
	shorttitle = {How to Train a {GAN}?},
	author = {Chintala, Soumith},
	urldate = {2022-04-19},
	date = {2022-04-19},
	note = {original-date: 2016-12-09T16:09:27Z},
}

@online{brownlee_how_2019-2,
	title = {How to Implement the Frechet Inception Distance ({FID}) for Evaluating {GANs}},
	url = {https://machinelearningmastery.com/how-to-implement-the-frechet-inception-distance-fid-from-scratch/},
	titleaddon = {Machine Learning Mastery},
	author = {Brownlee, Jason},
	urldate = {2022-04-19},
	date = {2019-08-29},
	langid = {american},
}

@software{noauthor_nvlabsstylegan2-ada-pytorch_2022,
	title = {{NVlabs}/stylegan2-ada-pytorch},
	url = {https://github.com/NVlabs/stylegan2-ada-pytorch},
	publisher = {{NVIDIA} Research Projects},
	urldate = {2022-04-20},
	date = {2022-04-20},
	note = {original-date: 2021-01-12T16:28:39Z},
}

@online{noauthor_pytorch_nodate,
	title = {{PyTorch} documentation — {PyTorch} 1.11.0 documentation},
	url = {https://pytorch.org/docs/stable/index.html},
	urldate = {2022-04-23},
}

@online{noauthor_keras_2020,
	title = {Keras vs Tensorflow vs Pytorch [Updated] {\textbar} Deep Learning Frameworks {\textbar} Simplilearn},
	url = {https://www.simplilearn.com/keras-vs-tensorflow-vs-pytorch-article},
	titleaddon = {Simplilearn.com},
	urldate = {2022-04-23},
	date = {2020-07-27},
	langid = {american},
}

@inreference{noauthor_pytorch_2022,
	title = {{PyTorch}},
	rights = {Creative Commons Attribution-{ShareAlike} License},
	url = {https://en.wikipedia.org/w/index.php?title=PyTorch&oldid=1082017868},
	booktitle = {Wikipedia},
	urldate = {2022-04-23},
	date = {2022-04-10},
	langid = {english},
	note = {Page Version {ID}: 1082017868},
}

@online{noauthor_numpy_nodate,
	title = {{NumPy} documentation — {NumPy} v1.22 Manual},
	url = {https://numpy.org/doc/stable/},
	urldate = {2022-04-23},
}

@online{noauthor_gpu_nodate,
	title = {{GPU} for Deep Learning},
	url = {https://www.run.ai/guides/gpu-deep-learning},
	urldate = {2022-04-23},
}

@online{noauthor_cuda_2017,
	title = {{CUDA} Zone},
	url = {https://developer.nvidia.com/cuda-zone},
	titleaddon = {{NVIDIA} Developer},
	urldate = {2022-04-23},
	date = {2017-07-18},
	langid = {american},
}

@online{ekoulier_answer_2018,
	title = {Answer to "Why is tanh almost always better than sigmoid as an activation function?"},
	url = {https://stats.stackexchange.com/a/330565},
	shorttitle = {Answer to "Why is tanh almost always better than sigmoid as an activation function?},
	titleaddon = {Cross Validated},
	author = {ekoulier},
	urldate = {2022-04-24},
	date = {2018-02-26},
}

@inreference{noauthor_activation_2022,
	title = {Activation function},
	rights = {Creative Commons Attribution-{ShareAlike} License},
	url = {https://en.wikipedia.org/w/index.php?title=Activation_function&oldid=1084483380},
	booktitle = {Wikipedia},
	urldate = {2022-04-25},
	date = {2022-04-24},
	langid = {english},
	note = {Page Version {ID}: 1084483380},
}

@online{noauthor_vanishing_2021,
	title = {Vanishing and Exploding Gradients in Deep Neural Networks},
	url = {https://www.analyticsvidhya.com/blog/2021/06/the-challenge-of-vanishing-exploding-gradients-in-deep-neural-networks/},
	titleaddon = {Analytics Vidhya},
	urldate = {2022-04-25},
	date = {2021-06-18},
	langid = {english},
}

@online{noauthor_tanhwolfram_nodate,
	title = {Tanh—Wolfram Language Documentation},
	url = {https://reference.wolfram.com/language/ref/Tanh.html},
	urldate = {2022-04-25},
}

@inproceedings{matougui_k-mer_2019,
	title = {A K-mer based Multi Convolutional Neural Network Classifier of Low-Ranking Taxonomic Bins from Metagenome},
	author = {Matougui, Brahim and Batouche, M. and Boukelia, Abdelbasset},
	date = {2019-09},
}

@online{brownlee_gentle_2019-2,
	title = {A Gentle Introduction to the Rectified Linear Unit ({ReLU})},
	url = {https://machinelearningmastery.com/rectified-linear-activation-function-for-deep-learning-neural-networks/},
	titleaddon = {Machine Learning Mastery},
	author = {Brownlee, Jason},
	urldate = {2022-04-25},
	date = {2019-01-08},
	langid = {american},
}

@online{brownlee_weight_2021,
	title = {Weight Initialization for Deep Learning Neural Networks},
	url = {https://machinelearningmastery.com/weight-initialization-for-deep-learning-neural-networks/},
	titleaddon = {Machine Learning Mastery},
	author = {Brownlee, Jason},
	urldate = {2022-04-27},
	date = {2021-02-02},
	langid = {american},
}

@article{he_delving_2015,
	title = {Delving Deep into Rectifiers: Surpassing Human-Level Performance on {ImageNet} Classification},
	url = {http://arxiv.org/abs/1502.01852},
	shorttitle = {Delving Deep into Rectifiers},
	journaltitle = {{arXiv}:1502.01852 [cs]},
	author = {He, Kaiming and Zhang, Xiangyu and Ren, Shaoqing and Sun, Jian},
	urldate = {2022-04-27},
	date = {2015-02-06},
	eprinttype = {arxiv},
	eprint = {1502.01852},
	note = {version: 1},
}

@inproceedings{yani_application_2019,
	title = {Application of transfer learning using convolutional neural network method for early detection of terry’s nail},
	volume = {1201},
	pages = {012052},
	booktitle = {Journal of Physics: Conference Series},
	publisher = {{IOP} Publishing},
	author = {Yani, Muhamad and {others}},
	date = {2019},
	note = {Issue: 1},
}

@article{skandarani_gans_2021,
	title = {Gans for medical image synthesis: An empirical study},
	journaltitle = {{arXiv} preprint {arXiv}:2105.05318},
	author = {Skandarani, Youssef and Jodoin, Pierre-Marc and Lalande, Alain},
	date = {2021},
}

@video{finnish_center_for_artificial_intelligence_fcai_tero_2021,
	title = {Tero Karras - Training Generative Adversarial Networks with Limited Data},
	url = {https://www.youtube.com/watch?v=hOx9NBwDkHY},
	author = {{Finnish Center for Artificial Intelligence FCAI}},
	urldate = {2022-05-01},
	date = {2021-01-20},
}

@inproceedings{krizhevsky_imagenet_2012,
	title = {{ImageNet} Classification with Deep Convolutional Neural Networks},
	volume = {25},
	url = {https://proceedings.neurips.cc/paper/2012/file/c399862d3b9d6b76c8436e924a68c45b-Paper.pdf},
	booktitle = {Advances in Neural Information Processing Systems},
	publisher = {Curran Associates, Inc.},
	author = {Krizhevsky, Alex and Sutskever, Ilya and Hinton, Geoffrey E},
	editor = {Pereira, F. and Burges, C. J. and Bottou, L. and Weinberger, K. Q.},
	date = {2012},
}

@article{gulrajani_improved_2017,
	title = {Improved Training of Wasserstein {GANs}},
	url = {http://arxiv.org/abs/1704.00028},
	journaltitle = {{arXiv}:1704.00028 [cs, stat]},
	author = {Gulrajani, Ishaan and Ahmed, Faruk and Arjovsky, Martin and Dumoulin, Vincent and Courville, Aaron},
	urldate = {2022-05-03},
	date = {2017-12-25},
	eprinttype = {arxiv},
	eprint = {1704.00028},
}

@online{brownlee_gentle_2019-3,
	title = {A Gentle Introduction to {StyleGAN} the Style Generative Adversarial Network},
	url = {https://machinelearningmastery.com/introduction-to-style-generative-adversarial-network-stylegan/},
	titleaddon = {Machine Learning Mastery},
	author = {Brownlee, Jason},
	urldate = {2022-05-07},
	date = {2019-08-18},
	langid = {american},
}

@inproceedings{karras_style-based_2019,
	title = {A style-based generator architecture for generative adversarial networks},
	pages = {4401--4410},
	booktitle = {Proceedings of the {IEEE}/{CVF} conference on computer vision and pattern recognition},
	author = {Karras, Tero and Laine, Samuli and Aila, Timo},
	date = {2019},
}

@inreference{noauthor_hotline_2022,
	title = {\textit{Hotline Miami}},
	rights = {Creative Commons Attribution-{ShareAlike} License},
	url = {https://en.wikipedia.org/w/index.php?title=Hotline_Miami&oldid=1082119366},
	booktitle = {Wikipedia},
	urldate = {2022-05-07},
	date = {2022-04-11},
	langid = {english},
	note = {Page Version {ID}: 1082119366},
}

@article{marchesi_megapixel_2017,
	title = {Megapixel size image creation using generative adversarial networks},
	journaltitle = {{arXiv} preprint {arXiv}:1706.00082},
	author = {Marchesi, Marco},
	date = {2017},
}

@online{laptrinhx_understanding_2021,
	title = {Understanding Optimization Algorithms},
	url = {https://laptrinhx.com/understanding-optimization-algorithms-3818430905/},
	titleaddon = {{LaptrinhX}},
	author = {{LaptrinhX}},
	urldate = {2022-05-11},
	date = {2021-03-14},
	langid = {english},
}

@online{md_these_2021,
	title = {These football clubs don’t exist — Sharing my experience with {StyleGAN}},
	url = {https://medium.com/analytics-vidhya/these-football-clubs-dont-exist-sharing-my-experience-with-stylegan-7d02e4b34914},
	titleaddon = {Analytics Vidhya},
	author = {{MD}, Jae Won Choi},
	urldate = {2022-05-12},
	date = {2021-06-14},
	langid = {english},
}

@inreference{noauthor_variational_2022,
	title = {Variational autoencoder},
	rights = {Creative Commons Attribution-{ShareAlike} License},
	url = {https://en.wikipedia.org/w/index.php?title=Variational_autoencoder&oldid=1087184794},
	booktitle = {Wikipedia},
	urldate = {2022-05-16},
	date = {2022-05-10},
	langid = {english},
	note = {Page Version {ID}: 1087184794},
}

@online{daemonmaker_answer_2014,
	title = {Answer to "What are the advantages of {ReLU} over sigmoid function in deep neural networks?"},
	url = {https://stats.stackexchange.com/a/126362},
	shorttitle = {Answer to "What are the advantages of {ReLU} over sigmoid function in deep neural networks?},
	titleaddon = {Cross Validated},
	author = {{DaemonMaker}},
	urldate = {2022-05-17},
	date = {2014-12-03},
}

@software{noauthor_nvlabsstylegan_2022,
	title = {{NVlabs}/stylegan},
	url = {https://github.com/NVlabs/stylegan},
	publisher = {{NVIDIA} Research Projects},
	urldate = {2022-05-17},
	date = {2022-05-17},
	note = {original-date: 2019-02-04T15:33:58Z},
}

@online{noauthor_papers_nodate,
	title = {Papers with Code - Leaky {ReLU} Explained},
	url = {https://paperswithcode.com/method/leaky-relu},
	urldate = {2022-05-18},
	langid = {english},
}

@online{brownlee_supervised_2016,
	title = {Supervised and Unsupervised Machine Learning Algorithms},
	url = {https://machinelearningmastery.com/supervised-and-unsupervised-machine-learning-algorithms/},
	titleaddon = {Machine Learning Mastery},
	author = {Brownlee, Jason},
	urldate = {2022-05-18},
	date = {2016-03-15},
	langid = {american},
}

@online{noauthor_classical_nodate,
	title = {Classical {ML} Equations in {LaTeX}},
	url = {https://blmoistawinde.github.io/ml_equations_latex/#cross-entropy},
	urldate = {2022-05-18},
}

@article{goodfellow_generative_2014,
	title = {Generative adversarial nets},
	volume = {27},
	journaltitle = {Advances in neural information processing systems},
	author = {Goodfellow, Ian and Pouget-Abadie, Jean and Mirza, Mehdi and Xu, Bing and Warde-Farley, David and Ozair, Sherjil and Courville, Aaron and Bengio, Yoshua},
	date = {2014},
}
